\pdfoutput=1
\documentclass{article} %
\usepackage{iclr2027_preprint,times}

\usepackage{amsmath,amsfonts,bm}

\def\eqref#1{equation~\ref{#1}}

\def\1{\bm{1}}

\def\mX{{\bm{X}}}

\DeclareMathAlphabet{\mathsfit}{\encodingdefault}{\sfdefault}{m}{sl}
\SetMathAlphabet{\mathsfit}{bold}{\encodingdefault}{\sfdefault}{bx}{n}

\def\gE{{\mathcal{E}}}

\def\gG{{\mathcal{G}}}

\def\gV{{\mathcal{V}}}

\usepackage{hyperref}
\usepackage{url}

\usepackage[labelfont=bf]{caption}
\usepackage{microtype}
\usepackage{graphicx}
\usepackage{subcaption}
\usepackage{booktabs} %
\usepackage{multicol}
\usepackage{multirow}
\usepackage{colortbl}
\usepackage{enumitem}
\usepackage[capitalize,noabbrev]{cleveref}

\usepackage[most]{tcolorbox}

\newtcolorbox{finding}{
  breakable,
  colback=babyblue!45,
  colframe=uclablue,
  boxrule=0.6pt,
  arc=2pt,
  left=7pt, right=7pt, top=3pt, bottom=3pt,
  fontupper=\small,
  before skip=5pt, after skip=7pt,
}
\newcommand{\take}{\textbf{\textcolor{uclablue}{Findings:}}\ }

\newcommand{\denovo}{\textit{de novo }}
\newcommand{\frag}{$f$-RAG }

\definecolor{uclablue}{rgb}{0.15, 0.45, 0.68}
\definecolor{aliceblue}{RGB}{178, 217, 245}
\newcommand{\CC}{\cellcolor{aliceblue}}
\definecolor{babyblue}{RGB}{217, 239, 251}

\definecolor{babypink}{RGB}{251, 231, 230}
\newcommand{\CP}{\cellcolor{babypink}}

\crefname{appendix}{Appx.}{Apps.}
\crefname{subappendix}{Appx.}{Apps.}
\crefname{subsubappendix}{Appx.}{Apps.}
\crefname{section}{\S}{\S}
\crefname{table}{Tab.}{Tabs.}

\title{A Large Scale Investigation of Scaling Limits in Chemical
Language Models}

\newlength{\authcolw}
\newcommand{\authline}[1]{\makebox[\authcolw][l]{#1}}
\author{%
Roshan Balaji\thanks{Equal contribution.} \\
\authline{BioSystems Engineering and Control Lab;} \\
Wadhwani School of Data Science and AI, \\
IIT Madras; \\
Centre for Integrative Biology and \\
Systems medicinE (IBSE) \\
\And
Kamran Chitsaz\footnotemark[1] \\
\authline{Chandar Research Lab;} \\
Mila -- Quebec AI Institute \\
\And
Quentin Fournier \\
\authline{Mila -- Quebec AI Institute} \\
\AND
Nirav Pravinbhai Bhatt\thanks{Equal supervision; corresponding author.} \\
\authline{BioSystems Engineering and Control Lab;} \\
Wadhwani School of Data Science and AI, \\
IIT Madras; \\
Centre for Integrative Biology and \\
Systems medicinE (IBSE); \\
IIT Madras Zanzibar \\
\texttt{niravbhatt@iitm.ac.in} \\
\And
Sarath Chandar\footnotemark[2] \\
\authline{Chandar Research Lab;} \\
Mila -- Quebec AI Institute; \\
Polytechnique Montr\'eal; \\
Canada CIFAR AI Chair \\
\texttt{sarath.chandar@mila.quebec}
}

\iclrfinalcopy
\begin{document}

\maketitle

\begin{abstract}
  Chemical Language Models (CLMs) are increasingly used in \denovo drug design, driven by recent growth in model scale, compute, and dataset size. However, the relationship between design choices, training dynamics, and downstream generation quality remains poorly understood. We present a compute-controlled scaling study of CLMs comprising more than 30,000 experiments across molecular representations (SMILES, SELFIES, SAFE), tokenizations (atom-level and byte-pair encoding), model scales (0.5M–1B parameters), leakage-controlled datasets (MOSES, ChEMBL, PubChem, ZINC-22), and architectures (decoder-only and encoder–decoder). By fitting IsoFLOP profiles, we establish clear scaling trends in pretraining loss, but find that these improvements do not translate into comparable gains in goal-directed molecular design. Layer-wise probing and sparse autoencoder analysis reveal continued development of chemical representations: chemical syntax saturates early, while semantic properties emerge more slowly and become increasingly accessible with further training and model scale. These representational gains coexist with diminishing improvements in goal-directed generation under the evaluated protocols and oracle budgets. Our resulting suite of models, NovoMolGen, achieves state-of-the-art results, outperforming prior CLMs and specialized generative models in goal-directed molecular generation across drug discovery tasks. These findings expose a disconnect between chemical representation learning and downstream molecular design, motivating the development of pretraining paradigms that more directly learn chemical semantics.
  
\end{abstract}

\begin{figure}[t]
  \centering
  \includegraphics[width=0.9\linewidth]{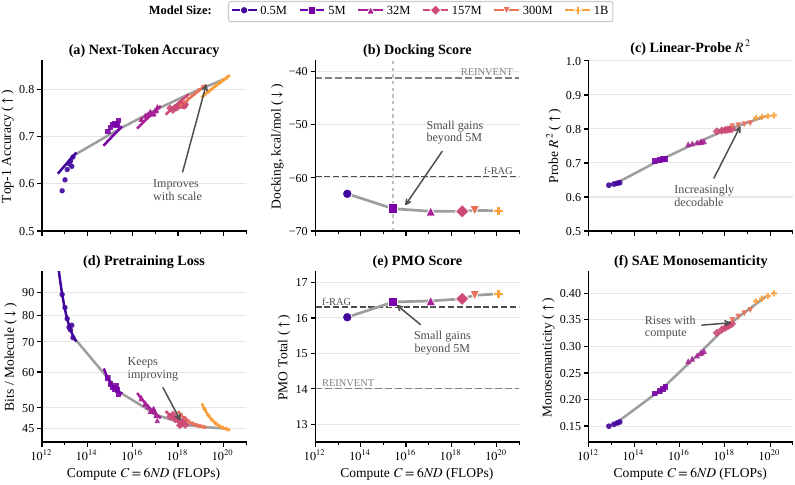}
  \caption{\textbf{Pretraining and internal representations improve
      with compute, while goal-directed generation shows diminishing
    returns.} Metrics against pretraining compute $C = 6ND$ for models
    from 0.5M to 1B parameters. \textbf{(a)}~Next-token accuracy rises
    with scale. \textbf{(b,\,e)}~After RL fine-tuning, the total top
    5\% docking score over five targets ($\downarrow$) and the total PMO score
    over 22 tasks show only small gains beyond $\sim$5M.
    \textbf{(c)}~Middle-layer linear-probe $R^2$,
    averaged over 36 property targets, keeps rising. \textbf{(d)}~Validation set
    bits per molecule decrease steadily as compute increases.
    \textbf{(f)}~SAE monosemanticity, the mean over features of the largest
  absolute property correlation, keeps rising.}
  \label{fig:teaser_figure}
\end{figure}

\section{Introduction}
\label{sec:intro}

Chemical Language Models (CLMs) have largely adopted the scaling
paradigm of Natural Language Processing (NLP), pretraining
increasingly larger autoregressive transformers on massive molecular
corpora. The implicit assumption is that minimising next-token
cross-entropy translates directly into better molecular
design~\citep{olivecrona_molecular_2017,bagal_molgpt_2022,ross_gp-molformer_2024}.
In NLP, scaling laws describe predictable improvements in language
modeling as model size, training data, and compute
increase~\citep{kaplan_scaling_2020,hoffmannTrainingComputeoptimalLarge2022}. 
However, \denovo molecular generation optimises candidates using
downstream iteration loops against biological and physicochemical
property oracles~\citep{gao_sample_2022}. 
Understanding molecular scaling therefore requires tracing its
effects from sequence modeling and learned chemical representations
to goal-directed generation. Whether continued improvements in the
former translate into comparable gains in molecular design remains
poorly understood. Recent studies have explored scaling behaviour in CLMs, revealing
different outcomes across evaluation settings. While
\citet{frey2023neural} observed scaling in CLMs, their analysis did not establish how these gains translate
to goal-directed molecular generation. More recently,
\citet{xu2026unveiling} studied compute-controlled scaling across
representations and property prediction, while
\citet{medinaDiversityBeatsSize2026} observed that hit-generation
rates saturate with model scale in reinforcement-learning-based
molecular optimisation. However, these studies do not examine how goal-directed generation
changes along the pretraining scaling curve, nor do they account for
the cost of downstream optimisation. These findings motivate our central hypothesis: \textit{improvements along the compute-optimal pretraining frontier do not necessarily translate into better goal-directed design.}

To establish how scaling translates into molecular design performance,
we study CLMs using IsoFLOP profiling across
model scales spanning 0.5M to 1B parameters. Our experimental design
includes controlled comparisons of transformer architectures
(decoder-only and encoder--decoder), string representations (SMILES,
  SELFIES,
SAFE)~\citep{weininger_smiles_1988,krenn_selfies_2022,noutahi_gotta_2024},
tokenisation strategies (atom-level and byte-pair
encoding)~\citep{wadellTokenizationMolecularFoundation2026}, and
datasets (MOSES~\citep{polykovskiy_molecular_2020},
  ChEMBL~\citep{Gaulton2012ChEMBL:Discovery},
  PubChem~\citep{Kim2016PubChemDatabases}, and
ZINC-22~\citep{tingle_zinc-22free_2023}). 
To support comparisons across tokenisations and representations,
we report validation set negative log-likelihood in bits per molecule,
normalising loss per molecule rather than per token.
Checkpoints are evaluated on the
Practical Molecular Optimisation (PMO)
benchmark~\citep{gao_sample_2022} and structure-based
docking~\citep{alhossary_fast_2015} under fixed oracle budgets, with
representations analysed using layer-wise
probing~\citep{belinkov2022probing} and BatchTopK Sparse Autoencoders
(SAEs)~\citep{bussmann2024batchtopk}.

Our large-scale empirical investigation in this work reveals a paradox between
pretraining efficiency, internal representations, and practical
generative performance (\cref{fig:teaser_figure}):
\begin{itemize}[leftmargin=14pt,noitemsep,topsep=2pt]
  \item Pretraining cross-entropy loss, measured in bits per molecule,
    decreases smoothly with increasing compute. IsoFLOP profiles
    reveal power-law relationships between compute-optimal model
    size and training tokens (\(N_{\mathrm{opt}} \propto C^{a}\),
    \(D_{\mathrm{opt}} \propto C^{b}\)), demonstrating that chemical
    sequence modelling follows compute-optimal scaling trends
    within the studied regime.

  \item Latent chemical representations become richer with
    increasing compute, as revealed by layer-wise linear probes
    and SAE feature dictionaries. Models acquire basic chemical
    syntax in early checkpoints, while molecular properties and
    functional-group features become more accessible during
    subsequent training.

  \item Downstream goal-directed generation largely saturates beyond
    approximately 5M parameters, despite improvements in bits per
    molecule and linear accessibility of chemical features. Performance on PMO and structure-based docking flattens under
    the evaluated optimisation protocols and oracle budgets.
    A 5M-parameter model achieves performance comparable to models
    scaling up to 1B parameters across multi-objective optimisation
    tasks.
\end{itemize}

These findings demonstrate that scaling parameter counts under
current self-supervised string objectives yields diminishing returns
in downstream molecular design, despite continued improvements in
pretraining loss and learned chemical representations. Because our
probing shows that additional chemical features remain accessible
during training, this divergence suggests that the evaluated
objectives may limit their downstream utility. Practically, compact
CLMs (\(\sim\)5M parameters) preserve competitive downstream
performance while substantially reducing the computational cost of
generation and policy updates. To motivate better evaluation
standards, we open-source the
NovoMolGen family of CLMs, releasing our pretraining checkpoints and downstream task pipelines.

\section{Experimental Setup}
\label{sec:methodology}

We study how pretraining scale and design choices affect \denovo
molecular generation. After describing the datasets and design axes
(\cref{sec:experimental_design}), we evaluate every model at three
levels: validation set likelihood (\cref{sec:scaling_protocol}), the linear
accessibility of chemical information in its representations
(\cref{sec:mechanistic_probing}), and goal-directed optimisation under
fixed oracle budgets (\cref{sec:finetuning}).

\subsection{Datasets and Design Axes}
\label{sec:experimental_design}

We pretrain our models on datasets of increasing size and diversity: MOSES~\citep{polykovskiy_molecular_2020},
ChEMBL~\citep{Gaulton2012ChEMBL:Discovery},
PubChem~\citep{Kim2016PubChemDatabases}, and
ZINC-22~\citep{tingle_zinc-22free_2023}. For ZINC-22, the largest
publicly available molecular library ($\approx$96B synthesisable
compounds), we sample 1.5B molecules proportionally
across its heavy-atom-count and logP tranches. To prevent data leakage,
we remove from every corpus all molecules with a Tanimoto similarity
above 0.5 (2,048-bit ECFP, radius 2) to any molecule in a set of
$\sim$21,000 known actives for the PMO and inhibitors for docking targets and
top 1\% molecules from the Tartarus benchmarks~\citep{nigam_tartarus_2023}.
This active (molecule) set also serves as an unseen set
for probing and interpretability. We create four evaluation splits with increasing levels of distributional shift. These include a random split, a scaffold split of held-out Bemis--Murcko scaffolds, a
molecular-weight split of the heaviest molecules, and a
maximum-dissimilarity split of the fingerprint clusters farthest from
the corpus. The random split is the in-distribution
reference, and the other three measure generalisation. We vary the model size (0.5M--1B parameters), molecular representation (SMILES, SELFIES, SAFE), tokenisation (atom-level, byte-pair encoding), pretraining corpus, and architecture (decoder-only,
encoder--decoder). The reference configuration, a decoder-only model
trained on PubChem SMILES with atom-level tokenisation, spans the full size range, and each remaining axis is varied one at a time at
matched pretraining compute. Preprocessing and the full experimental grid are detailed in~\cref{app:data_design}.

\subsection{Pretraining and Scaling Analysis}
\label{sec:scaling_protocol}

We introduce NovoMolGen, a family of 42 CLMs pretrained from scratch across 0.5M--1B parameters, using Qwen3~\citep{yangQwen3TechnicalReport2025} (decoder-only) backbone with next-token prediction and T5Gemma~\citep{zhangEncoderDecoderGemmaImproving2025} (encoder--decoder) backbone with prefix language modelling.
All models use AdamW with a warmup--stable--decay (WSD) schedule
and token-aware batching of approximately 25,000 non-padding
tokens per optimiser step.
Because different molecular representations and tokenisers produce different numbers of tokens, per-token losses are not directly comparable. We therefore report validation set negative log-likelihood in bits per molecule as \(L_{\mathrm{mol}}=-\frac{1}{M\ln 2}\sum_{i=1}^{M}\sum_{t=1}^{T_i}\ln p_{\theta}(x_{i,t}\mid x_{i,<t})\), where \(M\) is the number of molecules, \(T_i\) is the sequence length including EOS, \(x_{i,t}\) is the \(t\)-th token, and \(p_\theta\) is the model probability. For T5Gemma, the encoder
receives only a beginning-of-sequence token, so the decoder scores the
whole molecule autoregressively and $L_{\mathrm{mol}}$ stays comparable
to decoder-only models. We estimate non-embedding compute as $C \approx
6ND$, with $N$ the non-embedding parameter count and $D$ the cumulative
non-padding training tokens, and use $C \approx 3ND$
for T5Gemma, whose scored tokens pass through the decoder, which
holds half the non-embedding parameters. To locate the compute-optimal frontier, we build IsoFLOP profiles that
compare model sizes at a fixed compute budget. The loss-minimising size at
each budget gives $N_{\mathrm{opt}}(C)$ and $D_{\mathrm{opt}}(C)$.
Model configurations and
fitting procedures are detailed in~\cref{app:pretraining_config}.

\subsection{Probing Internal Chemical Representations}
\label{sec:mechanistic_probing}

For each frozen checkpoint, we mean-pool residual-stream activations of
the active set molecules over non-padding positions at early, middle, and
late layers. Linear ridge-regression
probes~\citep{belinkov2022probing} predict 36 properties, grouped into
physicochemical descriptors (e.g.\ molecular weight, logP), PMO
objectives and docking scores, and a logistic probe measures chemical
syntax by separating valid from corrupted sequences. Probes use a fixed
80/20 split with regularisation from five-fold cross-validation, and we
report validation set $R^2$ and accuracy. We
compute each probe's improvement over the randomly initialised network
as a fraction of its final improvement relative to training tokens to compare the model's learning pace. To characterise individual features, we train BatchTopK sparse
autoencoders (SAEs) on the token-level activations of the same
molecules~\citep{bussmann2024batchtopk,cunningham2023sparse}. We
measure reconstruction quality, sparsity and monosemanticity, the mean
over live features of the largest absolute property correlation, and
compare linear probes on SAE features with probes on residual-stream
neurons for detecting the ten most common functional groups, each
trained on the training split and scored on the validation split (F1).
Further details on the probing
setup, SAE training and extended analyses are provided
in~\cref{app:mech_probing_appendix}.

\subsection{Goal-Directed Fine-Tuning and Evaluation}
\label{sec:finetuning}

Finally, we test whether pretraining gains translate into goal-directed
design under limited oracle calls. We fine-tune each checkpoint with the
REINVENT framework~\citep{olivecrona_molecular_2017} using Augmented
Hill-Climb~\citep{thomas2022augmented}, a well-established reinforcement learning (RL) approach shown to perform competitively against more complex policy optimisers like PPO~\citep{bou2024acegen, schulman2017proximal}. The checkpoint initialises a
fixed prior $P_{\mathrm{prior}}(x)$ and a trainable agent
$P_{\mathrm{agent}}(x)$ that learns to generate molecules $x$ with high
task-specific reward $s(x)$. For the highest-scoring molecules in each
batch, the agent's log-likelihood is regressed towards the augmented
target $\log P_{\mathrm{prior}}(x) + \sigma s(x)$, and an
experience-replay buffer reintroduces the best molecules found. We evaluate on 22 PMO tasks (excluding Valsartan SMARTS) with 10,000
oracle calls per task and five protein--ligand docking targets with
3,000 calls per target. PMO performance is the budget-normalised area
under the top-10 reward curve (AUC top-10), capturing optimisation
quality and sample efficiency. For docking, we report the mean docking
score of the top 5\% of novel candidates. Reward definitions,
hyperparameter selection and oracle scoring are
provided in~\cref{app:finetuning_hyperparams}.

\begin{figure}[t]
  \centering
  \includegraphics[width=\linewidth]{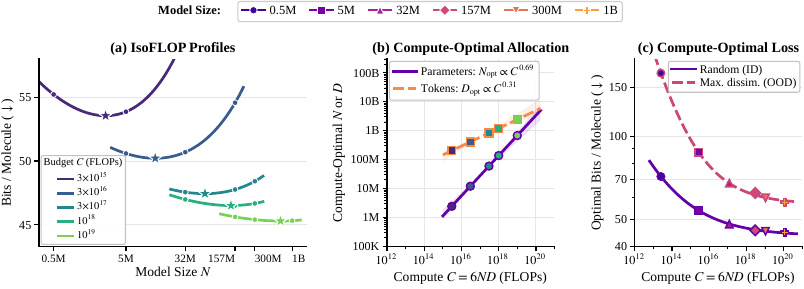}
  \caption{\textbf{Chemical sequence modelling follows compute-optimal
 scaling.} Validation set bits per molecule ($\downarrow$) of the
 PubChem--SMILES--atomwise decoder-only models. \textbf{(a)}~IsoFLOP
 profiles at the five compute budgets of our IsoFLOP grid,
 evaluated at each budget's bracketing model sizes; curves and points
 come from a parametric loss surface fit to all training
 trajectories, and stars mark the loss-minimising size.
    \textbf{(b)}~Compute-optimal parameters and training tokens follow
 power laws in compute, with shaded bootstrap intervals.
    \textbf{(c)}~Compute-optimal loss on the random (in-distribution) and
 maximum-dissimilarity (out-of-distribution) splits, with colour and
 marker denoting model size.}
  \label{fig:isoflop}
\end{figure}

\section{Results}
\label{sec:pretraining_scaling}

We first establish how validation set
loss scales with model size, and compute and locate the compute-optimal
allocation of parameters and tokens (\cref{sec:loss_scaling}). We then
ask which design choices change the loss, comparing representations,
tokenisers, datasets and architectures at matched compute
(\cref{sec:design_choices}), and whether lower loss transfers to
out-of-distribution splits and unconstrained sampling
(\cref{sec:unconstrained_generation}). We next show how chemical
information becomes linearly accessible inside the model as training
proceeds (\cref{sec:representation_dynamics}). Finally, we test whether
any of these pretraining gains transfer to goal-directed design under fixed
oracle budgets (\cref{sec:goal_directed_gen}).

\subsection{Chemical Sequence Modelling Follows Compute-Optimal Scaling}
\label{sec:loss_scaling}

For the PubChem--SMILES--atomwise reference configuration, validation set
loss decreases steadily with pretraining compute across model sizes
from 0.5M to 1B parameters (\cref{fig:teaser_figure}d). Trajectories of
different sizes overlap, so a larger model does not necessarily reach a
lower loss at the same budget. IsoFLOP profiles
capture the trade-off between model capacity and the number of training tokens. 
At a fixed budget, a larger model processes fewer training tokens, while a smaller model sees more tokens but has less capacity. Balancing these competing effects yields a U-shaped profile, whose minimum marks the compute-optimal model size (\cref{fig:isoflop}a).
We fit a parametric loss surface to the reference losses and summarise the loss-minimising allocation as \(N_{\mathrm{opt}}(C)\propto C^{a}\) and \(D_{\mathrm{opt}}(C)=\frac{C}{6N_{\mathrm{opt}}(C)}\propto C^{1-a}\), where \(N_{\mathrm{opt}}\) is the optimal model size, \(D_{\mathrm{opt}}\) is the corresponding data size, and \(a\) is the fitted scaling exponent. The fitted allocations follow an approximate power law, with
$N_{\mathrm{opt}} \propto C^{0.69}$ and $D_{\mathrm{opt}} \propto
C^{0.31}$
for models from 0.5M to 1B (\cref{fig:isoflop}b), so parameters grow
faster than tokens, close to the natural-language allocation ($N_{\mathrm{opt}}\propto C^{0.73}$, $D_{\mathrm{opt}}\propto C^{0.27}$)
of~\citet{kaplan_scaling_2020}. Our grid spans budgets from
$3\times10^{15}$ to $10^{19}$ FLOPs, where the compute-optimal size
ranges from $\sim$2.5M to $\sim$700M parameters. Training to 20 tokens
per parameter reaches within 1\% of the fitted optimal loss at every
budget, so we train and compare all sizes at $D = 20N$ and call these
checkpoints compute-optimal. The compute-optimal loss then follows a
saturating power law on both the random and maximum-dissimilarity splits
(\cref{fig:isoflop}c). On the random split, it falls from 71.5 bits per
molecule at 0.5M to 48.1 at 32M and 45.0 at 1B. Returns diminish at
larger budgets but never vanish, and the out-of-distribution loss
follows the same trend, its ratio to the in-distribution loss is narrowing
from 2.4 at 0.5M to 1.3 at 1B. The IsoFLOP grid and the loss fit are detailed in~\cref{app:isoflop_fits}.

\begin{finding}
  \take Validation set loss follows power laws in compute for chemical language with
  $N_{\mathrm{opt}}\propto C^{0.69}$ and $D_{\mathrm{opt}}\propto C^{0.31}$,
  similar to the allocation for natural language reported by~\citet{kaplan_scaling_2020}.
\end{finding}

\begin{figure}[t]
  \centering
  \includegraphics[width=\linewidth]{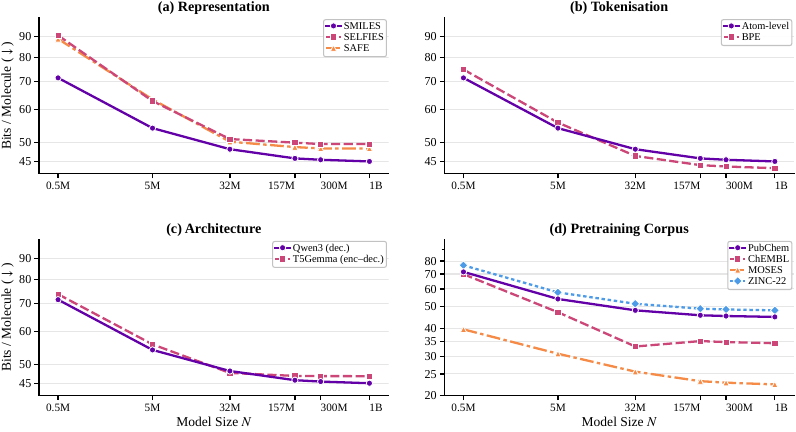}
  \caption{\textbf{Representation and corpus change the loss far more
 than tokenisation or architecture.} Validation set bits per molecule
 ($\downarrow$) at $D = 20N$, varying one design axis at a time
 around the PubChem--SMILES--atomwise decoder-only reference.
    \textbf{(a)}~SMILES is lowest at every scale. \textbf{(b)}~Byte-pair
 tokenisation is slightly worse than atom-level tokenisation at small
 scale and lower from 32M onwards.
    \textbf{(c)}~T5Gemma matches the decoder-only model at 32M and trails
 it at larger scales. \textbf{(d)}~Corpora are scored on their own
 validation splits, so their ordering reflects corpus diversity.}
  \label{fig:design_axes}
\end{figure}

\begin{figure}[t]
  \centering
  \includegraphics[width=\linewidth]{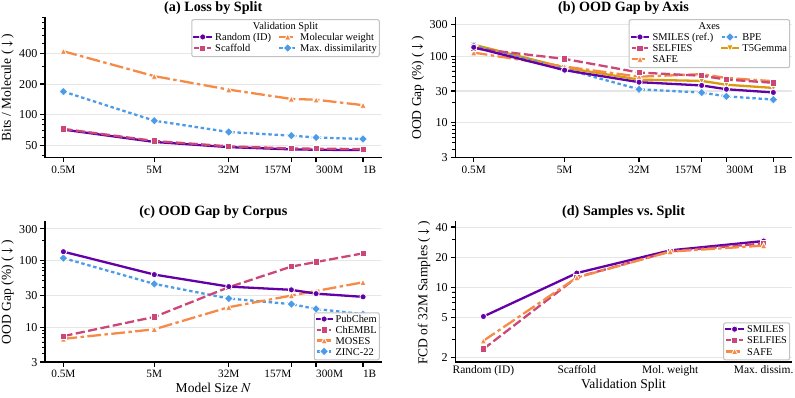}
  \caption{\textbf{Out-Of-Distribution (OOD) generalisation improves with
 scale but depends on representation and corpus.} The
 OOD gap is the relative increase in bits per
 molecule from the random to the maximum-dissimilarity split ($\downarrow$). \textbf{(a)}~Validation set loss of the reference
 configuration on each split. \textbf{(b)}~OOD gap for each design
 axis. \textbf{(c)}~The OOD gap for each corpus,
 measured on its own splits, narrows with scale on PubChem and ZINC-22 but
 widens on MOSES and ChEMBL.
    \textbf{(d)}~FCD between 30,000 molecules sampled from each 32M model
 and each split.}
  \label{fig:id_ood}
\end{figure}

\subsection{Molecular Representation and Corpus Change the Pretraining Loss}
\label{sec:design_choices}

We vary representation, tokenisation, architecture and corpus at
matched model size and training tokens (\cref{fig:design_axes}).
Representation matters most. SMILES reaches lower bits per molecule
than SAFE and SELFIES at every scale from 0.5M to 1B, with gaps of 2.1
and 2.8 bits per molecule at 32M that persist at larger scales
(\cref{fig:design_axes}a). SELFIES tokens are individually easier to
predict and give the lowest bits per token, but a SELFIES molecule
needs about 155 tokens against 47 for SMILES, so its per-molecule loss
stays higher. Tokenisation has a smaller, scale-dependent effect.
Byte-pair encoding is slightly worse than atom-level at 5M (55.8 versus
54.1) and better from 32M onwards (46.4 versus 48.1), as larger models
exploit the larger vocabulary (\cref{fig:design_axes}b). Qwen3 and T5Gemma reach 48.1 and 47.6 bits
per molecule at 32M, and T5Gemma then trails the decoder-only model by
1.1--1.8 bits from 157M to 1B (\cref{fig:design_axes}c), showing that model architecture does not affect pretraining loss. The corpus
changes the loss most, from 25.6 bits per molecule on MOSES to 33.2 on
ChEMBL, 48.1 on PubChem and 51.5 on ZINC-22 at 32M
(\cref{fig:design_axes}d). The ordering follows corpus diversity, from
narrow, drug-like MOSES to the broad make-on-demand space of ZINC-22,
and not sequence length, since ChEMBL molecules are longer than PubChem
molecules yet score lower. \Cref{app:bits_per_token} breaks these losses into bits per token
and tokens per molecule.

\begin{finding}
 \take Molecular representation and corpus have the largest effect improving pretraining loss, whereas tokenisation and architecture change it by only a few per cent. Per-molecule loss is a better metric than per-token loss for CLM model ranking.
\end{finding}

\subsection{Studies on Scaling Models and OOD Generalization}
\label{sec:unconstrained_generation}

The four evaluation splits analyse increasingly strong OOD shifts. The scaffold split behaves almost as in-distribution, at
$\sim$2\% higher bits per molecule than the random split at every scale,
while the molecular-weight and maximum-dissimilarity splits are
genuinely out of distribution, at 5.9$\times$ and 2.4$\times$ the
random-split loss at 0.5M and 2.8$\times$ and 1.3$\times$ at 1B
(\cref{fig:id_ood}a). The gap also depends on the design axes and
corpus. At 32M it is the smallest for byte-pair SMILES (+32\%), then
atom-level SMILES (+41\%), and largest for SELFIES (+57\%)
(\cref{fig:id_ood}b). The gap reduces with scale on PubChem and ZINC-22 corpus, but widens on the smaller MOSES and ChEMBL, which suggests a closer alignment to their distributions
(\cref{fig:id_ood}c). For unconstrained generation, we sample 30,000
molecules per model at temperature $T=1.0$ without top-$k$ or top-$p$
filtering and measuring their Fr\'echet ChemNet Distance (FCD)~\citep{Preuer2018FrechetDiscovery} to each
split, averaging over pretraining seeds at 32M. On the smaller MOSES
and ChEMBL corpora, FCD falls with loss, from 1.82 to 0.14 and from 3.13
to 0.22 between 0.5M and 32M. On PubChem, 32M SMILES
has the lowest bits per molecule but an FCD of 5.1, against 2.4 for
SELFIES and 2.9 for SAFE, and within PubChem SMILES, the FCD rises from 2.8 to 5.1 between 5M and 32M, even as the loss falls from 54.1 to 48.1.
Sampled molecules show alignment to the training distribution, their
FCD rising from 2.4--5.1 against the random split to 26--29 against the
maximum-dissimilarity split for every representation (\cref{fig:id_ood}d).
Better loss on OOD molecules does not influence the sampling process, so split loss alone cannot effectively select a better CLM. This
motivates testing whether scaling pretraining supports goal-directed
optimisation. Full unconstrained-generation
metrics are reported in~\cref{app:ood_generation}.

\begin{finding}
  \take Scaling CLMs improve the OOD split losses, but does not lead to better generalization in unconstrained molecular generation.
\end{finding}

\begin{figure}[t]
  \centering
  \includegraphics[width=\linewidth]{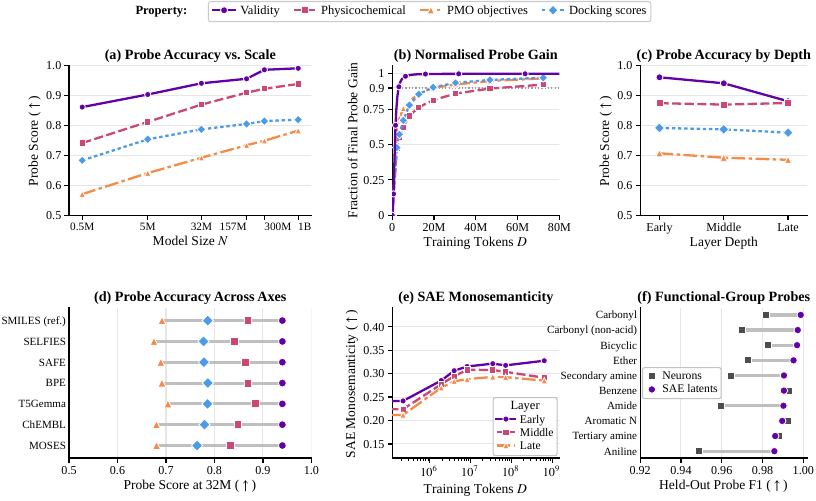}
  \caption{\textbf{Chemical syntax is learned rapidly, while semantic
    properties emerge later and more slowly.} Linear probes and SAEs on
    the active set activations. Validity is the accuracy of separating valid
    molecules from corrupted sequences, and the other properties report
    probe $R^2$. \textbf{(a)}~Middle-layer probe score from 0.5M to 1B
    parameters. \textbf{(b)}~Probe gain over the randomly initialised
    network as a fraction of the gain at the final checkpoint (mean over
    layers), against training tokens expressed for the 32M model. \textbf{(c)}~Probe score by depth at 32M.
    \textbf{(d)}~Probe score at 32M across design axes. \textbf{(e)}~SAE
    monosemanticity of the 32M model during training by depth.
    \textbf{(f)}~Validation set F1 of linear probes on SAE latents and on
    residual-stream neurons for ten common functional groups at 32M.}
  \label{fig:pretraining_results}
\end{figure}

\subsection{Learning of Internal Chemical Representations at Scale}
\label{sec:representation_dynamics}

We ask whether chemical information becomes more accessible as
pretraining progresses. Using frozen checkpoints from 0.5M--1B models,
we probe 36 molecular properties and validity on the active set at
early, middle and late layers (\cref{fig:pretraining_results}). Every
property probe improves with scale, rising between 0.5M and 1B from
0.74 to 0.94 $R^2$ for physicochemical descriptors, 0.68 to 0.82 for
docking scores and 0.57 to 0.78 for PMO objectives, while validity is
already 0.86 at 0.5M and reaches 0.99 at 1B
(\cref{fig:pretraining_results}a). The validity probe reaches 90\% of its final gain after
0.1 tokens per parameter, whereas docking and PMO objectives need a median of 0.6 and physicochemical properties need 1.5
(\cref{fig:pretraining_results}b). Linear accessibility also varies with depth. Docking and PMO $R^2$ fall slightly towards
late layers (0.79 to 0.78 and 0.71 to 0.68), while physicochemical
descriptors stay similarly decodable (\cref{fig:pretraining_results}c).
This ordering of higher validity and lower PMO objectives holds for
every design axis at 32M
(\cref{fig:pretraining_results}d). BatchTopK SAEs show stronger feature--property associations as training
progresses. For the 32M model, early-layer monosemanticity rises from
0.24 to 0.33, while middle- and late-layer values peak mid-training and
then decline slightly (\cref{fig:pretraining_results}e), and the
fraction of features above the property-association threshold rises from
0.10 to 0.21. These measures indicate alignment with the evaluated
properties rather than single-concept encoding. For 7 of the 10
common functional groups, a linear probe on SAE latents beats a probe on
residual-stream neurons by up to 0.04 F1
(\cref{fig:pretraining_results}f). Probe
configurations, SAE curves, and
functional-group F1 scores are in~\cref{app:mech_probing_appendix}.

\begin{finding}
  \take CLMs learn what is a valid molecule (syntax) early and learn the molecular properties (semantics) much later during pretraining and keep improving with scale.
\end{finding}

\begin{figure}[t]
  \centering
  \includegraphics[width=\linewidth]{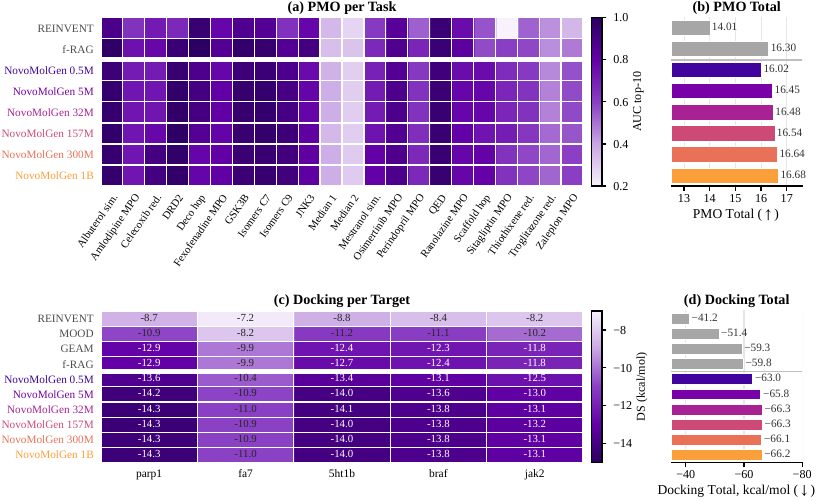}
  \caption{\textbf{NovoMolGen outperforms strong baselines on
      goal-directed generation from 5M parameters, with little gain from
    further scale.} Rows compare baselines (grey) with
    NovoMolGen-SMILES models from 0.5M to 1B parameters.
    \textbf{(a)}~PMO AUC top-10 per task with 10,000 oracle calls
    ($\uparrow$). \textbf{(b)}~Total PMO score over the 22 tasks.
    \textbf{(c)}~Top-5\% docking score of novel molecules per target with
    3,000 oracle calls, in kcal/mol.
    \textbf{(d)}~Total top 5\% docking score over the five targets in kcal/mol
    ($\downarrow$). Baseline results are taken
    from~\citet{lee_molecule_2024}.}
    \label{fig:PMO_model_size}
  \end{figure}

\subsection{Impact of Scaling on Goal-Directed Molecular Generation}
\label{sec:goal_directed_gen}

We evaluate goal-directed generation on two complementary benchmarks,
the Practical Molecular Optimisation (PMO)
benchmark~\citep{gao_sample_2022} and a multi-objective protein--ligand docking benchmark following~\citet{lee_exploring_2023}. For PMO we
follow the original protocol of 10,000 oracle calls per task and
fine-tune each pretrained model with
hyperparameters shared across tasks (\cref{app:finetuning_hyperparams}).
NovoMolGen surpasses REINVENT~\citep{olivecrona_molecular_2017} and \frag~\citep{lee_molecule_2024} from 5M parameters onwards,
scoring 16.45 against 16.30 for \frag (\cref{fig:PMO_model_size}a,b).
Scaling to 1B raises the total only to 16.68, a diminishing and often
negligible gain (\cref{sec:appendix_pmo_extend}). For each docking target (\textit{parp1}, \textit{fa7}, \textit{5ht1b},
\textit{braf}, \textit{jak2}), we generate 3,000 molecules and keep
unique candidates with Tanimoto similarity below 0.4 to the training
set. A novel hit scores below the median of known actives and satisfies
QED $>0.5$ and SA $<5$, the mean
docking score of the top 5\% of novel candidates. Against fragment-based,
3D and diffusion baselines, including GEAM~\citep{lee2024drug},
MOOD~\citep{lee_exploring_2023} and $f$-RAG, NovoMolGen attains better
novel docking scores on every target (\cref{fig:PMO_model_size}c,d). Gains
beyond 5M are marginal, the total docking score moving only from $-65.8$ at
5M to $-66.2$ at 1B. Docking setup, reward curves and top molecules are in~\cref{sec:appendix_docking}.

\begin{finding}

  \take NovoMolGen model with 5M parameters outperforms REINVENT and
  \frag on PMO and docking targets. Scaling NovoMolGen models upto 1B parameters yield marginal improvement in goal-directed molecular generation.
\end{finding}

\section{Discussion and Recommendations}
\label{sec:discussion}

Our analysis, over 30,000 experiments and more than one billion
evaluated molecules, reveals a divergence between pretraining and
downstream performance in CLMs. Validation set loss keeps improving from 0.5M to 1B (\cref{fig:isoflop}), and the same holds across representations,
tokenisers and corpora (\cref{fig:design_axes}) and even on the
out-of-distribution splits (\cref{fig:id_ood}). Probe $R^2$ for PMO objectives and SAE monosemanticity rises between 0.5M and 1B parameter models (\cref{fig:pretraining_results}). PMO and docking scores saturate
at $\sim$5M for every
representation, tokeniser and architecture, where a compute-optimally
trained 5M model already surpasses REINVENT and $f$-RAG and matches a 1B
model. Despite a significant decrease in pretraining loss (9 bits per molecule) between 5M and 1B parameter models, the total PMO score increases only from 16.45 to 16.68 (\cref{fig:PMO_model_size}). Similarly, docking scores lie within 0.2~kcal/mol across model sizes on every target (\cref{fig:PMO_model_size}). Six of the 22 PMO tasks
already reach AUC top-10 $\geq 0.9$ at 32M. Median~1 and Median~2 scores stay below 0.4 for every model and baseline. At 10,000 and 3,000 oracle calls, these benchmarks mainly measure how efficiently fine-tuning exploits the
oracle and only needs a small pretrained prior proposing valid and
diverse molecules. These results indicate that this saturation is not a
limitation of the
transformer architecture, but a consequence of the current pretraining
data, modelling objectives and benchmarks. Unlike genomic or proteomic
data, where
evolutionary pressure has embedded a rich, functional learning signal,
chemical libraries such as PubChem and ZINC-22 are curated for coverage
and synthesisability, so their dominant signal teaches chemical syntax
rather than functional semantics. Similar limits of scale-driven
approaches have been reported for proteins~\citep{fournier2024protein}
and DNA~\citep{boshar2025foundational}. Further progress will require
objectives beyond chemical validity, for example, signals from
protein--ligand interactions, physicochemical properties or experimental
bioactivity combined with self-supervision, or reinforcement learning
introduced earlier in training as a ``fitness'' signal analogous to
natural selection. Our results suggest several practical guidelines
for building and evaluating next-generation CLMs:
\begin{itemize}[leftmargin=14pt,noitemsep,topsep=2pt]
  \item \textbf{Train better sample-efficient algorithms, not larger
 models.} A compute-optimal 5M model uses $\sim$$4\times10^{4}$ fewer
 pretraining FLOPs than a 1B model, and makes each policy update
    $\sim$200 times cheaper, at comparable PMO and docking performance.
  \item \textbf{SMILES remains the best representation.} SMILES with a
 decoder-only model gives the lowest loss and the highest downstream
 scores, while tokenisation and architecture change the loss by only
 a few per cent.
  \item \textbf{Report bits per molecule and test on OOD splits.}
 Per-token loss favours long representations such as SELFIES, and the
 scaffold split is effectively in-distribution, so molecular-weight
 and maximum-dissimilarity splits are more informative.
  \item \textbf{Better benchmarks.} Harder multi-objective optimisation
 tasks, targets absent from public training data and stricter oracle
 budgets would provide a better test of whether larger models help.
\end{itemize}

\section{Conclusions}
\label{sec:conclusion}
We presented a compute-controlled study of CLMs from 0.5M to 1B
parameters across representation, tokenisation, corpus and architecture.
Pretraining loss follows compute-optimal scaling, and internal
representations are enriched in two phases, chemical syntax first and chemical semantics next. Yet, goal-directed generation
saturates at $\sim$5M parameters, where NovoMolGen already surpasses
REINVENT and $f$-RAG and matches a model two orders of magnitude larger. Although NLP-inspired assumptions that larger models and longer
training hold for pretraining and chemical representations, current benchmarks cannot capture the complexity of chemistry. This motivates pretraining modelling objectives that enable CLMs to learn chemical semantics directly. NovoMolGen can also be extended to other tasks, such as text-instructed design and fragment-constrained generation.

\subsection*{AI use statement}

In this work, we used generative AI tools to polish
the writing, improve clarity and readability and assist with the retrieval and discovery of related literature. We have
reviewed all AI-assisted work and revised the text to reflect
our intended meaning and scientific claims. We take
responsibility for the final content of this work, including
text, claims, and artefacts produced with the aid of
generative AI.

\subsection*{Reproducibility statement}

\Cref{app:data_design} describes the datasets, standardisation, leakage
filtering and evaluation splits. \Cref{app:pretraining_config} specifies
the model architectures, optimiser, learning-rate schedule and
token-aware batching. \Cref{app:isoflop_fits} reports the IsoFLOP grid,
the fitted loss surface and the bootstrap procedure, and
\cref{app:mech_probing_appendix} describes the probing and
sparse-autoencoder protocols. \Cref{app:finetuning_hyperparams} details
the Augmented Hill-Climb objective, reward definitions, oracle budgets
and hyperparameter search. We will release the NovoMolGen model suite,
intermediate pretraining checkpoints, sparse-autoencoder dictionaries,
and the code for pretraining, probing and downstream optimisation to reproduce every reported result.

\bibliography{iclr2027_conference}
\bibliographystyle{iclr2027_conference}

\newpage
\appendix
\crefalias{section}{appendix}
\crefalias{subsection}{subappendix}
\crefalias{subsubsection}{subsubappendix}

\counterwithin{figure}{section}
\counterwithin{table}{section}
\setcounter{figure}{0}
\setcounter{table}{0}
\renewcommand\thefigure{\thesection\arabic{figure}}
\renewcommand\thetable{\thesection\arabic{table}}

\section{Related Work}

\subsection{Representation of Molecules}
Molecules are commonly depicted using structural diagrams,
traditionally drawn with pen and paper, to represent bonds and atoms
visually. However, in chem-informatics, more advanced representations
are needed for the computational processing of molecular structures.
In this context, "molecular representations" encompass any encoding
of a chemical compound that can be employed for computational
exploration of the chemical space. The current approaches to
representing molecules can be broadly classified into four types: (i)
Vector-based representations, (ii) Graph-based representations, (iii)
3D-based representations, and (iv) String-based representations.

\textbf{Vector-based}: Topological fingerprints, such as Extended
Connectivity FingerPrints
(ECFP)~\citep{rogers_extended-connectivity_2010} and Molecular ACCess
System (MACCS)~\citep{durant_reoptimization_2002}, have traditionally
been employed for substructure and molecule similarity searches.
These fingerprints encode molecules as a sequence of bits in an
identifier list, each denoting the presence or absence of a specific
substructure. Although each molecular structure can be
deterministically mapped to a fingerprint, the fingerprints are only
partially invertible~\citep{le_neuraldecipher_2020}, which limits
their applicability in \denovo molecule
generation~\citep{gomez-bombarelli_automatic_2018,
tazhigulov_molecular_2022}. Additionally, the fingerprints can be
augmented with 2D molecular descriptors such as Molecular Weight, QED
score, and Number of Aromatic Rings to help impose specific
constraints on the generated molecules and make them more aligned
with desired chemical properties or biological activity.

\textbf{Graph-based}: A 2D molecular graph is defined as $\gG =
(\gV,\gE)$ where $\gV$ is the set of nodes (atoms) and $\gE$ is the
set of edges (bonds). The type of atoms and edges can be represented
using a feature matrix $\mX$. Graph Neural Networks (GNNs) have been
used to learn the representations of
molecules~\citep{kipf_semi-supervised_2017, xu_how_2018,
xiong_graph_2021} for tasks such as reaction prediction, property
prediction and drug discovery. The initial frameworks for learning
molecule representation used Message Passing Neural Networks (MPNNs)
to compute the atom embedding based on neighbourhood information
capturing local interaction effects~\citep{gilmer_neural_2017}.
Although many variants of GNNs have been
proposed~\citep{morris_weisfeiler_2019, maron_provably_2019,
  zhang_heterogeneous_2019, brockschmidt_gnn-film_2020,
xiong_pushing_2020, li_chemical_2024}, challenges remain in terms of
higher-order expressivity, scalability, and computational cost.

\textbf{3D-based}: While using Graph Neural Networks (GNNs) on 2D
molecular graphs is convenient and seem to be the obvious choice, the
resulting representations often overlook crucial spatial information,
such as the spatial direction and torsion angles between
atoms~\citep{guo_graph-based_2023}. Recent advancements in molecule
representation learning have focused on integrating 3D coordinate
information into 2D molecular graphs~\citep{luo_3d_2024,
fang_geometry-enhanced_2022, li_geomgcl_2022}.
Uni-Mol~\citep{zhou_uni-mol_2022} introduced a pretraining framework
capable of directly utilizing 3D positions as inputs and outputs.
However, a significant challenge in this approach is the existence of
multiple low-energy conformations for a given molecule. These
conformations are not easily accessible and are particularly
difficult to compute, especially for large molecules and the vast
chemical space, which spans billions of possible molecules.

\textbf{String-based}: 2D molecular structures can also be encoded as
linear notations, which use specialized languages to represent
molecular structures and compositions in chemistry. The earliest
example of such a molecular language was developed in the 1980s by
\citet{weininger_smiles_1988}. The SMILES (Simplified Molecular Input
Line Entry System) notation encodes atoms, bonds, and connectivity
patterns using ASCII strings, where atoms are represented by
characters (e.g., `C' for carbon, `N' for nitrogen) and bonds by
special characters (e.g., `-' for a single bond, `=' for a double
bond, `\#' for a triple bond). However, the syntax rules and
restrictive grammar of SMILES can result in many invalid molecules
during parsing, even when the string appears to represent a plausible
molecular structure. To address some of these limitations,
DeepSMILES~\citep{oboyle_deepsmiles_2018} was introduced, which
avoids the issue of unbalanced parentheses by using only closing
parentheses, where the number of parentheses indicates the branch
length. More recently, SELFIES (Self-Referencing Embedded
Strings)~\citep{krenn_selfies_2022} was developed as a linear
notation that is 100\% robust; every SELFIES string corresponds to a
valid molecule, even for entirely random strings. Additionally, SAFE
(Sequential Attachment-based Fragment
Embedding)~\citep{noutahi_gotta_2024} introduced a framework for
fragment-constrained molecule generation tasks while maintaining
compatibility with existing SMILES parsers.

String-based molecular representations offer a computationally
efficient and scalable approach to exploring the vast chemical space
using unlabelled data without relying on additional information such
as 3D geometry or complex optimization techniques. Despite their
simplicity, these methods capture essential chemical information and
structural features, making them valuable tools in computational
chemistry and drug discovery.

\subsection{Deep Generative Models for \textit{De Novo} Molecule Generation}
Deep generative models have become a key approach for \denovo
molecule generation, facilitating the discovery of novel compounds by
capturing complex patterns within the vast chemical space. Numerous
methods have emerged, each focused on different molecular
representations and assembly
strategies~\citep{olivecrona_molecular_2017, jin_junction_2018,
  polykovskiy_molecular_2020, jin_hierarchical_2020, bagal_molgpt_2022,
  eckmann_limo_2022, jo_score-based_2022, irwin_chemformer_2022,
  fang_domain-agnostic_2023, lee_exploring_2023, yang_hit_2024,
ross_gp-molformer_2024}.

\textbf{Assembly Methods}: Early
approaches~\citep{gomez-bombarelli_automatic_2018, jin_junction_2018,
winter_efficient_2019, tazhigulov_molecular_2022} primarily utilized
Variational Autoencoders (VAEs) to transform SMILES representations
into a continuous latent space, followed by sampling and decoding to
generate discrete molecular structures. Graph-based generative models
have emerged as a natural extension, directly leveraging the
molecular graph structure where atoms are represented as nodes and
bonds as edges~\citep{cao_molgan_2022}. GraphVAE (Graph Variational
Autoencoder)~\citep{simonovsky_graphvae_2018} encodes and decodes
molecules using edge-conditioned graph convolutions.
MoFlow~\citep{zang_moflow_2020}, a flow-based graph generative model
learns invertible mappings between molecular graphs and their latent
representations. Graph generation approaches utilize small molecular
building blocks such as atoms and their performance degrades
significantly for larger molecules. To tackle this problem recent
works employ significantly larger and more flexible graph motifs as
basic building blocks~\citep{kuznetsov_molgrow_2021}. In parallel, 3D
molecule generation has gained substantial attention, particularly
through the use of diffusion models.
G-Schnet~\citep{gebauer_symmetry-adapted_2019}, for instance,
utilizes an autoregressive process to iteratively sample atoms and
bonds in 3D space. Similarly, inspired by the success of diffusion
models in other domains~\citep{sohl-dickstein_deep_2015,
ho_denoising_2020, kong_autoregressive_2023},
\citet{hoogeboom_equivariant_2022} proposed an equivariant diffusion
model for generating novel 3D molecular structures.

\textbf{Optimization Methods}: Molecule optimization involves
navigating an immense and complex chemical space, which requires
sophisticated algorithms capable of efficiently searching for and
generating molecules with optimal characteristics. Several
computational approaches have been developed to tackle this
challenge, each with distinct mechanisms for exploring the design
space and optimizing molecular properties. Reinforcement Learning
treats molecule optimization as a sequential decision-making problem.
In this context, the state typically represents a partially generated
molecule, and actions correspond to modifications at the graph or
string level. The reward function is based on the properties of the
generated molecules, guiding the model toward desirable outcomes.
Bayesian Optimization operates by learning a continuous latent space
of molecular representations, optimizing target properties by
navigating through this latent space. Genetic algorithms, inspired by
natural evolutionary processes, explore the chemical space through
operations such as mutation and crossover applied to a pool of
candidate molecules, promoting diversity and exploration. Gradient
ascent methods, on the other hand, estimate the gradient of a
molecular property across the chemical space and use backpropagation
to optimize molecular structures. Hill Climbing is an iterative
optimization method with high-performing molecules from previous
rounds incorporated into the training data to refine the generative
model progressively.

While graph-based and 3D deep generative models have made significant
strides in generating molecular structures, recent advances in
natural language processing have opened new possibilities for \denovo
molecule generation and optimization. Large Language Models (LLMs)
offer a novel approach for navigating the large chemical space,
presenting new opportunities for optimizing molecular properties in a
scalable and computationally feasible manner. Combining these models
with traditional optimization techniques can further enhance the
search for \denovo molecules with desired properties, marking a
significant step forward in drug discovery.

\subsection{Language Models in Molecule Generation}
LLMs can effectively model sequential data and have shown remarkable
proficiency in understanding and generating human
language~\citep{dubey2024llama,
jiang_mistral_2023,groeneveld_olmo_2024}. These architectures are now
being repurposed to explore and generate molecular structures. When
treated as sequences of tokens, 1D molecular representations
inherently encode chemical information, including 2D bonding topology
patterns, while LLMs further enhance this by learning to generate
diverse molecular structures. These models leverage unlabelled
molecular data from across the chemical space, offering a novel
approach to \denovo molecule generation and broadening the potential
for chemical discovery. MolGPT~\citep{bagal_molgpt_2022} employs a
decoder-only transformer architecture, inspired by GPT, to predict
SMILES token sequences for molecular generation. Building on these
advancements, models like cMolGPT~\citep{wang_cmolgpt_2023} have been
developed to generate target-specific compounds by incorporating
conditional training for property optimization.
Taiga~\citep{mazuz_molecule_2023} extends this approach by employing
a two-stage framework: first, the model treats molecular generation
as a language modeling task by predicting the next token in SMILES
strings. Subsequently, reinforcement learning (RL) is applied to
optimize simple chemical properties such as QED (Quantitative
Estimate of Drug-likeness) and logP (Octanol-Water Partition
Coefficient). MolGen~\citep{fang_domain-agnostic_2023}, in contrast,
utilizes an encoder-decoder BART architecture, focusing on generating
chemically valid molecules through the SELFIES notation. It
incorporates a chemical feedback mechanism to align generative
probabilities with real-world chemical preferences.
SAFE-GPT~\citep{noutahi_gotta_2024} introduces a new line notation
and trains a GPT-like model on 1.1 billion SAFE notations,
demonstrating versatile and robust performance in both \denovo and
fragment-constrained molecule generation tasks.

Tokenizers, which convert raw text sequences into tokens, are a
critical component of modern language models. In CLMs, token
vocabularies are often constructed using a predefined regular
expression developed by \citet{schwaller_molecular_2019}, which
splits SMILES strings into relevant tokens representing each atom
(e.g., 'C' for carbon, 'N' for nitrogen). Less commonly, subword
tokenization algorithms such as Byte Pair Encoding
(BPE)~\citep{sennrich_neural_2016} or
Unigram~\citep{kudo_subword_2018} are employed, sometimes combined
with atomwise pretokenization and BPE. Given that tokenizer design
impacts every stage of the modeling pipeline, this study explores the
effects of using learned tokenization methods versus hand-crafted approaches.

\section{Dataset Preparation and Design Axes Details}
\label{app:data_design}

We pretrain on four public molecular corpora of increasing size and
diversity. MOSES~\citep{polykovskiy_molecular_2020} contains about 1.9M
drug-like molecules with a restricted atom set.
ChEMBL~\citep{Gaulton2012ChEMBL:Discovery}
contains about 1.9M bioactive molecules drawn from the medicinal-chemistry
literature. PubChem~\citep{Kim2016PubChemDatabases} contributes about 100M
molecules after filtering, and forms the reference corpus for our scaling
sweep. ZINC-22~\citep{tingle_zinc-22free_2023} is the largest library, and
we sample 1.5B molecules proportionally across its heavy-atom-count and
logP tranches so that the sample preserves the coverage of the full
make-on-demand space. \Cref{tab:corpus_stats} summarises the corpora.

\subsection{Standardisation}
Each corpus passes through the same offline
pipeline. We canonicalise every SMILES string with RDKit, drop entries
that RDKit cannot parse, strip salts and normalise charges, and
deduplicate on the canonical string in a single streaming pass so that no
molecule appears twice. We then convert the surviving molecules to
SELFIES and SAFE, and build atom-level and byte-pair
tokenisers for each representation from the same underlying molecules.
This keeps every representation and tokeniser aligned to one shared set of
compounds.

\subsection{Leakage control}
To keep the evaluation targets unseen during
pretraining, we assemble an active set of about 21,000 known actives for
the PMO and docking targets together with the Tartarus benchmark
molecules~\citep{nigam_tartarus_2023}. We compute 2,048-bit ECFP
fingerprints of radius 2 and remove from every corpus any molecule whose
Tanimoto similarity to an active set member exceeds 0.5. The active set is
never used for training and later serves as the validation probe set in
\cref{app:mech_probing_appendix}.

\subsection{Evaluation splits}
From the filtered molecules we hold out
four splits of increasing distribution shift. The random split is a
uniform sample and serves as the in-distribution reference. The scaffold
split holds out entire Bemis-Murcko scaffolds. The molecular-weight split
holds out the heaviest molecules. The maximum-dissimilarity split holds
out the fingerprint clusters that sit farthest from the rest of the
corpus. Each split contains 175,000 molecules and is scored on its own,
so a single model is measured against progressively harder generalisation
rather than a single validation set.

\subsection{Design grid}
The reference configuration is a Qwen3
decoder-only model trained on PubChem SMILES with atom-level tokenisation,
and it spans the full range from 0.5M to 1B parameters. Each remaining
axis is varied one at a time around this reference at matched model size
and training tokens, as summarised in \cref{tab:design_grid}. This
one-axis-at-a-time design keeps
the number of pretraining runs tractable while still isolating the effect
of each design choice on validation set loss, learned representations and
goal-directed generation.

\begin{table}[htb]
  \caption{Pretraining corpora. Token counts use the atom-level SMILES
    tokeniser. Sizes are reported after standardisation and leakage
  filtering.}
  \label{tab:corpus_stats}
  \centering
  \resizebox{0.85\linewidth}{!}{
    \begin{tabular}{lcccl}
      \toprule
      Corpus & Molecules & Median heavy atoms & Median SMILES tokens
      & Character \\
      \midrule
      MOSES    & 1.9M            & 21 & 28 & Narrow, drug-like,
      restricted atoms \\
      ChEMBL   & 1.9M            & 27 & 48 & Bioactive, literature-derived \\
      PubChem  & 100M            & 24 & 47 & Broad drug-like
      (reference corpus) \\
      ZINC-22  & 1.5B (sampled)  & 29 & 52 & Make-on-demand, largest coverage \\
      \bottomrule
  \end{tabular}}
\end{table}

\begin{table}[htb]
  \caption{Experimental design grid. Each axis is varied around the
    PubChem--SMILES--atomwise Qwen3 reference at matched model size and
  training tokens.}
  \label{tab:design_grid}
  \centering
  \resizebox{0.95\linewidth}{!}{
    \begin{tabular}{lll}
      \toprule
      Axis & Variants & Model sizes \\
      \midrule
      Reference          & PubChem, SMILES, atom-level, Qwen3 & 0.5M,
      5M, 32M, 157M, 300M, 1B \\
      Representation     & SELFIES, SAFE                      & 0.5M,
      5M, 32M\\
      Tokenisation       & Byte-pair encoding                 & 0.5M,
      5M, 32M, 157M, 300M, 1B \\
      Architecture       & T5Gemma (encoder--decoder)         & 0.5M,
      5M, 32M, 157M, 300M, 1B \\
      Pretraining corpus & MOSES, ChEMBL, ZINC-22            & 0.5M,
      5M, 32M, 157M, 300M, 1B \\
      \bottomrule
  \end{tabular}}
\end{table}

\subsection{Dataset Diversity}
\label{sec:data_diversity}

\begin{figure}[htb]
  \centering
  \includegraphics[width=\linewidth]{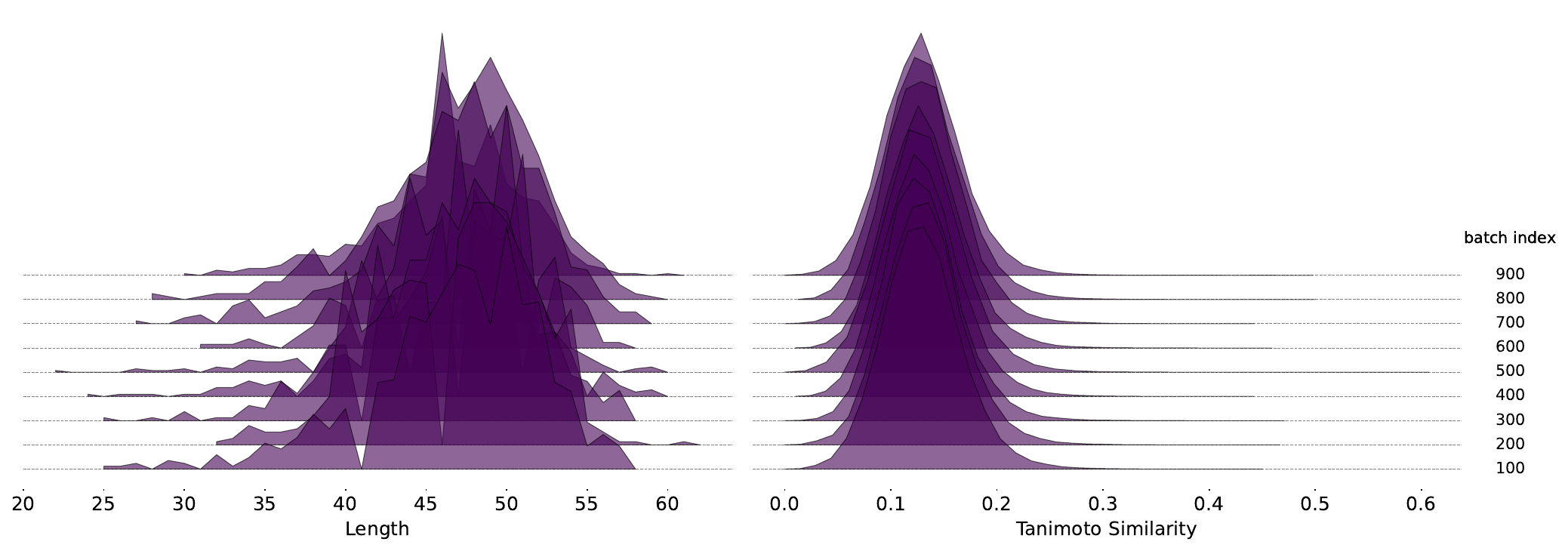}
  \caption{Diversity of the data across batches. The left plot shows
    the distribution of molecular lengths, tokenized using an atomwise
    tokenizer, indicating variation in molecule length within each
    batch. The right plot shows Tanimoto similarity between molecule
  pairs, demonstrating structural diversity in each batch.}
  \label{fig:data_quality}
\end{figure}

To ensure the diversity of the dataset, we confirmed that it contains
a broad range of molecular structures, which is crucial for
generating a wide variety of valid, novel, and unique molecules as
illustrated in~\cref{fig:data_quality}. The left plot shows the
distribution of molecular lengths, tokenized using an atomwise
tokenizer, demonstrating variation in molecule size across batches.
The right plot displays the Tanimoto similarity between molecule
pairs, showcasing the structural diversity in each batch an essential
factor for robust molecular generation.

\begin{figure}[htb]
  \centering
  \includegraphics[width=\linewidth]{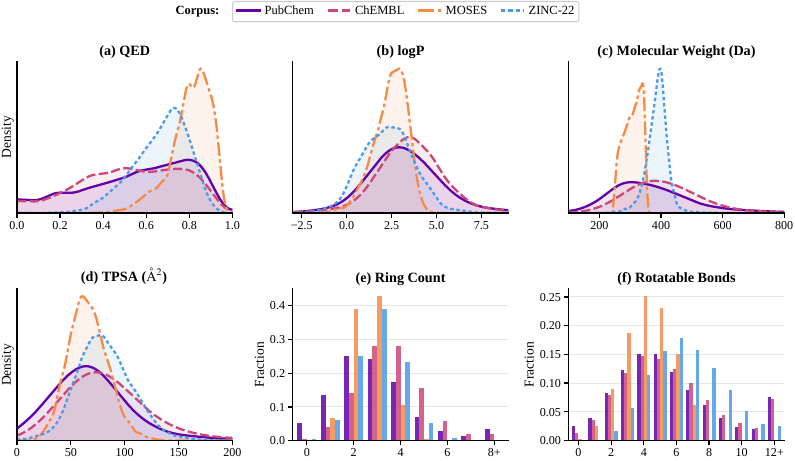}
  \caption{\textbf{Molecular-property distributions of the four
    pretraining corpora.} Each corpus is represented by 6,000 molecules
    sampled uniformly across all of its shards, with properties computed
    by RDKit on the largest fragment. \textbf{(a--d)}~Kernel density
    estimates of QED, logP, molecular weight and TPSA. \textbf{(e,\,f)}~Fraction
    of molecules by ring count and number of rotatable bonds, with the
  last bin including all larger values.}
  \label{fig:dataset_properties}
\end{figure}

The four pretraining corpora also differ in their molecular properties,
which helps explain why they are not equally hard to model.
\Cref{fig:dataset_properties} compares six properties across MOSES,
ChEMBL, PubChem and ZINC-22, each estimated from 6,000 sampled molecules.
MOSES, ChEMBL and PubChem are drug-like and overlap closely across most
properties, with QED concentrated above 0.7 and logP around 2 to 3.
ZINC-22 is the outlier. It is less drug-like, with QED peaking near 0.5
rather than above 0.7, and its molecules are heavier and more tightly
concentrated in molecular weight, with more rings and more rotatable bonds.
This broader make-on-demand chemistry matches the corpus ordering of
validation set loss, where the narrow, drug-like MOSES
distribution is the easiest to model and the broad ZINC-22 space is the
hardest.

\subsection{Property Distributions}
\label{app:property_distributions}

The distribution of properties serves as a valuable tool for visually
evaluating the generated structures. We present a kernel density
estimation of these distributions and calculate the Wasserstein-1
distance to compare the distributions of the generated and reference
datasets. We use the following properties:

\begin{enumerate}
  \item \textbf{Quantitative Estimation of Drug-likeness (QED)}: A
    metric derived from medicinal chemistry principles that
    quantifies the drug-likeness of molecules on a scale from 0 to 1.
  \item \textbf{Synthetic Accessibility (SA)}: A measure of a
    molecule's synthesizability, calculated based on the
    contributions of molecular fragments. The metric ranges from 10
    (difficult to synthesize) to 2 (easily synthesizable).
  \item \textbf{Octanol-Water Partition Coefficient (logP)}:
    Represents the ratio of a compound’s concentration in the octanol
    phase to its concentration in the aqueous phase in a two-phase
    octanol/water system, serving as an indicator of solubility.
  \item \textbf{Molecular Weight (MW)}: Evaluates whether the
    generated set is biased toward heavier or lighter molecules,
    computed as the sum of atomic weights.
  \item \textbf{Topological Polar Surface Area (TPSA)}: Estimated
    based on functional group contributions from a database of
    substructures, this metric reflects lipid solubility and
    molecular polarity. Higher TPSA values indicate reduced
    absorption and distribution within the body.
  \item \textbf{Bertz Complexity}: A graph-theoretical measure that
    quantifies molecular complexity using structural invariants and
    information-theoretic principles.
  \item \textbf{Number of Rings (NumRings) and Rotatable Bonds}:
    Represents the number of independent closed-ring structures and
    rotatable bonds within a molecule, which are essential for
    analyzing molecular topology and are commonly used in
    cheminformatics for compound classification and comparison.
\end{enumerate}

\begin{figure}[!htb]
  \centering

  \begin{subfigure}{0.4\linewidth}
    \centering
    \includegraphics[width=\linewidth]{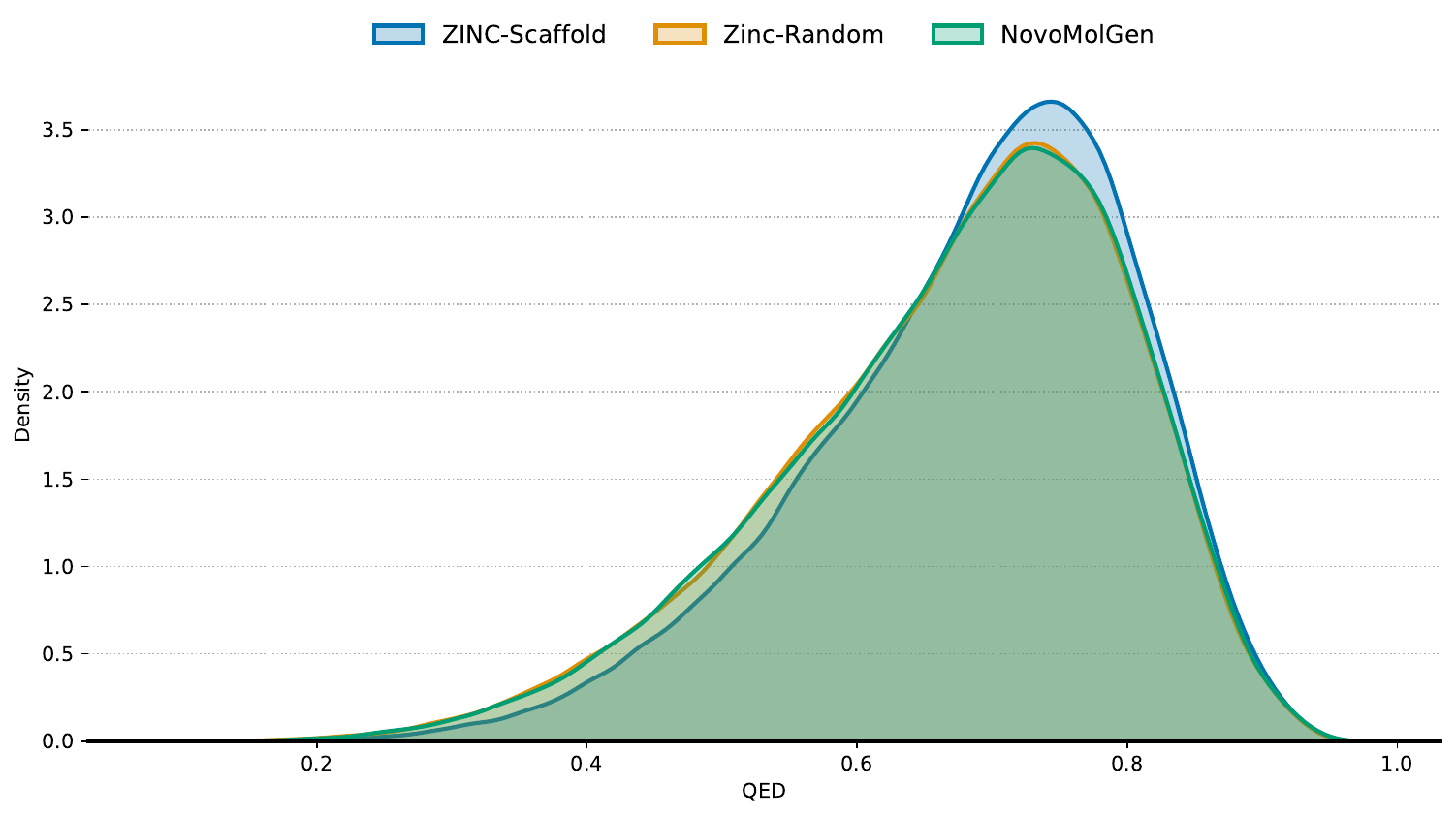}
    \caption{QED Distribution}
    \label{fig:subfig_qed}
  \end{subfigure}
  \hfill
  \begin{subfigure}{0.4\linewidth}
    \centering
    \includegraphics[width=\linewidth]{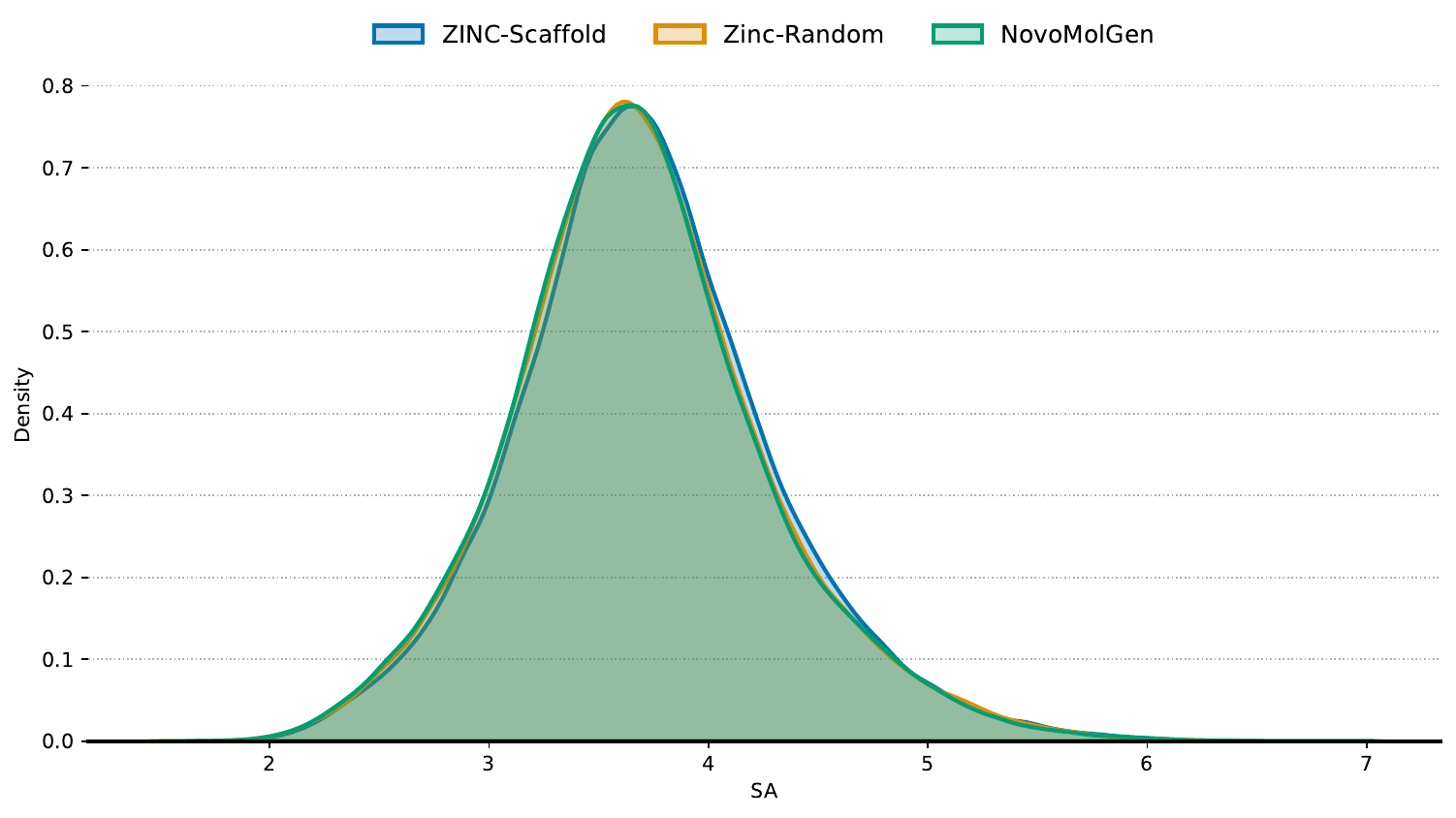}
    \caption{SA Distribution}
    \label{fig:subfig_sa}
  \end{subfigure}

  \vspace{0.5em}

  \begin{subfigure}{0.4\linewidth}
    \centering
    \includegraphics[width=\linewidth]{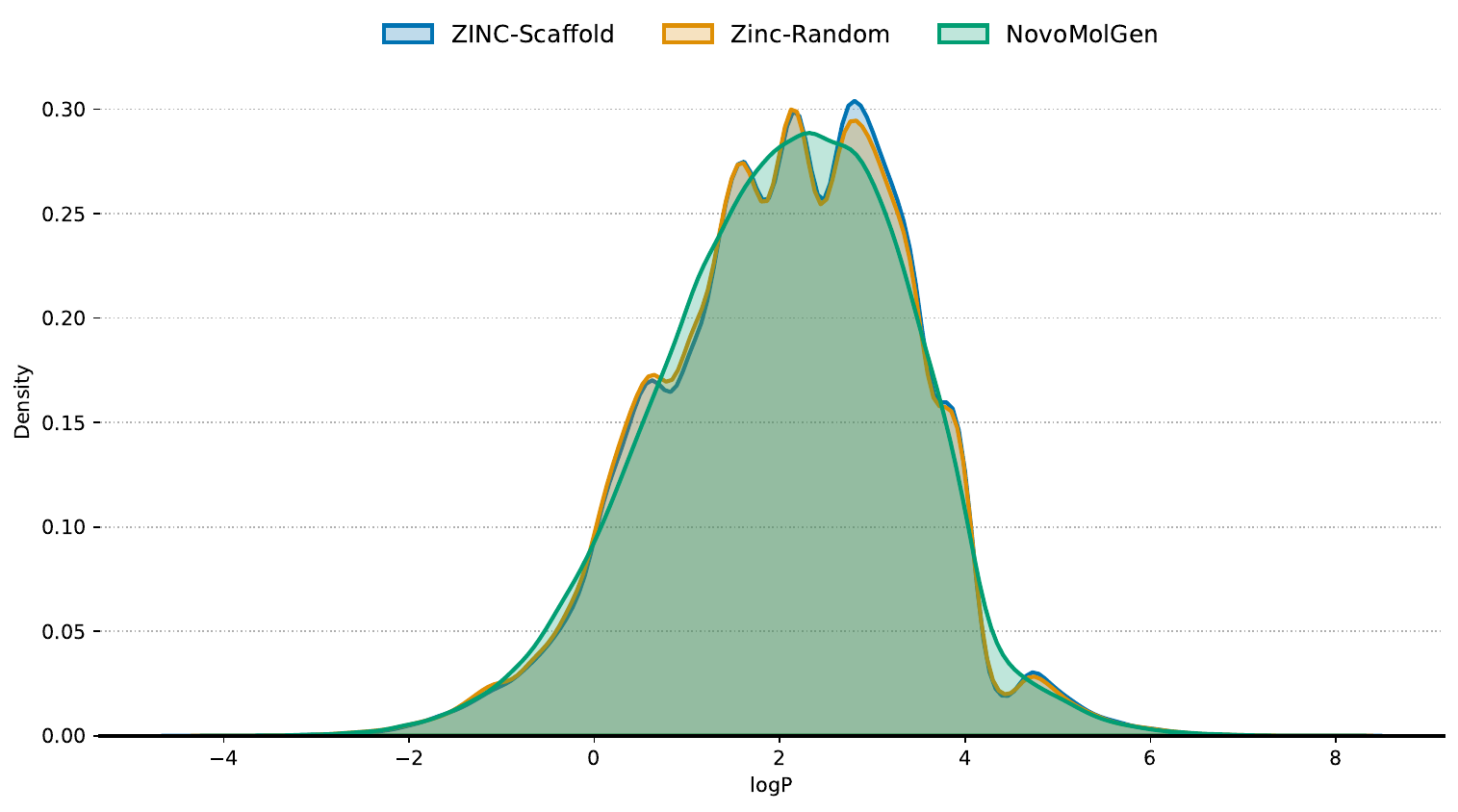}
    \caption{logP Distribution}
    \label{fig:subfig_logp}
  \end{subfigure}
  \hfill
  \begin{subfigure}{0.4\linewidth}
    \centering
    \includegraphics[width=\linewidth]{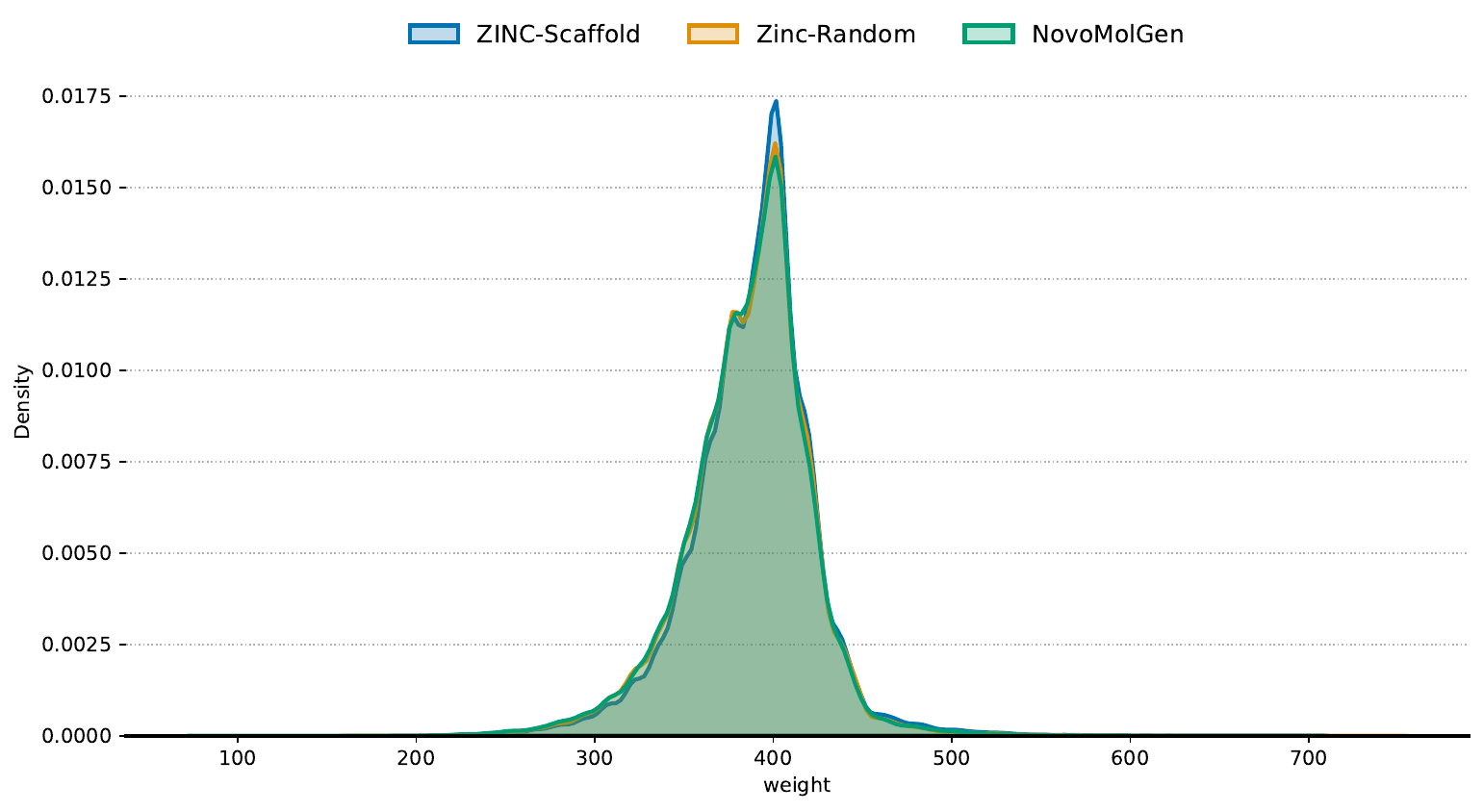}
    \caption{MW Distribution}
    \label{fig:subfig_weight}
  \end{subfigure}

  \vspace{0.5em}

  \begin{subfigure}{0.4\linewidth}
    \centering
    \includegraphics[width=\linewidth]{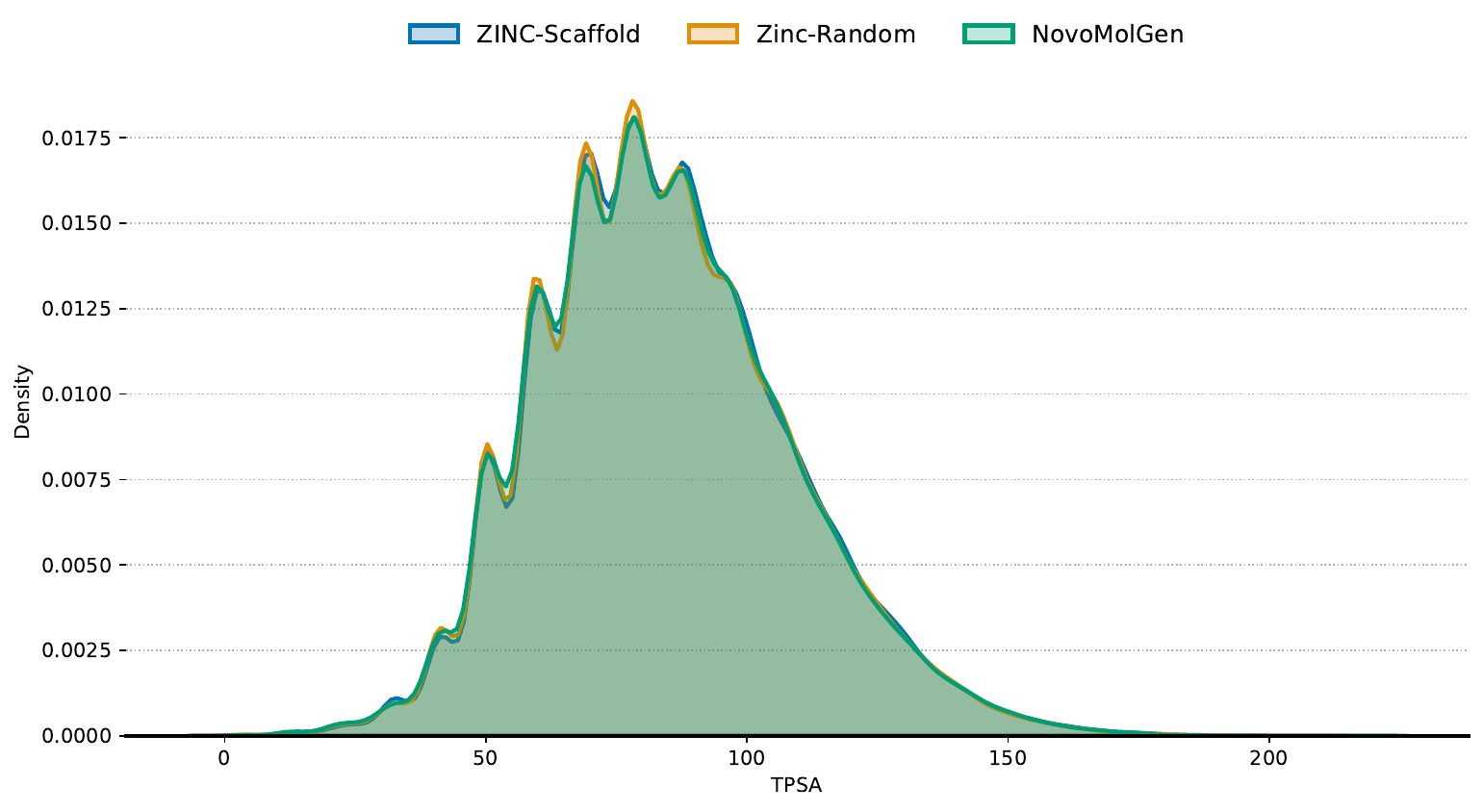}
    \caption{TPSA Distribution}
    \label{fig:subfig_tpsa}
  \end{subfigure}
  \hfill
  \begin{subfigure}{0.4\linewidth}
    \centering
    \includegraphics[width=\linewidth]{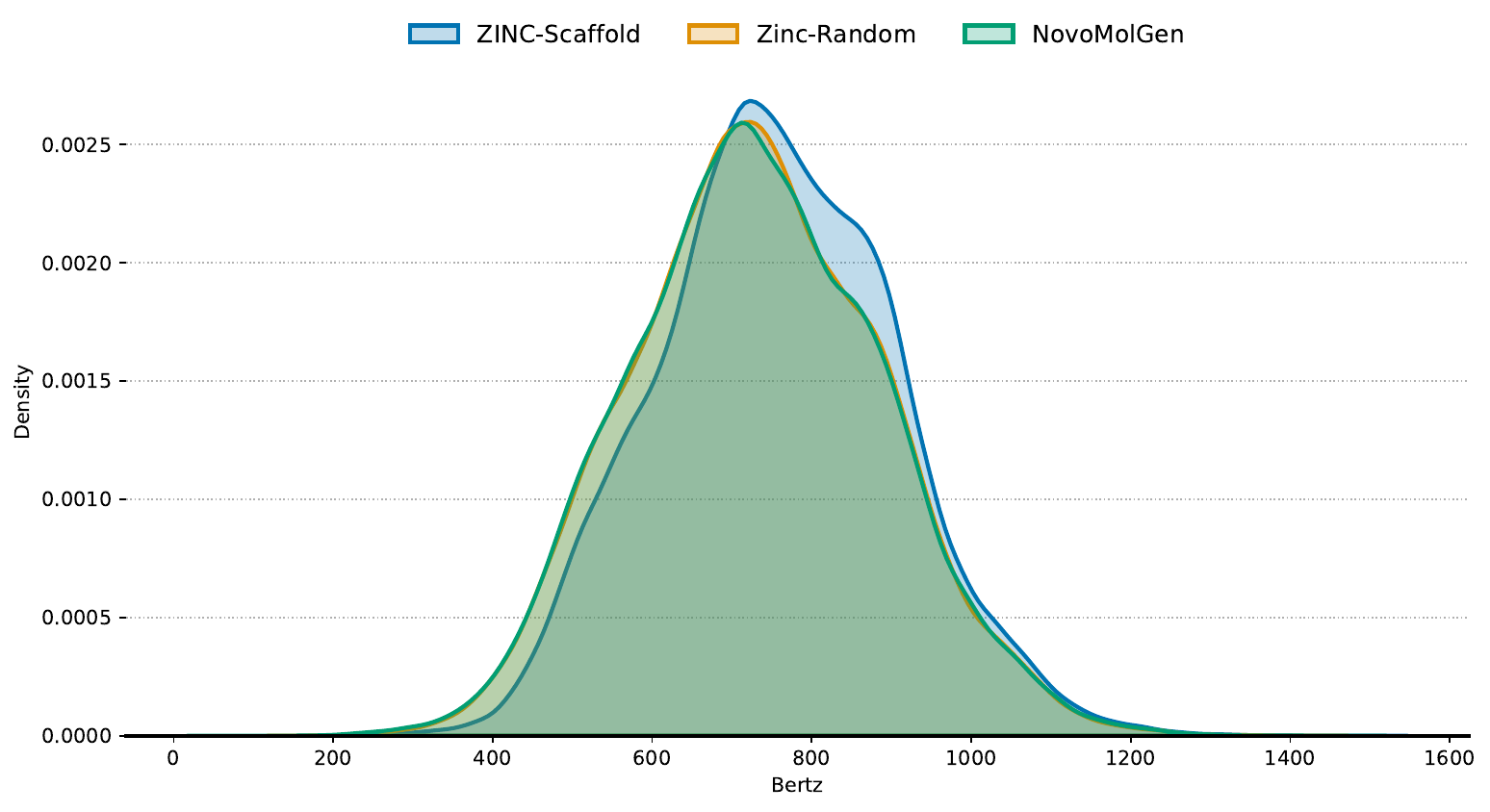}
    \caption{Bertz Complexity Distribution}
    \label{fig:subfig_bertz}
  \end{subfigure}

  \vspace{0.5em}

  \begin{subfigure}{0.4\linewidth}
    \centering
    \includegraphics[width=\linewidth]{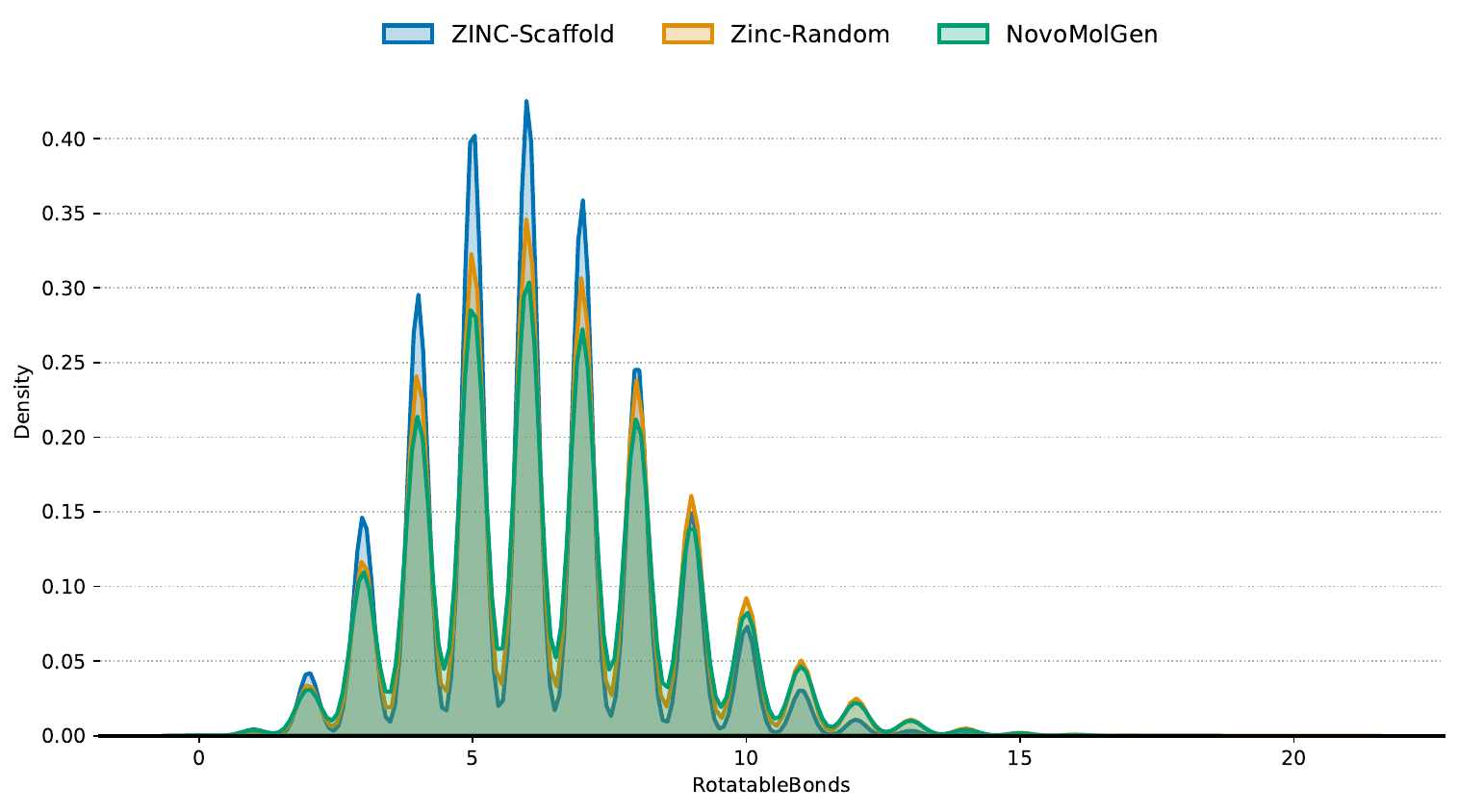}
    \caption{Rotatable Bonds Distribution}
    \label{fig:subfig_rotatable}
  \end{subfigure}
  \hfill
  \begin{subfigure}{0.4\linewidth}
    \centering
    \includegraphics[width=\linewidth]{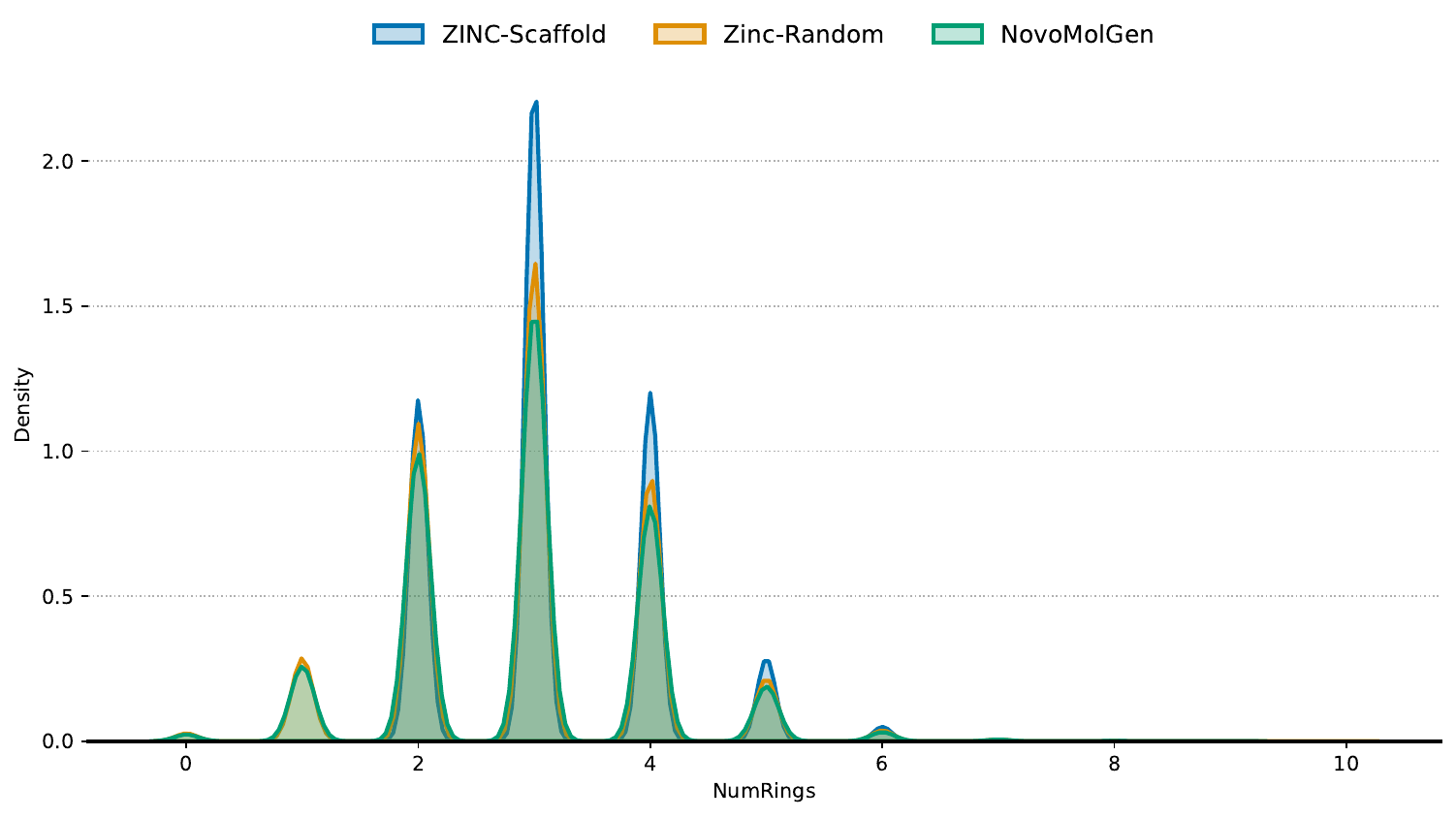}
    \caption{NumRings Distribution}
    \label{fig:subfig_rings}
  \end{subfigure}

  \caption{Molecular-property distributions for the random
    and scaffold reference splits (175{,}000 molecules each) and a
    generated set from \textbf{NovoMolGen-SMILES-BPE-32M} (100{,}000
    molecules). The properties include QED, SA, logP, MW, TPSA, Bertz
  complexity, number of rotatable bonds, and number of rings.}
  \label{fig:all_distributions}
\end{figure}

The kernel density estimation plots in~\cref{fig:all_distributions}
indicate that NovoMolGen successfully generates molecules whose
property distributions align closely with those of both the random
split and the scaffold split. Additionally, a lower
Wasserstein-1 distance is observed across all properties, further
demonstrating the model's ability to replicate the reference
distributions. The training dataset contains a higher proportion of
molecules with favorable drug-like properties, including higher
drug-likeness scores (QED $>$ 0.6), greater synthetic accessibility
(SA $<$ 4), optimal solubility (1 $<$ logP $<$ 4), and an ideal
molecular weight range (300 $<$ MW $<$ 500). By closely matching this
distribution, NovoMolGen generates molecules with an increased
likelihood of exhibiting drug-like characteristics. Furthermore, the
model effectively captures molecular topology, as reflected in the
distributions of NumRings, Rotatable Bonds, and Bertz Complexity.
These results highlight the potential of NovoMolGen in generating
chemically relevant and synthetically accessible molecules suitable
for drug discovery applications.

\section{Pretraining Configuration}
\label{app:pretraining_config}

We implement the Qwen3 decoder-only and T5Gemma encoder--decoder models
in the HuggingFace Trainer framework with FlashAttention
kernels~\citep{dao_flashattention_2024} for efficiency, and train in mixed
precision with bfloat16. We use the AdamW optimiser with a peak learning
rate of $6 \times 10^{-4}$ and a warmup--stable--decay (WSD) schedule. The
learning rate warms up over the first 2\% of steps, is held constant
through the stable phase, and each reported checkpoint is taken after a
short linear cooldown to $6 \times 10^{-5}$ branched from the stable
trajectory. This matches the schedule used for the IsoFLOP profiles in
\cref{app:isoflop_fits}.

We use token-aware batching of approximately 25{,}000 non-padding tokens
per optimiser step, with gradient accumulation and per-device batch size
chosen to fit 4 NVIDIA A100 GPUs (80 GB each). Weight decay is 0.01,
gradient clipping uses a maximum norm of 1.0, and AdamW uses $\beta_1 =
0.9$ and $\beta_2 = 0.95$. Design-axis comparisons match model size and
training tokens at $D = 20N$ rather than raw compute. For T5Gemma we
estimate compute as $C \approx 3ND$ because scored tokens pass only
through the decoder, so a T5Gemma model matched in size and tokens to a
Qwen3 model uses roughly half the training compute. The architectural
configurations are summarised in~\cref{tab:model_arch}, following the
width-to-depth ratio of~\citet{levine_depth_2020}.

\begin{table}[htb]
  \caption{NovoMolGen Configurations}
  \label{tab:model_arch}
  \centering
  \resizebox{0.7\linewidth}{!}{
    \begin{tabular}{l|c|c|c}
      \toprule
      Components        & NovoMolGen-32M & NovoMolGen-157M & NovoMolGen-300M \\
      \midrule
      Attention Heads   & 8              & 10              & 12              \\
      Hidden Layers     & 12             & 24              & 32              \\
      Hidden Size       & 512            & 640             & 768             \\
      Intermediate Size & 1024           & 2560            & 3072            \\
      \bottomrule
  \end{tabular}}
\end{table}

\section{Mechanistic Probing Methodology}
\label{app:mech_probing_appendix}

\subsection{Layer-wise Probing}

To quantify the explicit encoding of chemical properties within
NovoMolGen's internal representations, we employed the linear probing
method. This technique involves training simple linear classifiers or
regressors on the model's frozen activation vectors to predict
specific external properties, where high performance indicates that
the information is linearly accessible at that specific depth. Let
$h_\ell^{(i)} \in \mathbb{R}^{d}$ denote the activation vector of the
$i$-th molecule at layer $\ell$. We trained a separate linear probe
for each layer and property.

For continuous regression tasks (e.g., LogP, QED, Molecular Weight),
we utilised \textbf{Ridge Regression}. The probe learns weights $W
\in \mathbb{R}^{d}$ and bias $b \in \mathbb{R}$ to minimise the
regularised squared error:
\begin{equation}
  \mathcal{L}_{reg} = \sum_{i} ||y^{(i)} - (W^T h_\ell^{(i)} + b)||^2
  + \alpha ||W||^2_2
\end{equation}
where activations were standardised (zero mean, unit variance) prior
to training, and the regularisation strength $\alpha$ was selected by
five-fold cross-validation on the training split. For categorical
classification tasks, such as validity
checks, we utilised \textbf{Logistic Regression}, minimising the
cross-entropy loss with $L_2$ regularisation (using the multinomial
formulation for multiclass problems).

We evaluated the model on two distinct task categories. First, for
\textbf{Property Prediction}, we curated a set of over 30 chemical
descriptors, ranging from simple physicochemical properties (e.g.,
Molecular Weight, NumRings) to complex bioactivity scores (e.g.,
Docking Scores, QED), with ground-truth values computed using RDKit
and standard docking simulations. Second, for \textbf{Validity
Detection}, we assessed the model's ability to distinguish valid
chemical syntax from noise via a binary classification task. The
positive class consisted of activations from valid SMILES sequences,
while the negative class consisted of "corrupted" activations
extracted from structurally invalid or randomised sequences.

We use a fixed 80/20 train/test split and report performance on the
validation set 20\%, with the regularisation strength selected by five-fold
cross-validation on the training portion. For regression we report the
coefficient of determination ($R^2$), mean absolute error (MAE) and
Spearman rank correlation ($\rho$), and for classification we report
accuracy and $F_1$. All probes use the \texttt{scikit-learn}
library~\citep{kramer2016scikit}, with input activations frozen and
detached from the computational graph so that probing does not update the
language model parameters.

\subsection{Sparse Autoencoders Training}

To investigate the internal representations of NovoMolGen, we trained
Sparse Autoencoders (SAEs) on the residual stream activations of
specific transformer layers. We use the \textbf{BatchTopK} sparse
autoencoder~\citep{bussmann2024batchtopk}, which enforces sparsity across
a whole batch rather than per input. For a batch $X$ of $B$ activation
vectors it retains the $kB$ largest activations over the entire batch, so
the number of active features can vary between molecules while averaging
$k$ per input. This improves the trade-off between reconstruction fidelity
and sparsity compared with a fixed $L_1$ penalty. The SAE has an encoder
$W_{enc} \in \mathbb{R}^{d_{model} \times d_{dict}}$ and a decoder
$W_{dec} \in \mathbb{R}^{d_{dict} \times d_{model}}$, and the forward pass
is

\begin{align}
  a(x) &= \text{ReLU}(x W_{enc} + b_{enc}) \\
  f(X) &= \text{BatchTopK}_{k}\big(a(X)\big) \\
  \hat{x} &= f(x) W_{dec} + b_{dec}
\end{align}

where $\text{BatchTopK}_{k}$ keeps the $kB$ largest entries across the
batch $X$ and zeroes the rest, and $d_{dict} = M \times d_{model}$ is the
dictionary size.

Activations were extracted from different layers of (early, middle
and late) \textbf{NovoMolGen}. To ensure diverse, uncorrelated
training batches, we employed an activation-buffering strategy. The
source data consisted of the active set molecules,
which were tokenised using the model's native tokeniser (BPE or
AtomWise) with a fixed context length of 64 tokens. We utilised a
shuffled \texttt{ActivationBuffer} with a capacity of $3 \times 10^4$
contexts, continuously refreshed with batches of 512 sequences from
the language model to prevent distributional shift during training.

All SAEs were trained using the settings detailed in
Table~\ref{tab:sae_hyperparams}. We employed an auxiliary loss
($\mathcal{L}_{aux}$) to mitigate ``dead features'' (neurons that
never activate).

\begin{figure}[!ht]
  \centering
  \includegraphics[width=\linewidth]{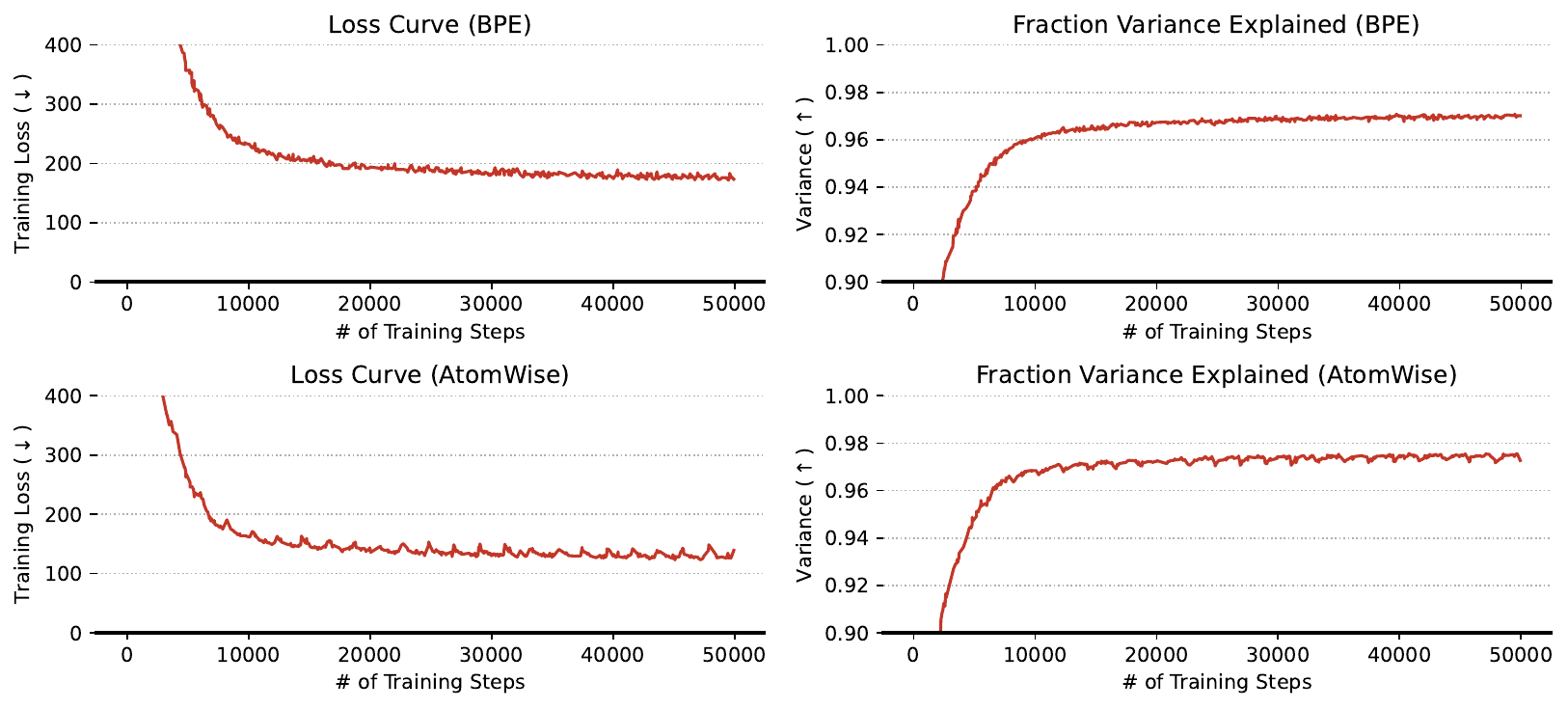}
  \caption{Training curves for loss and fraction of variance explained
  for NovoMolGen-SMILES-BPE-32M at checkpoint 75000 and layer 6.}
  \label{fig:sae_curves}
\end{figure}

\begin{table}[!ht]
  \centering
  \caption{Hyperparameters for Sparse Autoencoder Training.}
  \label{tab:sae_hyperparams}
  \resizebox{0.8\textwidth}{!}{
    \begin{tabular}{l l l}
      \toprule
      \textbf{Hyperparameter} & \textbf{Value} & \textbf{Description} \\
      \midrule
      Dictionary Multiplier & $16\times$ & Expansion factor ($16
      \times 512 = 8192$ features) \\
      Sparsity ($k$) & $\approx 2\%$ & Top-$k$ active features allowed \\
      Optimizer & Adam & Standard configuration \\
      Learning Rate & $3 \times 10^{-4}$ & Linear warmup over first
      2,500 steps \\
      Batch Size (SAE) & 4,096 & Activation vectors per optimization step \\
      Training Steps & 50,000 & Total updates ($\approx 2 \times
      10^8$ tokens) \\
      Dead Feature Loss ($\alpha$) & $1/32$ & Coefficient for auxiliary loss \\
      Threshold Adaption ($\beta$) & $0.999$ & EMA decay for
      threshold adaptation \\
      \bottomrule
  \end{tabular}}
\end{table}

To assess the quality of the learned dictionary, we evaluated each SAE on
a validation split of the active set using four metrics: (1) \textbf{Fraction
of Variance Explained (FVE)}, $1 - \|\hat{x} - x\|^2/\|x\|^2$; (2)
\textbf{$L_0$ norm}, the average number of active features per token
(target $\approx k$); (3) \textbf{Fraction Alive}, the fraction of
dictionary features that fire at least once; and (4) \textbf{CE Loss
Recovered}, the fraction of the language-modelling loss recovered when the
residual stream is replaced by the SAE
reconstruction~\citep{cunningham2023sparse}.
To relate features to chemistry, we correlate each live feature with the 36
property targets used for probing, namely the physicochemical descriptors,
PMO objectives and docking scores. For each molecule we max-pool a
feature's activations over its tokens to obtain one value per molecule, and
take its Pearson correlation with each property. \textbf{Monosemanticity}
is the mean over live features of the largest absolute property correlation,
and we call a feature property-associated when this value exceeds $0.4$.
The training curves in~\cref{fig:sae_curves} show that the fraction of
variance explained rises quickly and saturates early in training.

To empirically validate the hypothesis that Sparse Autoencoder (SAE)
features explicitly disentangle abstract chemical concepts better
than the model's raw basis directions, we conducted a controlled
substructure-detection experiment. We curated a set of 10 common
chemical functional groups defined by
SMARTS patterns and generated binary ground-truth labels $y \in \{0,
1\}^N$ for a random subset of $N=10,000$ valid molecules from the
active set using RDKit. For each molecule, we extracted the internal
activations from the target transformer layer and mean-pooled them over
non-padding positions, yielding one activation vector per molecule. We
then trained a Logistic Regression classifier with class balancing and
standard scaling, once on the SAE latents and once on the
residual-stream neurons, to predict the presence of each substructure.
Performance was evaluated by validation set $F_1$-score with 5-fold stratified
cross-validation to guard against class imbalance. We report the $F_1$ of
the probe on SAE latents against the probe on neurons
in~\cref{tab:substructure_f1_checkpoints}, and read a positive difference
as evidence that the concept is more linearly accessible in the SAE
dictionary. The SAE latents win for seven of the ten groups, and where
the neurons win the margin is negligible, so the wider dictionary offers
a modest but consistent gain rather than a decisive one.

\begin{table}[!ht]
  \centering
  \caption{Validation set $F_1$ for detecting the ten most common functional
    groups from a linear probe on SAE latents against a probe on
    residual-stream neurons (middle layer of the 32M
      PubChem--SMILES--atomwise reference at the final checkpoint, averaged
    over folds). SAE latents give the higher $F_1$ for seven of the ten
    groups, with gains of up to about 0.04. Where neurons win, the margin
  is within 0.003. \textbf{Bold} indicates the higher score.}
  \label{tab:substructure_f1_checkpoints}
  \resizebox{0.55\textwidth}{!}{%
    \begin{tabular}{l cc c}
      \toprule
      \textbf{Functional Group} & \textbf{SAE latents} &
      \textbf{Neurons} & \textbf{$\Delta$} \\
      \midrule
      Carbonyl            & \textbf{0.999} & 0.982 & $+0.017$ \\
      Carbonyl (non-acid) & \textbf{0.997} & 0.965 & $+0.032$ \\
      Bicyclic            & \textbf{0.997} & 0.984 & $+0.013$ \\
      Ether               & \textbf{0.995} & 0.968 & $+0.027$ \\
      Secondary amine     & \textbf{0.993} & 0.963 & $+0.030$ \\
      Amide               & \textbf{0.993} & 0.960 & $+0.033$ \\
      Aniline             & \textbf{0.986} & 0.952 & $+0.034$ \\
      Benzene             & 0.991 & \textbf{0.992} & $-0.001$ \\
      Aromatic N          & 0.990 & \textbf{0.993} & $-0.003$ \\
      Tertiary amine      & 0.988 & \textbf{0.989} & $-0.001$ \\
      \bottomrule
    \end{tabular}%
  }
\end{table}

\section{Fine-tuning Methodology}
\label{app:finetuning_hyperparams}

The agent is trained to minimize the cost function $J(X_k)$ over the
subset of top-$k$ molecules, $X_k$, defined as:
\begin{equation}
  J(X_k) = \frac{1}{k} \sum_{x \in X_k} \left[ \log
  P_\text{prior}(x)-\log P_\text{agent}(x) + \sigma \cdot s(x) \right]^2
  \label{eq:augmented_hill_climb_topk}
\end{equation}
where $\sigma$ is a scaling factor that controls the influence of the
reward $s(x)$. The reward is the task score mapped to $[0,1]$. For PMO
tasks it is the benchmark oracle score, and for docking it is the
QuickVina2 docking score min-max scaled to $[0,1]$ across the batch. This
loss balances the generation of high-reward molecules against adherence
to the chemically valid distribution of the prior. We add a penalty term,
$$J_p(X_k) = \frac{-1}{\frac{1}{k} \sum_{x \in X_k}\log P_\text{agent}(x)},$$
which grows as the agent becomes overconfident and its mean log-likelihood
approaches zero, and shrinks as that likelihood decreases. Minimising it
therefore discourages the agent from collapsing onto a narrow set of very
high likelihood molecules, which preserves exploration and stabilises
training. The final objective combines these components,
\begin{equation}
  J_\theta = J + \lambda \cdot J_p,
  \label{eq:total_loss}
\end{equation}
where $\lambda$ weights the penalty term. To improve sample efficiency we
use an experience replay buffer~\citep{guo2024saturn} that stores the 100
top-performing molecules found during fine-tuning. It is pre-initialised
with the 100 highest-reward ZINC-250K molecules per
task~\citep{irwin2012zinc}. These seed molecules are scored once before
optimisation and are not counted against the per-task oracle budget, and
they are excluded from the reported novel candidates so they cannot
inflate the downstream metrics.

To optimize our models for goal-directed design, we performed a
systematic hyperparameter search using a REINVENT-inspired Hill
Climbing framework. We ran a separate search for each of the two task
categories. For multi-property optimization we tuned on the
\textbf{Perindopril\_MPO} and \textbf{Zaleplon\_MPO} tasks from the PMO
benchmark, and for protein-ligand docking we tuned on the \textbf{fa7}
and \textbf{jak2} targets. Within each category, settings were selected by
the aggregated score over its two tasks, and each configuration was
evaluated over three random seeds.

The hyperparameter space explored includes the following, with the exact
grid varying slightly across representation and size variants:
\begin{itemize}
  \item \textbf{Penalty Coefficient (\(\lambda\))}: \([10, 100, 500, 2000]\)
  \item \textbf{Batch Size}: \([32, 64, 128]\)
  \item \textbf{Sigma (\(\sigma\))}: \([500, 1000, 2000, 2500]\)
  \item \textbf{Learning Rate (\(lr\))}: \([1 \times 10^{-3}, 5
    \times 10^{-4}, 1 \times 10^{-4}]\)
  \item \textbf{Fraction of Top-$k$ Molecules}: \([0.1, 0.5]\)
\end{itemize}

The default value for penalty coefficient from the REINVENT
implementation ($5000$) proved unsuitable for our models, likely due
to differences in the log-likelihood scale of generated molecules,
leading to unstable training and a high proportion of invalid solutions.

We evaluated 96 configurations, each on both task categories over three
seeds, for \textbf{576 runs} in total, where a run is one configuration
evaluated on one task category for one seed ($96 \times 2 \times 3 = 576$).
Our model could not generate any molecules for
the \textbf{Valsartan\_SMARTS} task, necessitating its exclusion from
our analysis. This limitation stems directly from our choice of
pre-training data, which lacks the specific chemical substructures
required by the task's SMARTS pattern. We note that this is a
data-dependent constraint, and other models may succeed where ours
could not. For instance, a method like \textbf{\frag}, which builds
its fragment vocabulary from the \texttt{UniChem} database, would
likely be able to generate valid solutions provided that this
database contains the relevant chemical motifs. The parallel
coordinates plot are shown
in~\cref{fig:smiles_32m_atom_parallel_plot,fig:smiles_32m_bpe_parallel_plot,fig:safe_parallel_plot,fig:safe_bpe_parallel_plot,fig:selfies_parallel_plot,fig:selfies_bpe_parallel_plot,fig:smiles_157m_atom_parallel_plot,fig:smiles_157m_bpe_parallel_plot,fig:smiles_300m_atom_parallel_plot,fig:smiles_300m_bpe_parallel_plot}.

\begin{figure}[htb]
  \centering
  \includegraphics[width=0.9\linewidth]{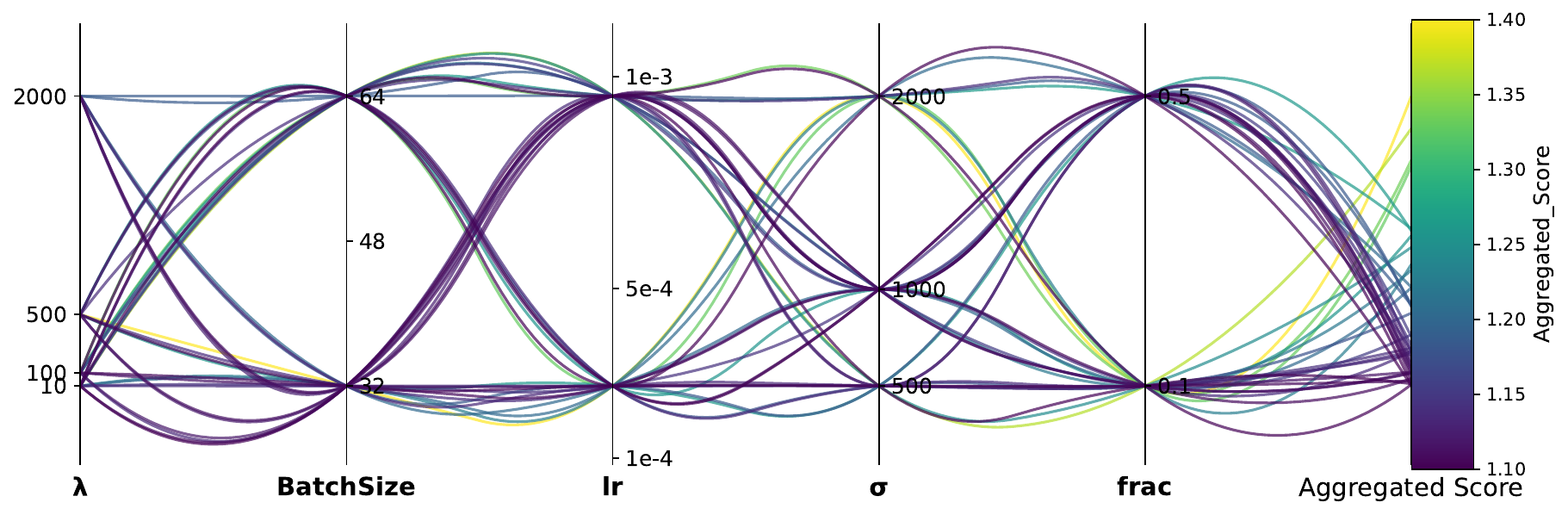}
  \caption{Parallel Coordinates Plot for the \textbf{SMILES
    (AtomWise)} model (32M), showing the importance and effects of
  hyperparameters on the Aggregated Score.}
  \label{fig:smiles_32m_atom_parallel_plot}
\end{figure}

\begin{figure}[htb]
  \centering
  \includegraphics[width=0.9\linewidth]{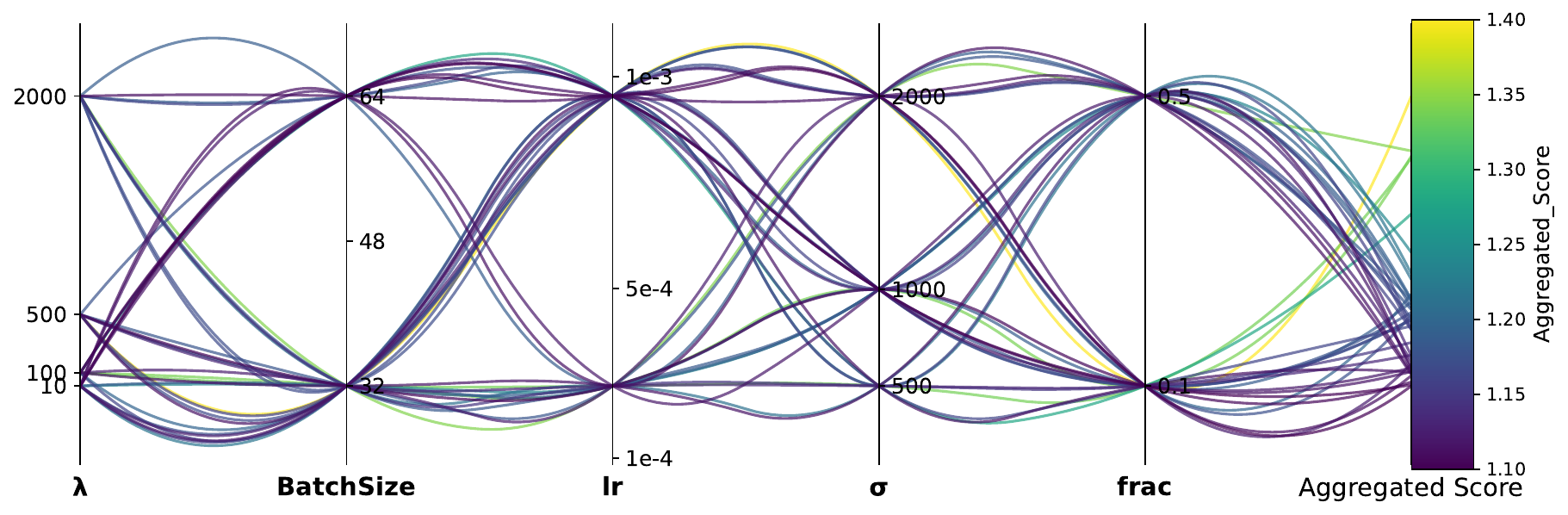}
  \caption{Parallel Coordinates Plot for the \textbf{SMILES (BPE)} model (32M).}
  \label{fig:smiles_32m_bpe_parallel_plot}
\end{figure}

\begin{figure}[htb]
  \centering
  \includegraphics[width=0.9\linewidth]{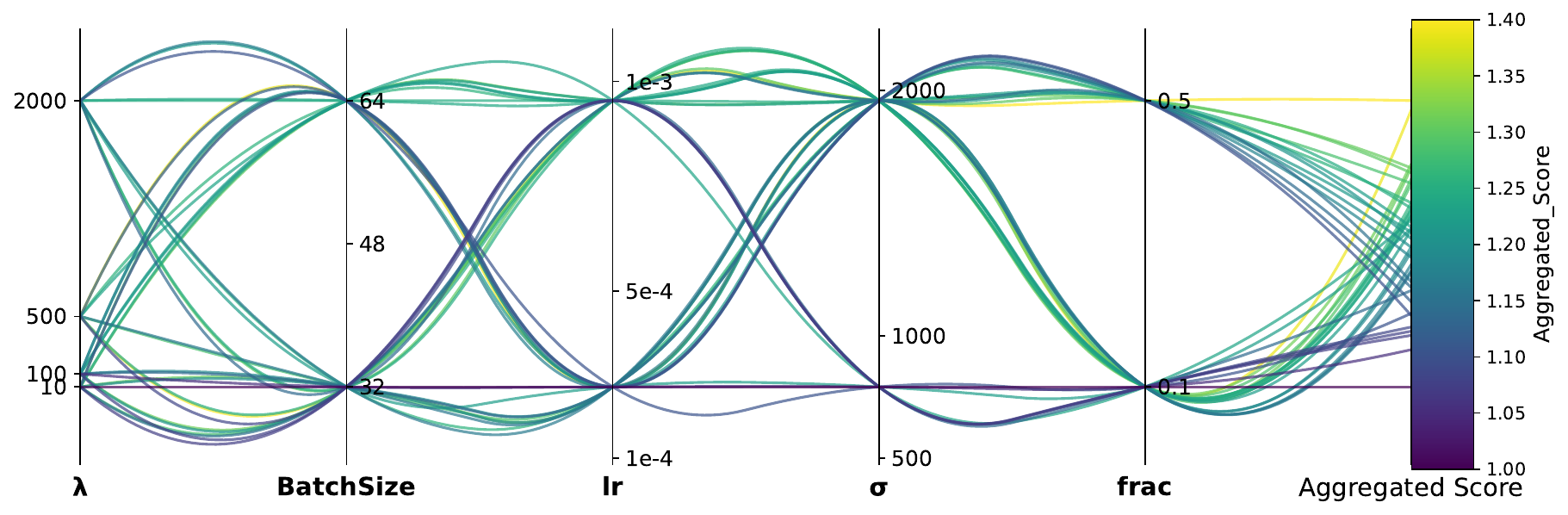}
  \caption{Parallel Coordinates Plot for the \textbf{SAFE (AtomWise)} model.}
  \label{fig:safe_parallel_plot}
\end{figure}

\begin{figure}[htb]
  \centering
  \includegraphics[width=0.9\linewidth]{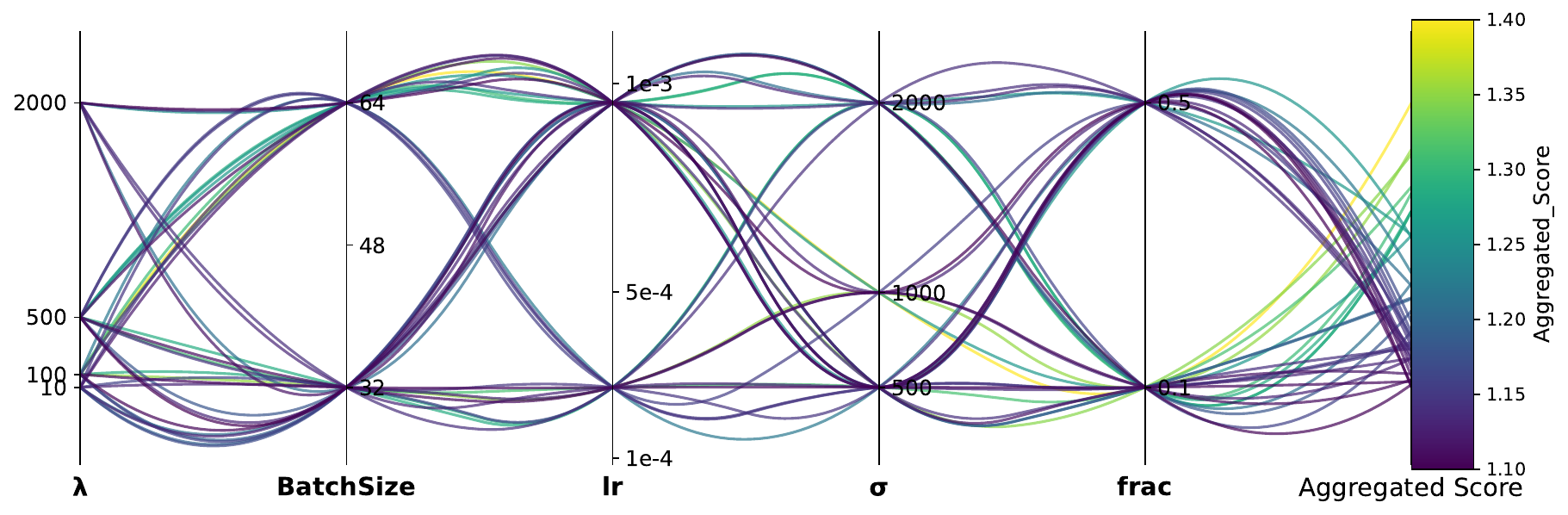}
  \caption{Parallel Coordinates Plot for the \textbf{SAFE (BPE)} model.}
  \label{fig:safe_bpe_parallel_plot}
\end{figure}

\begin{figure}[htb]
  \centering
  \includegraphics[width=0.9\linewidth]{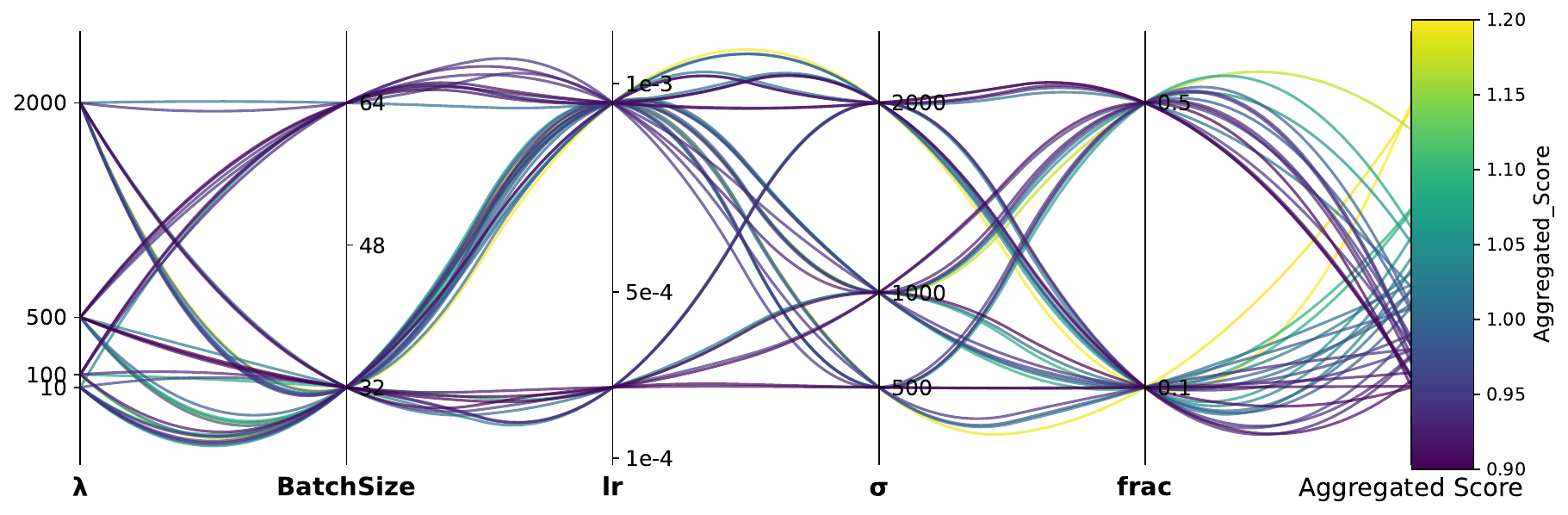}
  \caption{Parallel Coordinates Plot for the \textbf{SELFIES (AtomWise)} model.}
  \label{fig:selfies_parallel_plot}
\end{figure}

\begin{figure}[htb]
  \centering
  \includegraphics[width=0.9\linewidth]{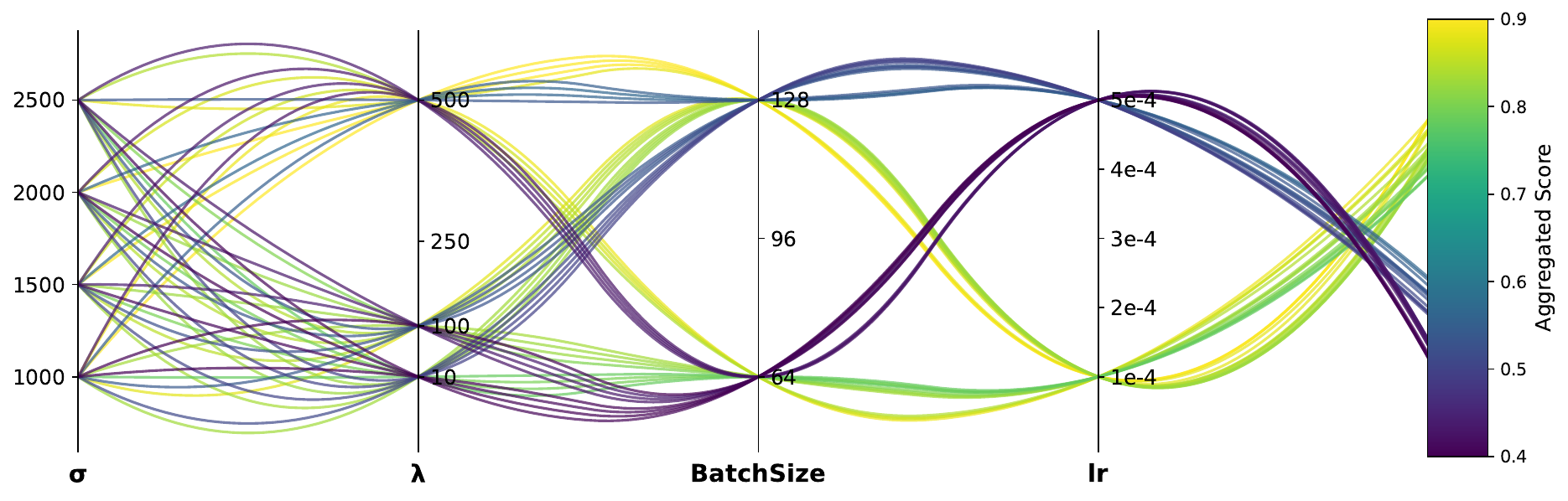}
  \caption{Parallel Coordinates Plot for the \textbf{SELFIES (BPE)} model.}
  \label{fig:selfies_bpe_parallel_plot}
\end{figure}

\begin{figure}[htb]
  \centering
  \includegraphics[width=0.9\linewidth]{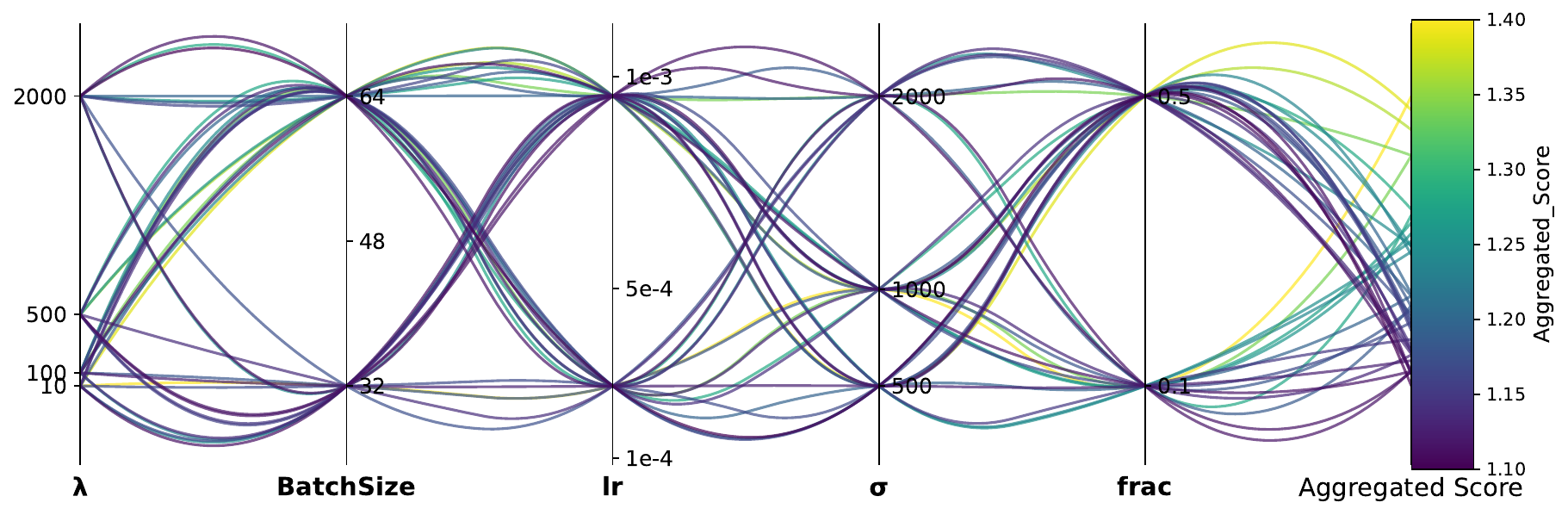}
  \caption{Parallel Coordinates Plot for the \textbf{SMILES
  (AtomWise)} model (157M).}
  \label{fig:smiles_157m_atom_parallel_plot}
\end{figure}

\begin{figure}[htb]
  \centering
  \includegraphics[width=0.9\linewidth]{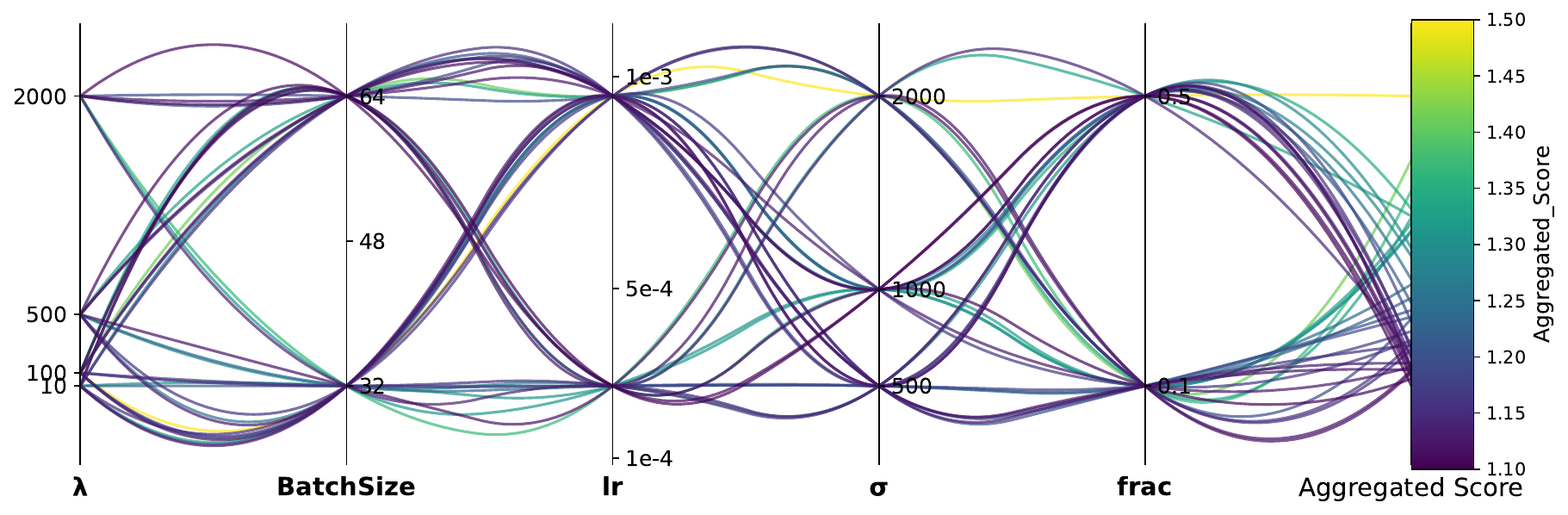}
  \caption{Parallel Coordinates Plot for the \textbf{SMILES (BPE)}
  model (157M).}
  \label{fig:smiles_157m_bpe_parallel_plot}
\end{figure}

\begin{figure}[htb]
  \centering
  \includegraphics[width=0.9\linewidth]{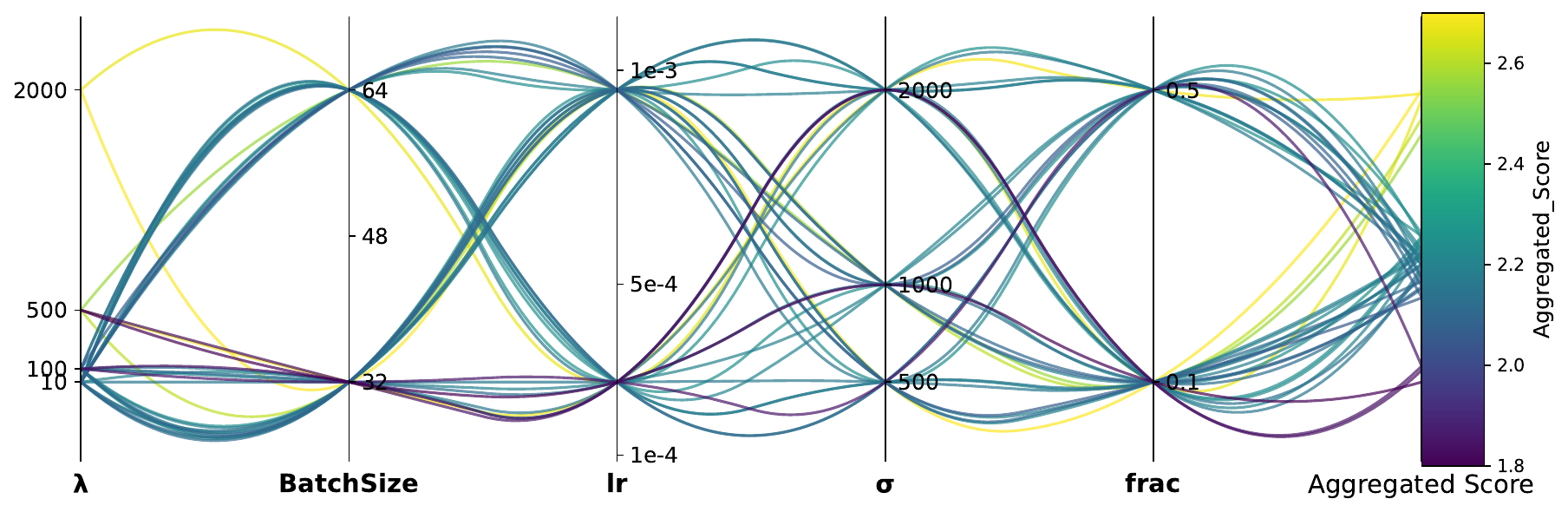}
  \caption{Parallel Coordinates Plot for the \textbf{SMILES
  (AtomWise)} model (300M).}
  \label{fig:smiles_300m_atom_parallel_plot}
\end{figure}

\begin{figure}[htb]
  \centering
  \includegraphics[width=0.9\linewidth]{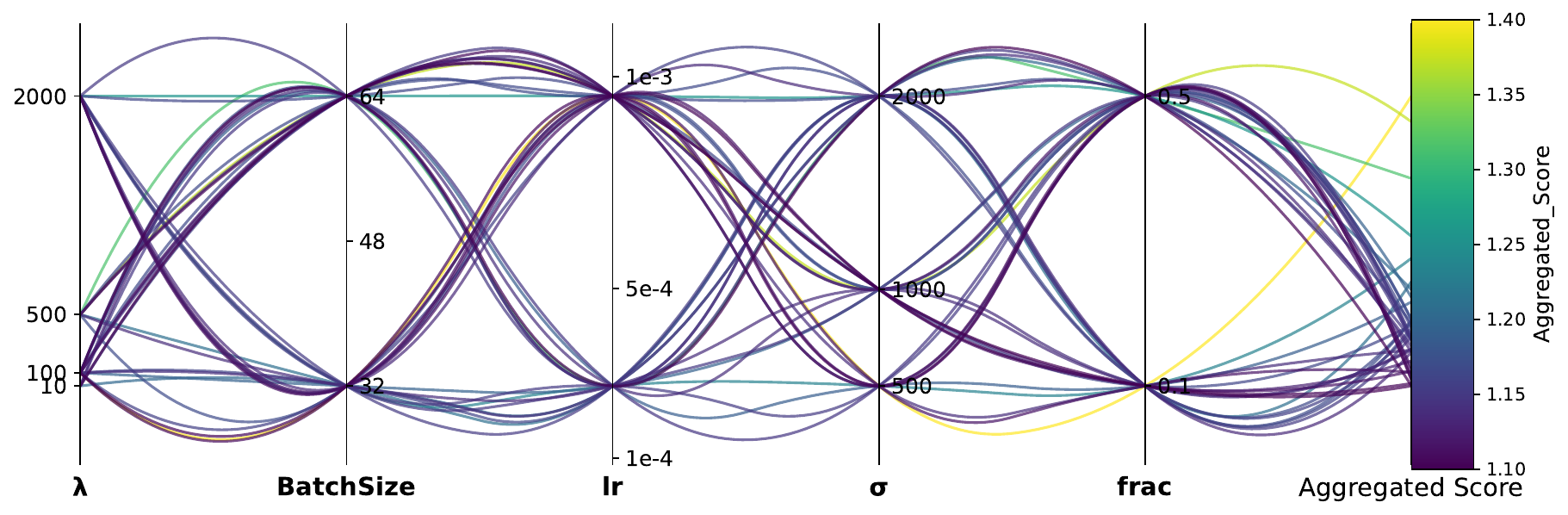}
  \caption{Parallel Coordinates Plot for the \textbf{SMILES (BPE)}
  model (300M).}
  \label{fig:smiles_300m_bpe_parallel_plot}
\end{figure}

\section{IsoFLOP Experiments}
\label{app:isoflop_fits}

We locate the compute-optimal frontier with IsoFLOP profiles, which
compare model sizes at a fixed compute budget $C \approx 6ND$. Training a
separate model to convergence at every $(N, C)$ pair would be
prohibitively expensive, so we use a single warmup--stable--decay
trajectory per model size. Each model runs one long warmup--stable phase,
and every IsoFLOP point is measured after a short linear cooldown branched
from that phase at the target token count. This gives a well-converged
checkpoint at each budget without retraining from scratch and keeps the
grid affordable.

\Cref{tab:isoflop_grid} lists the five compute budgets and the model sizes
evaluated near each budget's optimum.
Following~\citet{hoffmannTrainingComputeoptimalLarge2022},
we fit a parametric loss surface
\begin{equation}
  L(N, D) = E + \frac{A}{N^{\alpha}} + \frac{B}{D^{\beta}}
  \label{eq:loss_surface}
\end{equation}
to the validation set losses of the reference models, then minimise it at fixed
$C = 6ND$ to read the loss-minimising size $N_{\mathrm{opt}}(C)$ at each
budget. The minima follow an approximate power law, $N_{\mathrm{opt}}
\propto C^{0.69}$ and $D_{\mathrm{opt}} \propto C^{0.31}$, with a 68\%
bootstrap interval on the parameter exponent of 0.59 to 0.76 obtained by
resampling the training trajectories 1,000 times. Across the grid the
compute-optimal size grows from about 2.5M parameters at $3\times10^{15}$
FLOPs to about 700M at $10^{19}$ FLOPs, while the optimal token count grows
more slowly, consistent with an exponent below one-half.

The optimal token-to-parameter ratio drifts along the frontier, so we adopt
a single practical rule for all reported comparisons. Training every model to
$D = 20N$ stays within 1\% of the fitted optimal loss across the budget
range of the grid, $3\times10^{15}$ to $10^{19}$ FLOPs, so we train and
compare all sizes at $D = 20N$ and refer to these checkpoints as
compute-optimal. This is a near-optimal practical rule rather than the exact
optimum, since $D = 20N$ corresponds to $N \propto C^{0.5}$ whereas the
fitted allocation is $N_{\mathrm{opt}} \propto C^{0.69}$. For the smallest
and largest sizes in the reported sweep (0.5M and 1B), which fall just
outside the fitted budget range, the near-optimality is inferred from the
fitted loss surface rather than from bracketing IsoFLOP points.

\begin{table}[htb]
  \caption{IsoFLOP grid. For each compute budget we evaluate the model
    sizes that bracket the optimum and read the compute-optimal parameter
    count $N_{\mathrm{opt}}$ and token count $D_{\mathrm{opt}}$ from the
  fitted loss surface.}
  \label{tab:isoflop_grid}
  \centering
  \resizebox{0.9\linewidth}{!}{
    \begin{tabular}{lccc}
      \toprule
      Budget $C$ (FLOPs) & Evaluated sizes & $N_{\mathrm{opt}}$ &
      $D_{\mathrm{opt}}$ (tokens) \\
      \midrule
      $3\times10^{15}$ & 0.5M, 5M, 32M      & 2.5M & $2.0\times10^{8}$ \\
      $3\times10^{16}$ & 5M, 32M, 157M      & 12M  & $4.2\times10^{8}$ \\
      $3\times10^{17}$ & 5M, 32M, 157M      & 55M  & $9.1\times10^{8}$ \\
      $1\times10^{18}$ & 32M, 157M, 300M    & 130M & $1.3\times10^{9}$ \\
      $1\times10^{19}$ & 157M, 300M, 1B     & 700M & $2.4\times10^{9}$ \\
      \bottomrule
  \end{tabular}}
\end{table}

\section{Loss Metrics}
\label{app:bits_per_token}

We compare models in bits per molecule because per-token loss is not
comparable across representations and tokenisers that encode the same
molecule with different sequence lengths. \Cref{tab:bits_decomposition}
decomposes the validation set bits per molecule at 32M into bits per token and
tokens per molecule for each encoding and corpus, all on the random split.
The decomposition explains why the per-molecule ranking differs from the
per-token ranking. SELFIES has the lowest bits per token, yet it needs
about 155 tokens to encode a molecule against 47 for SMILES, so its bits
per molecule are higher. Byte-pair encoding raises bits per token but
shortens the sequence, which is why it overtakes atom-level tokenisation
from 32M onwards. Across corpora the per-molecule loss tracks corpus
diversity rather than sequence length, since ChEMBL molecules are longer
than PubChem molecules yet reach a lower loss.

\Cref{tab:loss_by_size} reports the validation set bits per molecule of the
PubChem--SMILES--atomwise reference across the full size range on the
random split.

\begin{table}[htb]
  \caption{Validation set loss at 32M decomposed into bits per token and tokens
    per molecule on the random split. Bits per molecule equals bits per
  token times tokens per molecule.}
  \label{tab:bits_decomposition}
  \centering
  \resizebox{0.95\linewidth}{!}{
    \begin{tabular}{llccc}
      \toprule
      Axis & Setting & Bits / molecule ($\downarrow$) & Bits / token
      & Tokens / molecule \\
      \midrule
      \multirow{3}{*}{Representation}
      & SMILES (atomwise)   & 48.1 & 1.02 & 47 \\
      & SELFIES (atomwise)  & 50.9 & 0.33 & 155 \\
      & SAFE (atomwise)     & 50.2 & 0.95 & 53 \\
      \midrule
      \multirow{2}{*}{Tokenisation}
      & SMILES (atom-level) & 48.1 & 1.02 & 47 \\
      & SMILES (byte-pair)  & 46.4 & 1.86 & 25 \\
      \midrule
      \multirow{4}{*}{Corpus (SMILES, atomwise)}
      & MOSES    & 25.6 & 0.91 & 28 \\
      & ChEMBL   & 33.2 & 0.69 & 48 \\
      & PubChem  & 48.1 & 1.02 & 47 \\
      & ZINC-22  & 51.5 & 0.99 & 52 \\
      \bottomrule
  \end{tabular}}
\end{table}

\begin{table}[htb]
  \caption{Validation set bits per molecule of the PubChem--SMILES--atomwise
  reference on the random split, trained at $D = 20N$.}
  \label{tab:loss_by_size}
  \centering
  \resizebox{0.85\linewidth}{!}{
    \begin{tabular}{lcccccc}
      \toprule
      Model size & 0.5M & 5M & 32M & 157M & 300M & 1B \\
      \midrule
      Bits / molecule ($\downarrow$) & 71.5 & 54.1 & 48.1 & 46.0 &
      45.5 & 45.0 \\
      \bottomrule
  \end{tabular}}
\end{table}

\subsection{BPE Substructures}
\label{sec:bpe_appendix}

\begin{figure}[htb]
  \centering
  \includegraphics[width=0.75\linewidth]{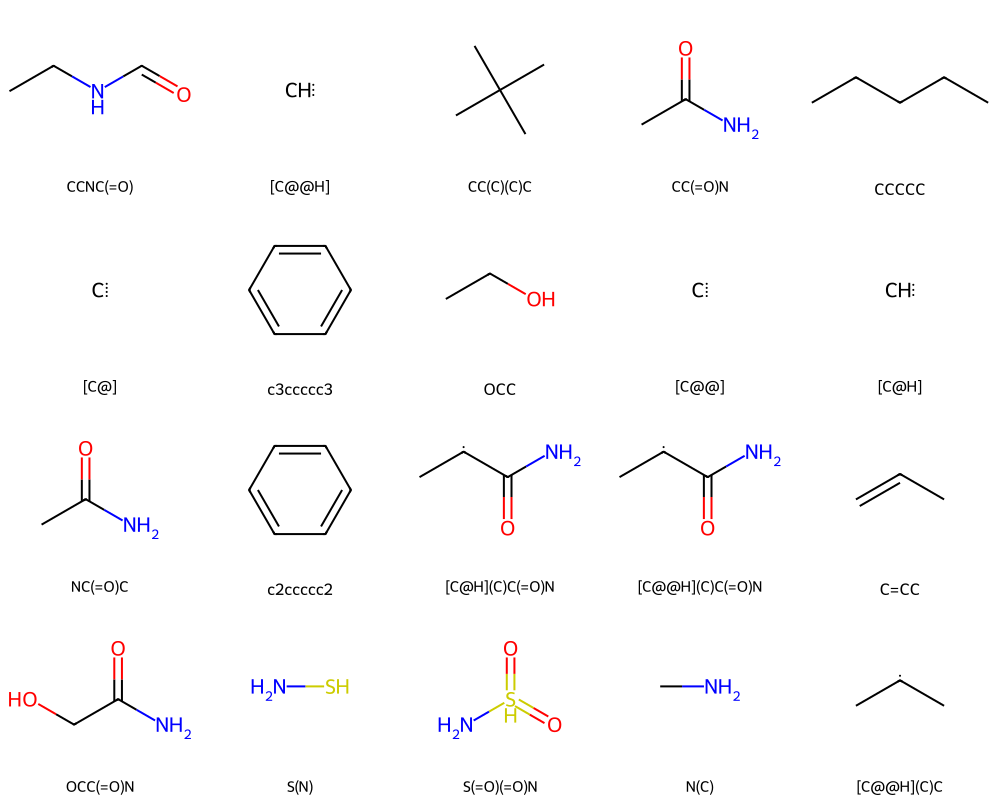}
  \caption{Substructures present in BPE tokenization}
  \label{fig:bpe_sub}
\end{figure}

We provide visualization of different substructures extracted from
the random 100M subset from ZINC-22 in~\cref{fig:bpe_sub}.

\section{OOD Generalization}
\label{app:ood_generation}

This section reports the per-split behaviour behind the out-of-distribution
analysis and the full unconstrained-generation metrics.
All values use the same $D = 20N$ checkpoints and the same 175,000-molecule
validation splits.

\subsection{Loss across splits}
\Cref{tab:ood_loss} reports the validation set
bits per molecule of the PubChem--SMILES--atomwise reference on all four
splits. The scaffold split stays close to the random split at every scale,
so it behaves almost as in-distribution. The molecular-weight and
maximum-dissimilarity splits are far harder, and their loss ratio to the
random split narrows as the model grows, from 5.9$\times$ and 2.4$\times$
at 0.5M to 2.8$\times$ and 1.3$\times$ at 1B. Larger models therefore
generalise better in likelihood, even though the shifted splits remain
much harder than the in-distribution reference.

\subsection{OOD gap across design axes}
The relative increase in bits per
molecule from the random to the maximum-dissimilarity split depends on the
encoding. At 32M this gap is smallest for byte-pair SMILES and largest for
SELFIES, as shown in \cref{tab:ood_gap}. Measured on their own splits, the
gap narrows with scale on the broad PubChem and ZINC-22 corpora but widens
on the narrower MOSES and ChEMBL corpora, which points to tighter fitting
of the smaller training distributions.

\subsection{Unconstrained generation}
We sample 30,000 molecules per model
at temperature $T = 1.0$ without top-$k$ or top-$p$ filtering and score
them against the validation splits. \Cref{tab:dist_metrics} reports validity,
internal diversity, novelty, and the distribution metrics on the random
split across the design axes and the reference size sweep. Lower loss does
not translate into a lower Fr\'echet ChemNet Distance. On PubChem the FCD
of the SMILES reference rises from 2.8 at 5M to 5.1 at 32M even though the
loss falls, and SELFIES and SAFE reach a lower FCD than SMILES at 32M
despite higher bits per molecule. On the narrower MOSES and ChEMBL corpora
the FCD is far lower and does fall with loss, since those distributions are
easier to match. \Cref{tab:fcd_splits} shows that sampled molecules stay
anchored to the training distribution, with FCD rising from a few units
against the random split to 26 to 29 against the maximum-dissimilarity
split for every representation.

\begin{table}[htb]
  \caption{Validation set bits per molecule of the PubChem--SMILES--atomwise
    reference on all four evaluation splits ($\downarrow$). The final row
    is the maximum-dissimilarity loss as a multiple of the random-split
  loss.}
  \label{tab:ood_loss}
  \centering
  \resizebox{0.9\linewidth}{!}{
    \begin{tabular}{lcccccc}
      \toprule
      Split & 0.5M & 5M & 32M & 157M & 300M & 1B \\
      \midrule
      Random               & 71.5 & 54.1 & 48.1 & 46.0 & 45.5 & 45.0 \\
      Scaffold             & 72.9 & 55.2 & 49.1 & 46.9 & 46.4 & 45.9 \\
      Molecular weight     & 422  & 243  & 168  & 143  & 132  & 126  \\
      Maximum dissimilarity& 172  & 92   & 68   & 62   & 60   & 59   \\
      \midrule
      Max-dissim.\ / random & 2.4 & 1.7 & 1.4 & 1.35 & 1.32 & 1.3 \\
      \bottomrule
  \end{tabular}}
\end{table}

\begin{table}[htb]
  \caption{Out-of-distribution gap at 32M, the relative increase in bits
    per molecule from the random to the maximum-dissimilarity split
  ($\downarrow$).}
  \label{tab:ood_gap}
  \centering
  \resizebox{0.8\linewidth}{!}{
    \begin{tabular}{llc}
      \toprule
      Axis & Setting & OOD gap (\%) \\
      \midrule
      \multirow{3}{*}{Representation}
      & SMILES (atomwise)     & 41 \\
      & SELFIES (atomwise)    & 57 \\
      & SAFE (atomwise)       & 50 \\
      \midrule
      \multirow{2}{*}{Tokenisation}
      & Atom-level & 41 \\
      & Byte-pair  & 32 \\
      \midrule
      \multirow{2}{*}{Architecture}
      & Qwen3 (decoder)        & 41 \\
      & T5Gemma (enc--dec)     & 46 \\
      \bottomrule
  \end{tabular}}
\end{table}

\begin{table}[htb]
  \caption{Unconstrained-generation metrics on the random split for
    NovoMolGen across design axes. Values are averaged over pretraining
    seeds where available. Each corpus is scored on its own random split,
    so the FCD scale differs across corpora. Arrows show the preferred
  direction.}
  \label{tab:dist_metrics}
  \centering
  \resizebox{\textwidth}{!}{
    \begin{tabular}{lllccccccc}
      \toprule
      Axis & Setting & Size & Valid ($\uparrow$) & IntDiv
      ($\uparrow$) & Novelty ($\uparrow$) & FCD ($\downarrow$) & SNN
      ($\uparrow$) & Frag ($\uparrow$) & Scaf ($\uparrow$) \\
      \midrule
      \multirow{6}{*}{Ref.\ size}
      & PubChem, SMILES & 0.5M & 0.310 & 0.892 & 0.989 & 3.92 & 0.386
      & 0.948 & 0.469 \\
      &                 & 5M   & 0.901 & 0.894 & 0.970 & 2.84 & 0.402
      & 0.892 & 0.722 \\
      &                 & 32M  & 0.961 & 0.884 & 0.979 & 5.12 & 0.395
      & 0.901 & 0.314 \\
      &                 & 157M & 0.968 & 0.883 & 0.980 & 5.05 & 0.398
      & 0.905 & 0.330 \\
      &                 & 300M & 0.972 & 0.882 & 0.980 & 4.95 & 0.399
      & 0.908 & 0.335 \\
      &                 & 1B   & 0.978 & 0.882 & 0.981 & 4.82 & 0.401
      & 0.910 & 0.340 \\
      \midrule
      \multirow{2}{*}{Representation}
      & SELFIES (atomwise)    & 32M & 0.994 & 0.894 & 0.956 & 2.43 &
      0.400 & 0.954 & 0.662 \\
      & SAFE (atomwise)       & 32M & 0.947 & 0.895 & 0.986 & 2.93 &
      0.383 & 0.963 & 0.595 \\
      \midrule
      Tokenisation & SMILES (byte-pair) & 32M & 0.965 & 0.885 & 0.982
      & 4.60 & 0.398 & 0.910 & 0.340 \\
      \midrule
      \multirow{3}{*}{Corpus}
      & MOSES   & 32M & 0.989 & 0.854 & 0.716 & 0.14 & 0.564 & 0.999 & 0.890 \\
      & ChEMBL  & 32M & 0.978 & 0.876 & 0.896 & 0.22 & 0.482 & 0.999 & 0.544 \\
      & ZINC-22 & 32M & 0.960 & 0.900 & 0.990 & 6.50 & 0.380 & 0.885 & 0.300 \\
      \bottomrule
  \end{tabular}}
\end{table}

\begin{table}[htb]
  \caption{Fr\'echet ChemNet Distance ($\downarrow$) between 30,000
    molecules sampled from each 32M PubChem model and each evaluation
    split. Samples stay anchored to the training distribution, so the FCD
  grows sharply on the shifted splits.}
  \label{tab:fcd_splits}
  \centering
  \resizebox{0.9\linewidth}{!}{
    \begin{tabular}{lcccc}
      \toprule
      Representation (32M) & Random & Scaffold & Molecular weight &
      Maximum dissimilarity \\
      \midrule
      SMILES (atomwise)  & 5.12 & 13.9 & 23.4 & 29.0 \\
      SELFIES (atomwise) & 2.43 & 12.6 & 22.9 & 27.4 \\
      SAFE (atomwise)    & 2.93 & 12.5 & 22.8 & 26.2 \\
      \bottomrule
  \end{tabular}}
\end{table}

\subsection{Generation Metric Definitions}
\label{sec:evaluation}
We define the distribution-learning metrics used to assess unconstrained
generation, following the MOSES
benchmark~\citep{polykovskiy_molecular_2020}. These metrics are
computed based on the generated set of molecules from the model,
denoted as $G$, and two reference sets, the random split (Test) and the
scaffold split (TestSF), each holding out molecules from the same corpus
the model was trained on. All reported metrics are calculated using the
subset of $G$ consisting of valid molecules identified through a
post-generation filtering process except for validity. Novelty is
computed against the full training corpus rather than a small reference
set, so it is not directly comparable to values that prior work reports
against smaller datasets such as MOSES. Similar care applies to other
distribution-based metrics, such as SNN and scaffold similarity.

\begin{enumerate}
  \item \textbf{Validity}: Validity is determined using RDKit's
    molecular structure parser, which verifies atomic valency and the
    consistency of bonds within aromatic rings. The metric ensures
    that the model adheres to relevant chemical constraints and
    measures the proportion of valid molecules generated within $G$.
    For molecular representations other than SMILES, we convert the
    generated set to SMILES and assess whether the corresponding
    decoder successfully decodes the molecule string to SMILES
    format. This step is essential, as representations that adhere to
    their respective syntactic rules may still produce chemically
    invalid molecules.
  \item \textbf{Novelty}: This metric quantifies the proportion of
    molecules in $G$ that do not appear in the training dataset. The
    molecules in $G$ are canonicalized and compared against the full
    training corpus.
  \item \textbf{Internal Diversity (IntDiv)}: This metric quantifies
    the chemical diversity within a set of generated molecules ($G$).
    It is calculated by taking 1 minus the average pairwise Tanimoto
    similarity of the Morgan fingerprints for all molecules in the
    set. Higher IntDiv scores, which range from 0 (no diversity) to 1
    (maximum diversity), indicate a more structurally varied set of
    generated molecules.
  \item \textbf{Fréchet ChemNet Distance (FCD)}: Derived from the
    activations of the penultimate layer of ChemNet, a deep neural
    network trained to predict the biological activities of drugs,
    this measure captures the chemical and biological properties of
    molecules. Activations for canonical SMILES representations of
    molecules are compared between the generated and reference sets.
    Lower values indicate better overlap, and the metric is non-negative.
  \item \textbf{Fragment Similarity (Frag)}: The distribution of
    BRICS fragments~\citep{degen_art_2008} is compared between the
    generated and reference sets. Higher values signify a closer
    match in fragment distributions, ensuring no fragment is
    disproportionately overrepresented or underrepresented.
  \item \textbf{Scaffold Similarity (Scaff)}: Similar to Fragment
    Similarity, this comparison uses Bemis-Murcko scaffolds instead
    of BRICS fragments to evaluate the resemblance between scaffolds
    in the generated and reference datasets.
  \item \textbf{Similarity to Nearest Neighbor (SNN)}: The average
    Tanimoto similarity between each molecule in the generated set
    and its nearest counterpart in the reference set is calculated.
    Lower values suggest the generated molecules are farther from the
    reference set's manifold, while higher values indicate closer alignment.
\end{enumerate}

\subsection{Distribution Learning on the Scaffold Split}
\label{sec:pretraining_additional}
We sample 30,000 molecules per model and score them against the random
split (Test) and the scaffold split (TestSF), each holding out 175,000
molecules, averaging over three seeds. \Cref{tab:scaffold_split} compares
the two splits for the reference size sweep and the representation axis at
32M. The scaffold split is harder in sample space than in likelihood.
Validation set loss on the scaffold split is only about 2\% above the random
split, yet the Fr\'echet ChemNet Distance and scaffold similarity degrade
sharply, because the held-out scaffolds are almost absent from the
generated distribution. Within the SMILES reference the metrics change
little from 32M onwards. The full set of random-split metrics across
every design
axis is given in \cref{tab:dist_metrics}.

\begin{table}[htb]
  \caption{Random split (Test) versus scaffold split (TestSF) for
    NovoMolGen. FCD is lower-is-better, SNN and scaffold similarity are
    higher-is-better. The scaffold split degrades sample-space metrics even
  though its validation loss stays close to the random split.}
  \label{tab:scaffold_split}
  \centering
  \resizebox{\textwidth}{!}{
    \begin{tabular}{llcccccc}
      \toprule
      \multirow{2}{*}{Representation} & \multirow{2}{*}{Size}
      & \multicolumn{2}{c}{FCD ($\downarrow$)}
      & \multicolumn{2}{c}{SNN ($\uparrow$)}
      & \multicolumn{2}{c}{Scaf ($\uparrow$)} \\
      \cmidrule(lr){3-4}\cmidrule(lr){5-6}\cmidrule(lr){7-8}
      & & Test & TestSF & Test & TestSF & Test & TestSF \\
      \midrule
      \multirow{3}{*}{SMILES (atomwise)}
      & 0.5M & 3.92 & 15.03 & 0.386 & 0.333 & 0.469 & 0.064 \\
      & 5M   & 2.84 & 13.17 & 0.402 & 0.347 & 0.722 & 0.063 \\
      & 32M  & 5.12 & 13.91 & 0.395 & 0.345 & 0.314 & 0.028 \\
      \midrule
      SELFIES (atomwise) & 32M & 2.43 & 12.61 & 0.400 & 0.345 & 0.662 & 0.041 \\
      SAFE (atomwise)    & 32M & 2.93 & 12.53 & 0.383 & 0.336 & 0.595 & 0.092 \\
      \bottomrule
  \end{tabular}}
\end{table}

\section{PMO Benchmark}
\label{sec:appendix_pmo_extend}

\begin{figure}[htb]
  \centering
  \includegraphics[width=\linewidth]{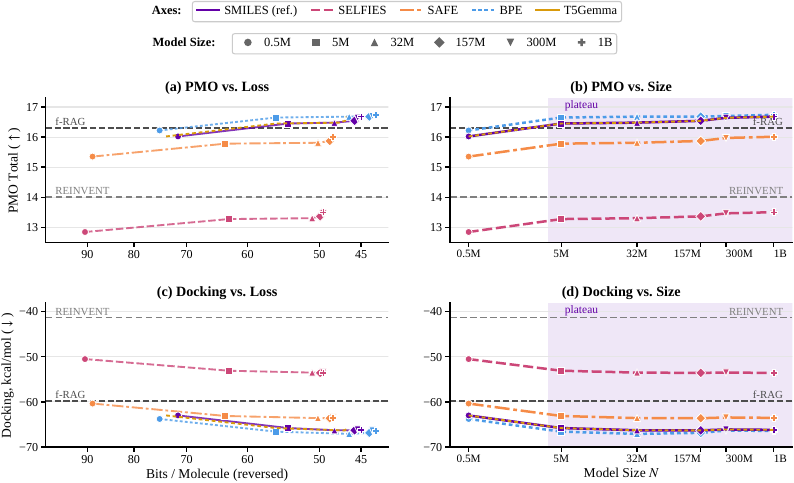}
  \caption{\textbf{Lower pretraining loss does not buy better
    goal-directed design.} Downstream performance after RL fine-tuning
    of decoder-only PubChem models across design axes (colour) and model
    sizes from 0.5M to 1B (marker). \textbf{(a)}~Total PMO score over 22
    tasks (higher is better) against validation set bits per molecule, with the
    loss axis reversed. \textbf{(b)}~Every axis plateaus in total PMO
    score beyond $\sim$5M (shaded). \textbf{(c,\,d)}~The same comparisons
    for the total top-5\% novel docking score over five targets (lower is
  better). Dashed lines mark \frag and REINVENT~\citep{lee_molecule_2024}.}
  \label{fig:PMO_moltype_overall}
\end{figure}

Before the per-task tables, \cref{fig:PMO_moltype_overall} summarises the
central downstream finding across design axes and model sizes. Lower
pretraining loss does not translate into better goal-directed design, and
every axis plateaus in both total PMO score and total docking score beyond
$\sim$5M parameters.

\begin{figure}[htb]
  \centering
  \includegraphics[width=\linewidth]{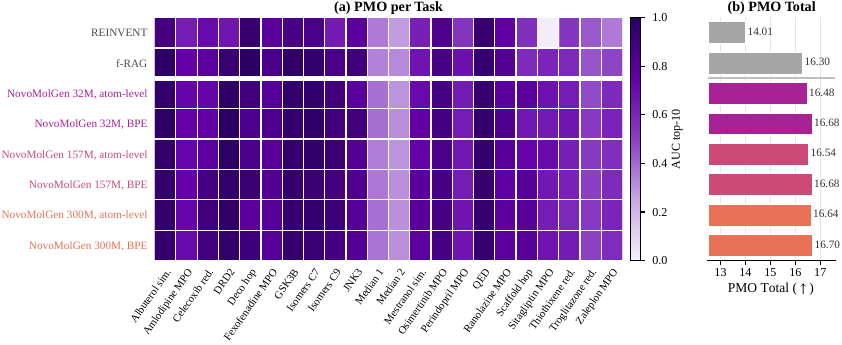}
  \caption{PMO benchmark results for \textbf{NovoMolGen-SMILES} across
    model sizes (32M, 157M, 300M) and tokenization strategies. The
    \textbf{heatmap} (left) shows normalized scores per task, while the
    \textbf{bar chart} (right) presents total scores (higher is
  better). Baselines are REINVENT and \frag.}
  \label{fig:PMO_model_size_all}
\end{figure}

\begin{figure}[htb]
  \centering
  \includegraphics[width=\linewidth]{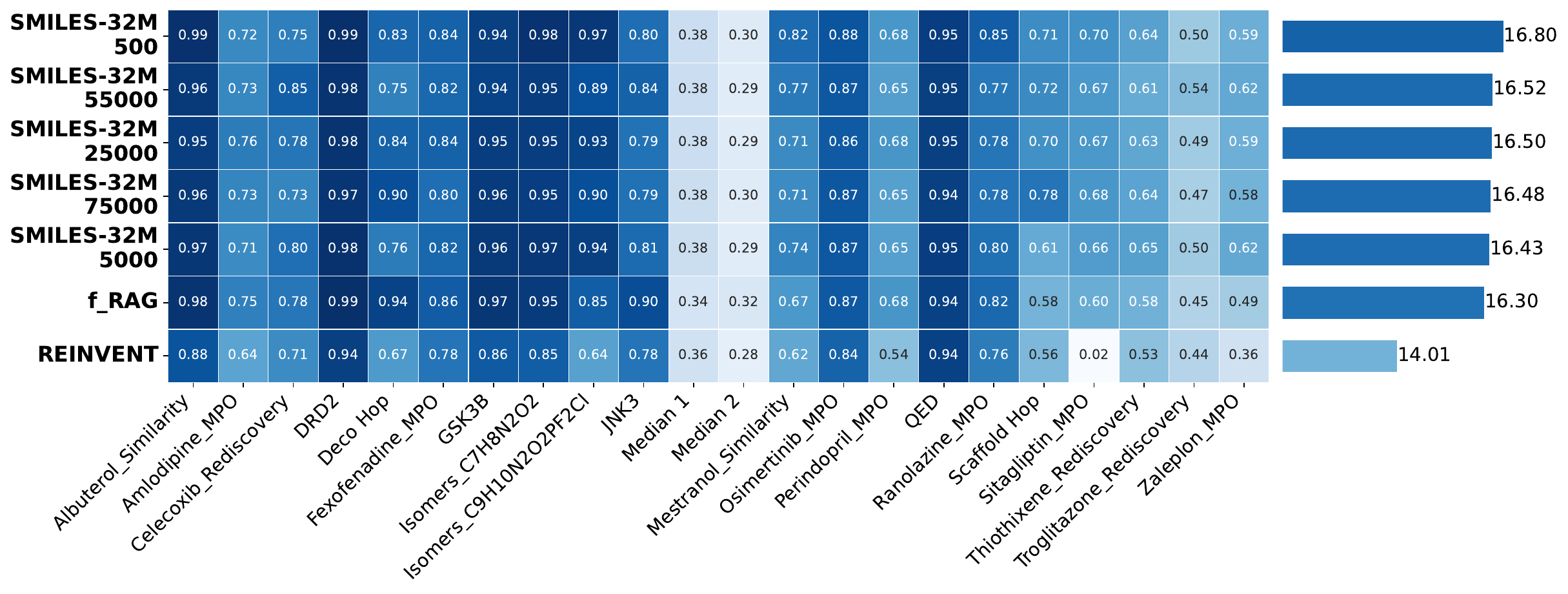}
  \caption{PMO benchmark results for \textbf{NovoMolGen-SMILES-32M
    (AtomWise)} across intermediate training checkpoints. The
    \textbf{heatmap} (left) shows normalized scores per task, while the
    \textbf{bar chart} (right) presents total scores (higher is better).
  Baselines are REINVENT and \frag.}
  \label{fig:PMO_training_progress_full}
\end{figure}

\begin{figure}
  \centering
  \includegraphics[width=0.8\linewidth]{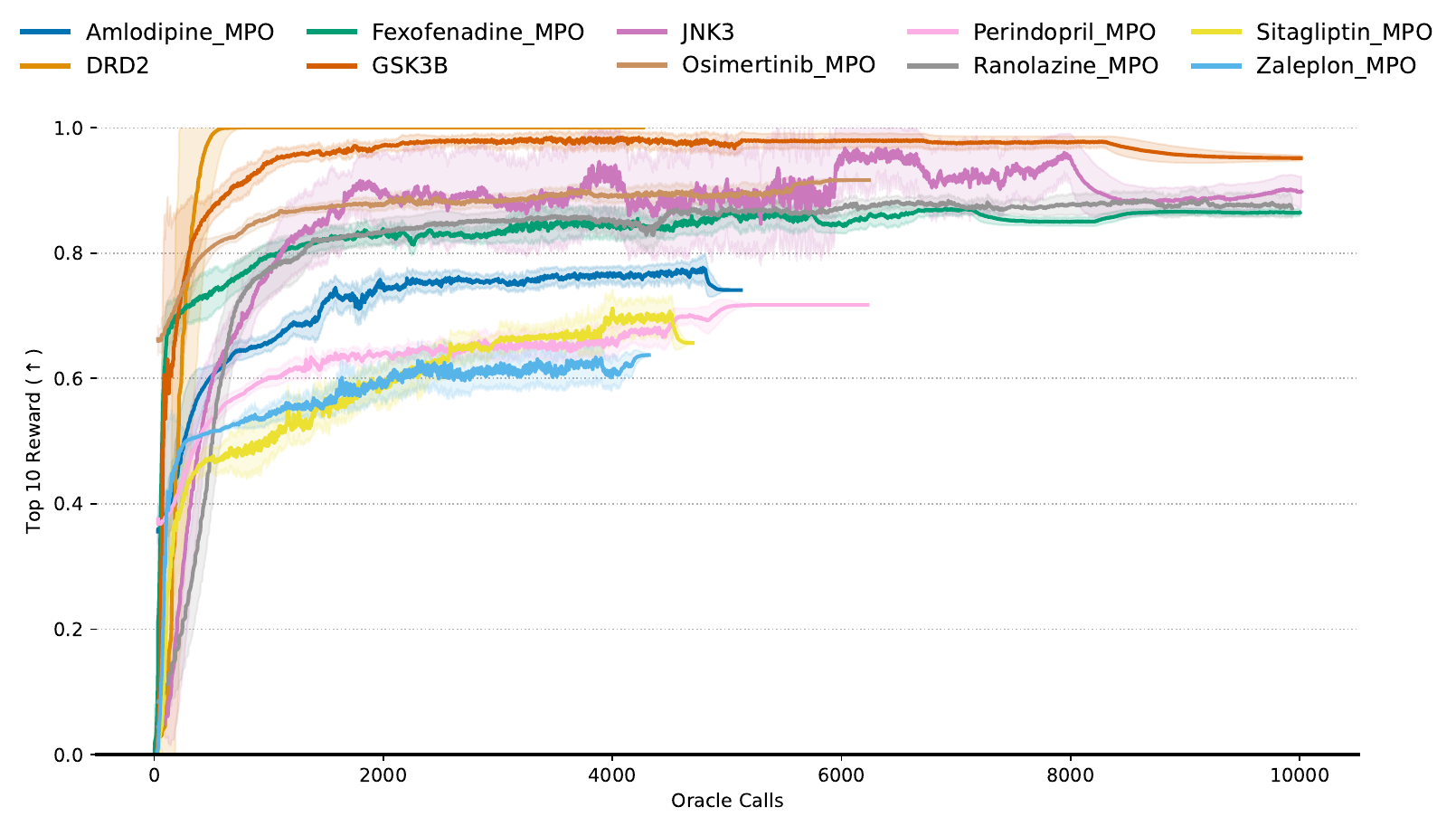}
  \caption{Sample efficiency of
    \textbf{NovoMolGen-SMILES-BPE-32M} on goal-directed generation tasks
    from the PMO benchmark. The plot tracks the top-10 reward as a
    function of oracle calls, showing how quickly the model discovers
  high-reward molecules.}
  \label{fig:pmo_reward_curve}
\end{figure}

This appendix provides a comprehensive analysis of the PMO benchmark
results for \textbf{NovoMolGen}, assessing its performance across
varying model sizes, tokenization strategies, and intermediate
training checkpoints. In \cref{fig:PMO_model_size_all}, the results
are presented, highlighting the influence of model size and
tokenization approaches on the performance of NovoMolGen.
Additionally, \cref{fig:PMO_training_progress_full} tracks the
performance progression of NovoMolGen-32M-Atomwise across
intermediate checkpoints, demonstrating the evolution of model
performance with the optimal hyperparameter configuration and
atomwise tokenization. The evaluation includes comparisons with the
REINVENT and \frag baselines, with the mean and standard deviation of
3 independent runs presented
in~\cref{tab:novomol_comparison_atomwise,tab:novomol_comparison_bpe}.
The reward curve for MPO tasks is shown in~\cref{fig:pmo_reward_curve}.

\begin{figure}[htb]
  \centering
  \includegraphics[width=\linewidth]{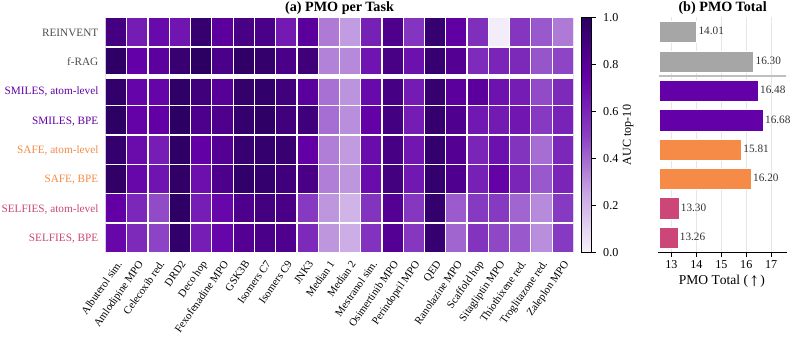}
  \caption{PMO benchmark results for all molecule types
    \textbf{NovoMolGen-32M}. The \textbf{heatmap} (left) shows
    normalized scores per task, while the \textbf{bar chart} (right)
  presents total scores (higher is better). Baselines are REINVENT and \frag.}
  \label{fig:PMO_benchmark_all_moltype}
\end{figure}

\begin{table}[htb]
  \centering
  \caption{PMO \textbf{AUC Top-10} results. Results are reported as
    mean$_{\mathrm{(std)}}$ over three independent model
    initializations for \textbf{SMILES} molecule type.
    \colorbox{aliceblue}{\textbf{Blue}} denotes the best performing
    model, while \colorbox{babypink}{\textbf{Pink}} represents the
  second-best model by mean score.}
  \label{tab:novomol_comparison_atomwise}
  \resizebox{\columnwidth}{!}{%
    \begin{tabular}{lccccc}
      \toprule
      \textbf{Oracle} & \textbf{\frag} & \textbf{REINVENT} &
      \textbf{NovoMolGen‑32M (AtomWise)} & \textbf{NovoMolGen‑157M
      (AtomWise)} & \textbf{NovoMolGen‑300M (AtomWise)}\\
      \midrule
      albuterol\_similarity& {\CC\textbf{0.977$_{(0.002)}$}} &
      $0.882_{(0.006)}$ & $0.958_{(0.003)}$ &
      {\CP\textbf{0.968$_{(0.002)}$}} & $0.954_{(0.002)}$\\
      amlodipine\_mpo& {\CC\textbf{0.749$_{(0.019)}$}} &
      $0.635_{(0.035)}$ & {\CP\textbf{0.732$_{(0.024)}$}} &
      $0.730_{(0.037)}$ & $0.730_{(0.007)}$\\
      celecoxib\_rediscovery& {\CP\textbf{0.778$_{(0.007)}$}} &
      $0.713_{(0.067)}$ & $0.732_{(0.009)}$ & $0.777_{(0.132)}$ &
      {\CC\textbf{0.904$_{(0.022)}$}}\\
      drd2  & $0.936_{(0.011)}$ & $0.666_{(0.044)}$ &
      {\CP\textbf{0.973$_{(0.001)}$}} &
      {\CC\textbf{0.983$_{(0.002)}$}} & $0.972_{(0.001)}$\\
      deco\_hop& {\CC\textbf{0.992$_{(0.000)}$}} &
      {\CP\textbf{0.945$_{(0.007)}$}} & $0.902_{(0.007)}$ &
      $0.853_{(0.115)}$ & $0.785_{(0.130)}$\\
      fexofenadine\_mpo& {\CC\textbf{0.856$_{(0.016)}$}} &
      $0.784_{(0.006)}$ & $0.802_{(0.008)}$ & $0.790_{(0.028)}$ &
      {\CP\textbf{0.803$_{(0.018)}$}}\\
      gsk3b & {\CC\textbf{0.969$_{(0.003)}$}} & $0.865_{(0.043)}$ &
      {\CP\textbf{0.955$_{(0.006)}$}} & $0.949_{(0.005)}$ & $0.941_{(0.012)}$\\
      isomers\_c7h8n2o2& $0.955_{(0.008)}$ &
      $0.852_{(0.036)}$ & {\CP\textbf{0.961$_{(0.001)}$}} &
      {\CC\textbf{0.965$_{(0.002)}$}} & $0.949_{(0.005)}$\\
      isomers\_c9h10n2o2pf2cl& $0.850_{(0.005)}$ & $0.642_{(0.054)}$
      & $0.898_{(0.032)}$ & {\CC\textbf{0.926$_{(0.024)}$}} &
      {\CP\textbf{0.915$_{(0.022)}$}}\\
      jnk3  & {\CC\textbf{0.904$_{(0.004)}$}} & $0.783_{(0.023)}$ &
      $0.787_{(0.014)}$ & {\CP\textbf{0.809$_{(0.066)}$}} & $0.808_{(0.045)}$\\
      median1 & $0.340_{(0.007)}$ & $0.356_{(0.009)}$ &
      {\CC\textbf{0.380$_{(0.003)}$}} & $0.354_{(0.027)}$ &
      {\CP\textbf{0.379$_{(0.002)}$}}\\
      median2 & {\CC\textbf{0.323$_{(0.005)}$}} & $0.276_{(0.008)}$ &
      $0.300_{(0.013)}$ & $0.299_{(0.001)}$ & {\CP\textbf{0.304$_{(0.019)}$}}\\
      mestranol\_similarity& $0.671_{(0.021)}$ & $0.618_{(0.048)}$ &
      $0.712_{(0.047)}$ & {\CP\textbf{0.739$_{(0.068)}$}} &
      {\CC\textbf{0.796$_{(0.046)}$}}\\
      osimertinib\_mpo & $0.866_{(0.009)}$ & $0.837_{(0.009)}$ &
      {\CC\textbf{0.874$_{(0.013)}$}} & $0.858_{(0.001)}$ &
      {\CP\textbf{0.871$_{(0.011)}$}}\\
      perindopril\_mpo & {\CC\textbf{0.681$_{(0.017)}$}} &
      $0.537_{(0.016)}$ & $0.637_{(0.050)}$ & $0.649_{(0.037)}$ &
      {\CP\textbf{0.668$_{(0.011)}$}}\\
      qed & $0.939_{(0.001)}$ & $0.941_{(0.000)}$ & $0.945_{(0.000)}$
      & {\CC\textbf{0.946$_{(0.000)}$}} & {\CP\textbf{0.945$_{(0.000)}$}}\\
      ranolazine\_mpo& {\CC\textbf{0.820$_{(0.016)}$}} &
      $0.760_{(0.009)}$ & $0.784_{(0.007)}$ &
      {\CP\textbf{0.797$_{(0.023)}$}} & $0.779_{(0.024)}$\\
      scaffold\_hop & $0.576_{(0.014)}$ & $0.560_{(0.019)}$ &
      {\CP\textbf{0.783$_{(0.123)}$}} & $0.735_{(0.140)}$ &
      {\CC\textbf{0.791$_{(0.160)}$}}\\
      sitagliptin\_mpo& $0.601_{(0.011)}$ & $0.021_{(0.009)}$ &
      {\CP\textbf{0.677$_{(0.018)}$}} &
      {\CC\textbf{0.708$_{(0.038)}$}} & $0.635_{(0.043)}$\\
      thiothixene\_rediscovery& $0.584_{(0.009)}$ & $0.534_{(0.013)}$
      & {\CC\textbf{0.635$_{(0.018)}$}} &
      {\CP\textbf{0.624$_{(0.050)}$}} & $0.584_{(0.062)}$\\
      troglitazone\_rediscovery& $0.448_{(0.017)}$ &
      $0.441_{(0.032)}$ & $0.470_{(0.042)}$ &
      {\CP\textbf{0.516$_{(0.010)}$}} & {\CC\textbf{0.524$_{(0.058)}$}}\\
      zaleplon\_mpo & $0.486_{(0.004)}$ & $0.358_{(0.062)}$ &
      {\CP\textbf{0.579$_{(0.023)}$}} & $0.559_{(0.013)}$ &
      {\CC\textbf{0.599$_{(0.000)}$}}\\
      \midrule
      \textbf{Sum} & 16.30 & 14.01 & 16.48 & {\CP\textbf{16.54}} &
      {\CC\textbf{16.64}}\\
      \bottomrule
  \end{tabular}}
\end{table}

\begin{table}[htb]
  \centering
  \caption{PMO \textbf{AUC Top-10} results. Results are reported as
    mean$_{\mathrm{(std)}}$ over three independent model
    initializations for \textbf{SMILES} molecule type.
    \colorbox{aliceblue}{\textbf{Blue}} denotes the best performing
    model, while \colorbox{babypink}{\textbf{Pink}} represents the
  second-best model by mean score.}
  \label{tab:novomol_comparison_bpe}
  \resizebox{\columnwidth}{!}{%
    \begin{tabular}{lccccc}
      \toprule
      \textbf{Oracle} & \textbf{\frag} & \textbf{REINVENT} &
      \textbf{NovoMolGen-32M (BPE)} & \textbf{NovoMolGen-157M (BPE)}
      & \textbf{NovoMolGen-300M (BPE)}\\
      \midrule
      albuterol\_similarity& {\CP\textbf{0.977$_{(0.002)}$}} &
      $0.882_{(0.006)}$ & {\CC\textbf{0.981$_{(0.001)}$}} &
      $0.951_{(0.000)}$ & $0.949_{(0.001)}$\\
      amlodipine\_mpo& {\CC\textbf{0.749$_{(0.019)}$}} &
      $0.635_{(0.035)}$ & {\CP\textbf{0.739$_{(0.021)}$}} &
      $0.739_{(0.026)}$ & $0.713_{(0.010)}$\\
      celecoxib\_rediscovery& $0.778_{(0.007)}$ & $0.713_{(0.067)}$ &
      $0.751_{(0.190)}$ & {\CP\textbf{0.872$_{(0.061)}$}} &
      {\CC\textbf{0.890$_{(0.018)}$}}\\
      drd2 & $0.936_{(0.011)}$ & $0.666_{(0.044)}$ &
      {\CC\textbf{0.987$_{(0.001)}$}} & $0.967_{(0.001)}$ &
      {\CP\textbf{0.967$_{(0.000)}$}}\\
      deco\_hop & {\CC\textbf{0.992$_{(0.000)}$}} &
      {\CP\textbf{0.945$_{(0.007)}$}} & $0.844_{(0.121)}$ &
      $0.937_{(0.008)}$ & $0.914_{(0.051)}$\\
      fexofenadine\_mpo & {\CC\textbf{0.856$_{(0.016)}$}} &
      $0.784_{(0.006)}$ & {\CP\textbf{0.837$_{(0.022)}$}} &
      $0.810_{(0.012)}$ & $0.804_{(0.002)}$\\
      gsk3b & {\CC\textbf{0.969$_{(0.003)}$}} & $0.865_{(0.043)}$ &
      {\CP\textbf{0.953$_{(0.013)}$}} & $0.947_{(0.000)}$ & $0.944_{(0.006)}$\\
      isomers\_c7h8n2o2 & {\CP\textbf{0.955$_{(0.008)}$}} &
      $0.852_{(0.036)}$ & {\CC\textbf{0.975$_{(0.003)}$}} &
      $0.929_{(0.006)}$ & $0.928_{(0.006)}$\\
      isomers\_c9h10n2o2pf2cl & $0.850_{(0.005)}$ & $0.642_{(0.054)}$
      & {\CC\textbf{0.897$_{(0.027)}$}} &
      {\CP\textbf{0.876$_{(0.005)}$}} & $0.874_{(0.011)}$\\
      jnk3 & {\CC\textbf{0.904$_{(0.004)}$}} & $0.783_{(0.023)}$ &
      {\CP\textbf{0.899$_{(0.049)}$}} & $0.839_{(0.044)}$ & $0.818_{(0.025)}$\\
      median1 & $0.340_{(0.007)}$ & $0.356_{(0.009)}$ &
      {\CC\textbf{0.388$_{(0.003)}$}} & $0.364_{(0.013)}$ &
      {\CP\textbf{0.374$_{(0.002)}$}}\\
      median2 & {\CC\textbf{0.323$_{(0.005)}$}} & $0.276_{(0.008)}$ &
      {\CP\textbf{0.310$_{(0.031)}$}} & $0.305_{(0.016)}$ & $0.307_{(0.021)}$\\
      mestranol\_similarity& $0.671_{(0.021)}$ & $0.618_{(0.048)}$ &
      $0.745_{(0.028)}$ & {\CC\textbf{0.789$_{(0.018)}$}} &
      {\CP\textbf{0.780$_{(0.040)}$}}\\
      osimertinib\_mpo & $0.866_{(0.009)}$ & $0.837_{(0.009)}$ &
      {\CC\textbf{0.885$_{(0.012)}$}} &
      {\CP\textbf{0.873$_{(0.008)}$}} & $0.870_{(0.008)}$\\
      perindopril\_mpo & {\CC\textbf{0.681$_{(0.017)}$}} &
      $0.537_{(0.016)}$ & $0.635_{(0.018)}$ & $0.636_{(0.036)}$ &
      {\CP\textbf{0.669$_{(0.049)}$}}\\
      qed & $0.939_{(0.001)}$ & $0.941_{(0.000)}$ &
      {\CC\textbf{0.947$_{(0.000)}$}} & $0.945_{(0.000)}$ &
      {\CP\textbf{0.945$_{(0.000)}$}}\\
      ranolazine\_mpo& {\CP\textbf{0.820$_{(0.016)}$}} &
      $0.760_{(0.009)}$ & {\CC\textbf{0.830$_{(0.010)}$}} &
      $0.760_{(0.018)}$ & $0.780_{(0.012)}$\\
      scaffold\_hop & $0.576_{(0.014)}$ & $0.560_{(0.019)}$ &
      $0.650_{(0.002)}$ & {\CC\textbf{0.803$_{(0.063)}$}} &
      {\CP\textbf{0.790$_{(0.134)}$}}\\
      sitagliptin\_mpo & $0.601_{(0.011)}$ & $0.021_{(0.009)}$ &
      $0.643_{(0.034)}$ & {\CP\textbf{0.650$_{(0.040)}$}} &
      {\CC\textbf{0.673$_{(0.044)}$}}\\
      thiothixene\_rediscovery & $0.584_{(0.009)}$ &
      $0.534_{(0.013)}$ & {\CC\textbf{0.657$_{(0.012)}$}} &
      {\CP\textbf{0.615$_{(0.010)}$}} & $0.638_{(0.003)}$\\
      troglitazone\_rediscovery & $0.448_{(0.017)}$ &
      $0.441_{(0.032)}$ & {\CC\textbf{0.517$_{(0.032)}$}} &
      {\CP\textbf{0.497$_{(0.016)}$}} & $0.496_{(0.003)}$\\
      zaleplon\_mpo & $0.486_{(0.004)}$ & $0.358_{(0.062)}$ &
      {\CC\textbf{0.611$_{(0.027)}$}} &
      {\CP\textbf{0.578$_{(0.041)}$}} & $0.574_{(0.010)}$\\
      \midrule
      \textbf{Sum} & 16.30 & 14.01 & {\CP\textbf{16.68}} &
      {\CP\textbf{16.68}} & {\CC\textbf{16.70}}\\
      \bottomrule
  \end{tabular}}%
\end{table}

\begin{figure}
  \centering
  \includegraphics[width=0.74\linewidth]{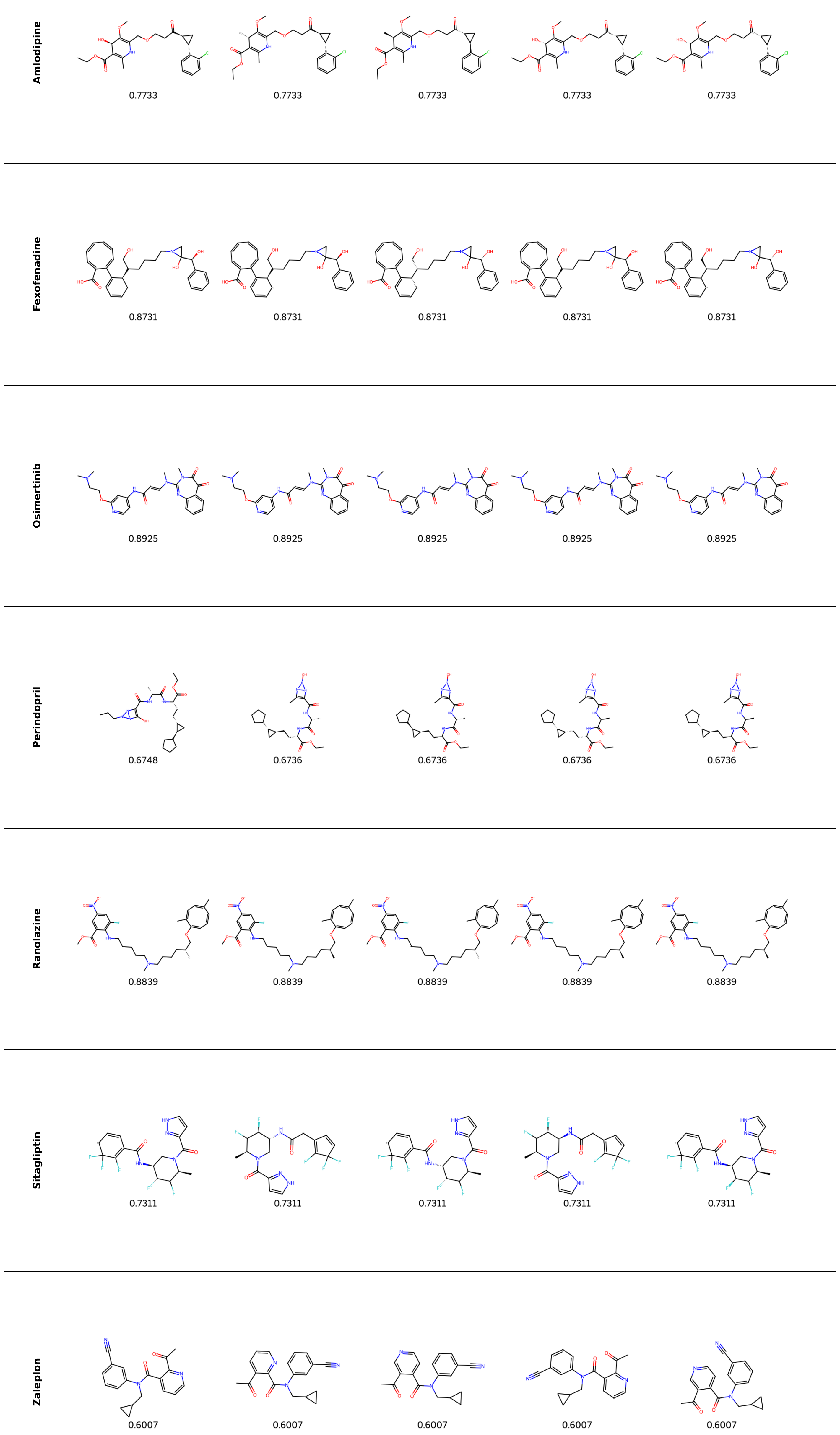}
  \caption{Examples of the generated top-5
    molecules from a single run of \textbf{NovoMolGen-SMILES-BPE-32M}.
  The scores are provided at the bottom of each generated molecule.}
  \label{fig:molecules_pmo}
\end{figure}

\section{Protein-Ligand Docking}
\label{sec:appendix_docking}
The docking score reflects the strength of interaction between a
ligand and its target protein. In this study, we focus on optimizing
binding affinity for five human proteins, each implicated in various diseases:

\begin{enumerate}
  \item \textbf{\textit{parp1}}: Poly(ADP-ribose) polymerase 1
    (PARP1) is a nuclear enzyme involved in DNA repair,
    transcriptional regulation, and cell survival. PARP1 inhibitors
    are widely studied for cancer therapy, particularly in tumors
    with defective DNA repair mechanisms.
  \item \textbf{\textit{fa7}}: Factor VII (FA7), also known as
    proconvertin, is a coagulation protein essential for blood
    clotting. Deficiencies in FA7 are associated with bleeding
    disorders, including hemophilia-like conditions.
  \item \textbf{\textit{5ht1b}}: The 5-hydroxytryptamine receptor 1B
    (5-HT1B) is a serotonin receptor involved in neurotransmitter
    regulation. It plays a role in mood disorders, anxiety, and
    migraine, making it a therapeutic target for psychiatric and
    neurological conditions.
  \item \textbf{\textit{braf}}: B-Raf (serine/threonine-protein
    kinase B-Raf) is a key regulator of the MAP kinase/ERK signaling
    pathway, which controls cell proliferation and differentiation.
    Mutations in BRAF are commonly associated with various cancers,
    including melanoma and colorectal cancer.
  \item \textbf{\textit{jak2}}: Janus kinase 2 (JAK2) is a tyrosine
    kinase involved in the JAK/STAT signaling pathway, which
    regulates hematopoiesis and immune function. Mutations in JAK2
    are linked to myeloproliferative disorders such as polycythemia
    vera and essential thrombocythemia.
\end{enumerate}

\begin{figure}
  \centering
  \includegraphics[width=0.8\linewidth]{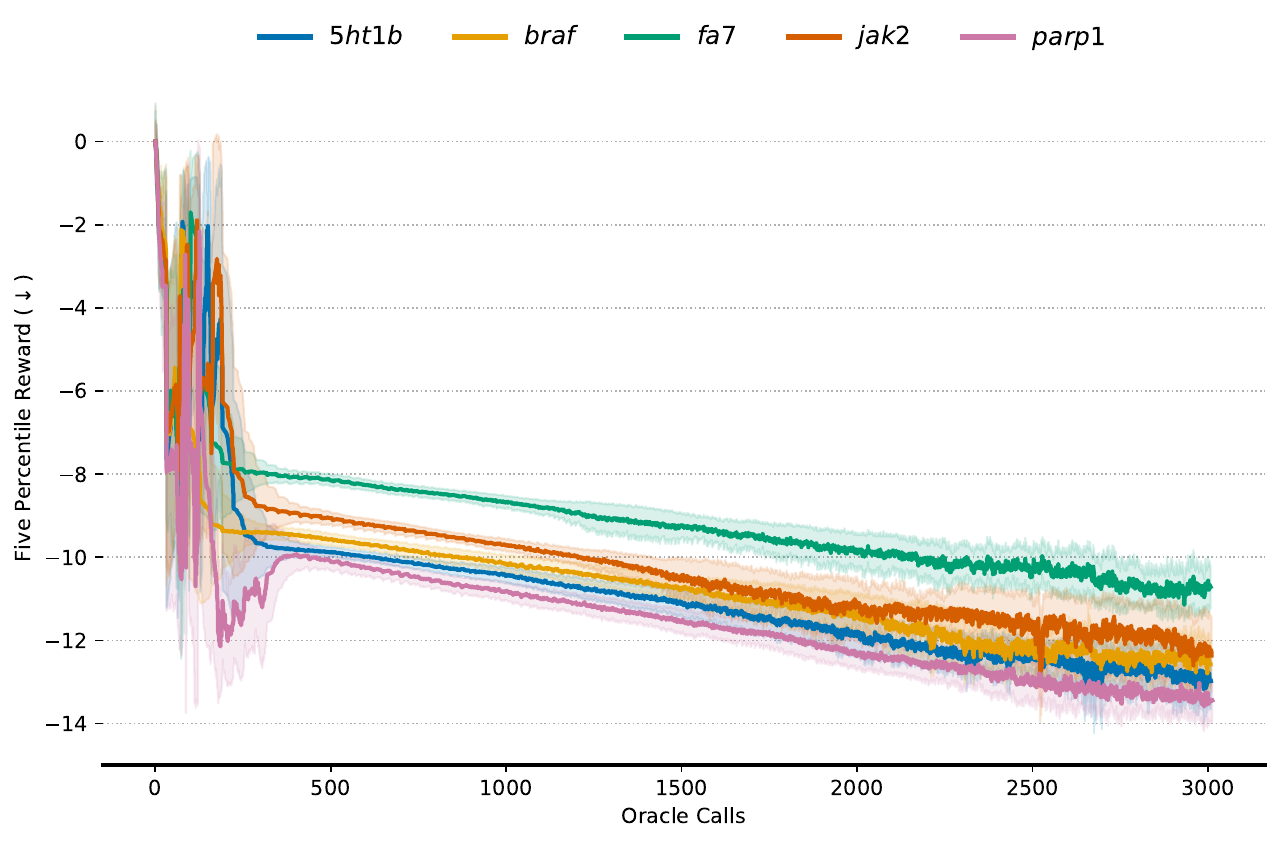}
  \caption{Sample efficiency of \textbf{NovoMolGen-SMILES-BPE-32M} on
    Protein-ligand docking goal-directed generation tasks. The plot
    tracks the top 5th percentile of the reward distribution as a
    function of oracle calls, demonstrating our model's ability to
  rapidly discover high-reward molecules.}
  \label{fig:docking_reward_curve}
\end{figure}

\begin{figure}
  \centering
  \includegraphics[width=0.9\linewidth]{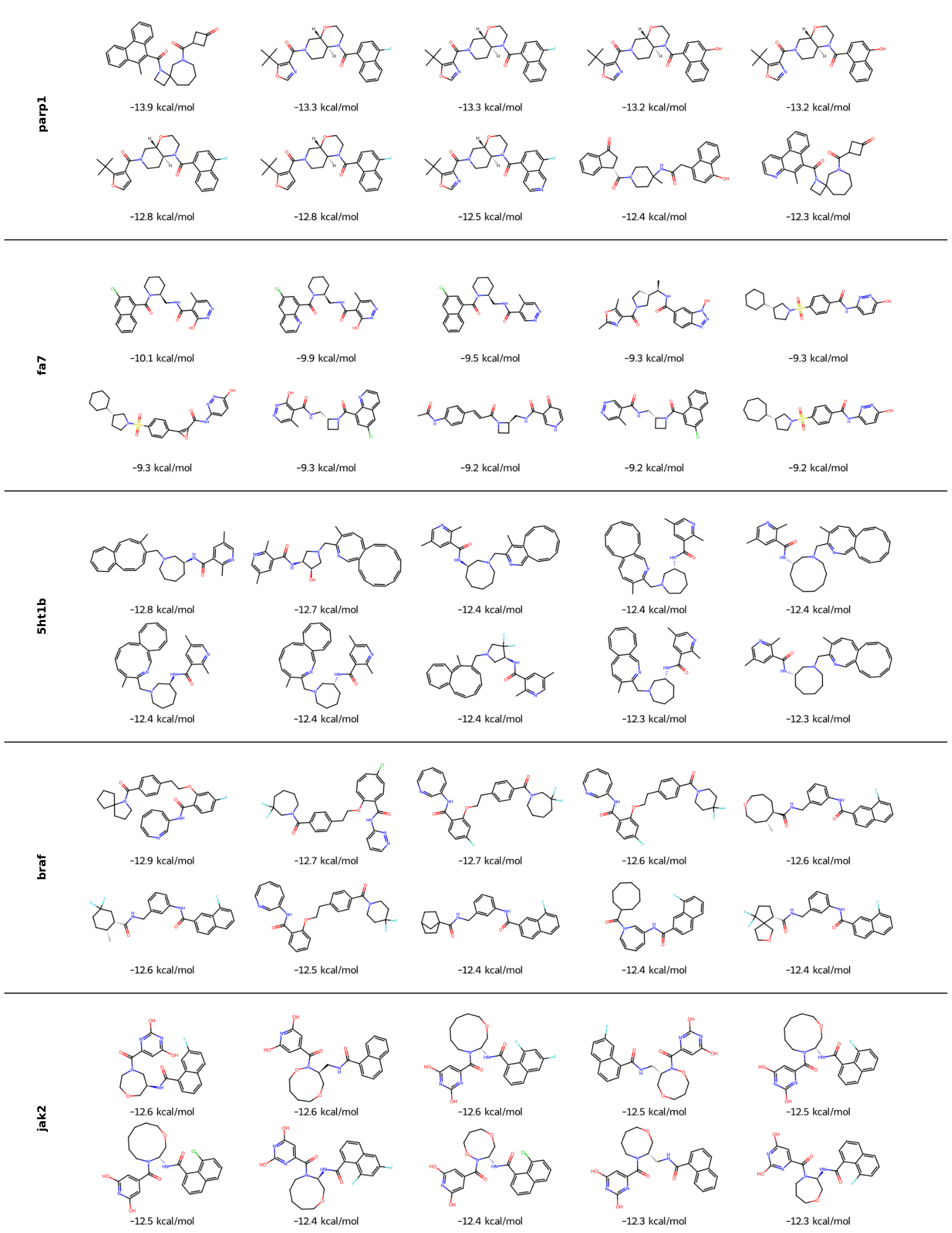}
  \caption{Examples of the generated top-10 molecules from a single run of
    \textbf{NovoMolGen-SMILES-BPE-32M}, shown for the stricter showcase
    filter QED $>0.5$ and SA $<4$ rather than the SA $<5$ used in the
    novel-hit definition. The docking scores are given at the bottom of
  each molecule.}
  \label{fig:molecules_docking}
\end{figure}

We employ the docking program QuickVina 2~\citep{alhossary_fast_2015}
to compute docking scores, setting the exhaustiveness parameter to 1.
The timeout for 3D conformer generation is set to 20 seconds, while
the docking score calculation timeout is set to 50 seconds. PDB
files, box sizes and center coordinates for the target proteins are
taken from~\citet{lee_exploring_2023}. To compute the maximum
Tanimoto similarity with the training molecules, we use the ZINC-250K
molecules to identify the unique and novel compounds. This approach
is adopted because searching against the full training corpus is
computationally infeasible. The novelty results are presented
in~\cref{tab:novelty_docking} and the visualization of generated
novel hits are presented in~\cref{fig:molecules_docking}. The reward
curve is shown in~\cref{fig:docking_reward_curve}.

\Cref{tab:docking_scores} reports the top-5\% novel docking score per
target for NovoMolGen and the fragment-based, 3D and diffusion baselines.
NovoMolGen reaches a better score than every baseline on all five targets
from 5M parameters, and the total score improves only marginally from
$-65.8$ at 5M to $-66.2$ at 1B, so the docking benefit of scale is small
once a compact prior is in place.

\begin{table}[htb]
  \centering
  \caption{Top-5\% novel docking score per target (kcal/mol,
    $\downarrow$). Baselines are taken
    from~\citet{lee_molecule_2024}. \colorbox{aliceblue}{\textbf{Blue}}
    denotes the best score per target. Totals are the reported sums, so
  rounded per-target entries may not add up exactly.}
  \label{tab:docking_scores}
  \resizebox{\textwidth}{!}{%
    \begin{tabular}{lcccccc}
      \toprule
      \textbf{Method} & parp1 & fa7 & 5ht1b & braf & jak2 & \textbf{Sum} \\
      \midrule
      REINVENT & $-8.7$ & $-7.2$ & $-8.8$ & $-8.4$ & $-8.2$ & $-41.2$ \\
      MOOD     & $-10.9$ & $-8.2$ & $-11.2$ & $-11.1$ & $-10.2$ & $-51.4$ \\
      GEAM     & $-12.9$ & $-9.9$ & $-12.4$ & $-12.3$ & $-11.8$ & $-59.3$ \\
      $f$-RAG  & $-12.9$ & $-9.9$ & $-12.7$ & $-12.4$ & $-11.8$ & $-59.8$ \\
      \midrule
      NovoMolGen-0.5M & $-13.6$ & $-10.4$ & $-13.4$ & $-13.1$ &
      $-12.5$ & $-63.0$ \\
      NovoMolGen-5M   & $-14.2$ & $-10.9$ & $-14.0$ & $-13.6$ &
      $-13.0$ & $-65.8$ \\
      NovoMolGen-32M  & \CC$\mathbf{-14.3}$ & \CC$\mathbf{-11.0}$ &
      \CC$\mathbf{-14.1}$ & \CC$\mathbf{-13.8}$ & $-13.1$ &
      \CC$\mathbf{-66.3}$ \\
      NovoMolGen-157M & \CC$\mathbf{-14.3}$ & $-10.9$ & $-14.0$ &
      \CC$\mathbf{-13.8}$ & \CC$\mathbf{-13.2}$ & \CC$\mathbf{-66.3}$ \\
      NovoMolGen-300M & \CC$\mathbf{-14.3}$ & $-10.9$ & $-14.0$ &
      \CC$\mathbf{-13.8}$ & $-13.1$ & $-66.1$ \\
      NovoMolGen-1B   & \CC$\mathbf{-14.3}$ & \CC$\mathbf{-11.0}$ &
      $-14.0$ & \CC$\mathbf{-13.8}$ & $-13.1$ & $-66.2$ \\
      \bottomrule
  \end{tabular}}
\end{table}

\begin{table}[htb]
  \centering
  \caption{Novelty comparison on protein-ligand docking tasks. We
    report Novelty (\%), calculated with respect to the ZINC-250k
    dataset. Results show the mean and standard deviation from 3
    independent runs. Baseline data is sourced
    from~\citet{lee2024drug}. \colorbox{aliceblue}{\textbf{Blue}}
    denotes the best model by mean score, while
    \colorbox{babypink}{\textbf{Pink}} represents the second-best model
  by mean score.}
  \label{tab:novelty_docking}
  \resizebox{\textwidth}{!}{%
    \begin{tabular}{@{}lccccc@{}}
      \toprule
      \textbf{Method} & parp1 & fa7 & 5ht1b & braf & jak2 \\
      \midrule
      REINVENT & $9.894_{(2.178)}$ & $10.731_{(1.516)}$ &
      $11.605_{(3.688)}$ & $8.715_{(2.712)}$ & $11.456_{(1.793)}$ \\
      MORLD & {\CC $\mathbf{98.433_{(1.189)}}$} & {\CC
      $\mathbf{97.967_{(1.764)}}$} & {\CC
      $\mathbf{98.787_{(0.743)}}$} & {\CC
      $\mathbf{96.993_{(2.787)}}$} & {\CC $\mathbf{97.720_{(0.995)}}$} \\
      HierVAE & $60.453_{(17.165)}$ & $24.853_{(15.416)}$ &
      $48.107_{(1.988)}$ & $59.747_{(16.403)}$ & $85.200_{(14.262)}$ \\
      FREED & $74.640_{(2.953)}$ & $78.787_{(2.132)}$ &
      $75.027_{(5.194)}$ & $73.653_{(4.312)}$ & $75.907_{(5.916)}$ \\
      MOOD & $84.180_{(2.123)}$ & $83.180_{(1.519)}$ &
      $84.613_{(0.822)}$ & $87.413_{(0.830)}$ & $83.273_{(1.455)}$ \\
      GEAM & {\CP $\mathbf{88.611_{(3.107)}}$} & {\CP
      $\mathbf{89.378_{(2.619)}}$} & {\CP
      $\mathbf{84.222_{(2.968)}}$} & {\CP
      $\mathbf{90.322_{(3.467)}}$} & {\CP
      $\mathbf{89.222_{(1.824)}}$} \\ \midrule \midrule

      \textsc{SMILES-32M-Atomwise} & $30.34_{(7.0)}$ &
      $45.84_{(26.0)}$ &
      $15.38_{(2.0)}$ &
      $23.14_{(3.0)}$ &
      $41.48_{(8.0)}$   \\

      \textsc{SMILES-157M-Atomwise} & $54.31_{(5.0)}$ &
      $46.26_{(8.0)}$ &
      $27.28_{(9.0)}$ &
      $29.01_{(16.0)}$ &
      $44.68_{(4.0)}$   \\

      \textsc{SMILES-300M-Atomwise} & $37.78_{(16.0)}$ &
      $66.12_{(10.0)}$ &
      $42.69_{(17.0)}$ &
      $33.23_{(18.0)}$ &
      $37.83_{(6.0)}$   \\ \midrule

      \textsc{SMILES-32M-BPE} & $56.28_{(6.0)}$ &
      $62.02_{(6.0)}$ &
      $33.69_{(8.0)}$ &
      $52.73_{(8.0)}$ &
      $48.42_{(7.0)}$  \\

      \textsc{SMILES-157M-BPE} & $45.98_{(5.0)}$ &
      $62.06_{(2.0)}$ &
      $43.62_{(4.0)}$ &
      $63.26_{(5.0)}$ &
      $51.64_{(4.0)}$  \\

      \textsc{SMILES-300M-BPE} & $50.12_{(9.0)}$ &
      $31.37_{(4.0)}$ &
      $40.44_{(3.0)}$ &
      $53.24_{(3.0)}$ &
      $45.87_{(1.0)}$  \\

      \bottomrule

    \end{tabular}
  }
\end{table}

\end{document}